\documentclass[11pt,a4paper]{article}
\usepackage{times,latexsym}
\usepackage{url}
\usepackage[T1]{fontenc}
\usepackage{forest}
\usepackage{enumitem}
\usepackage[acceptedWithA]{tacl2021v1}
\usepackage{xspace,mfirstuc,tabulary}
\usepackage{booktabs}
\usepackage{graphicx}
\usepackage{amssymb}
\usepackage{bbding}
\usepackage{xcolor,colortbl}
\usepackage{multirow}

\usepackage{CJKutf8}
\usepackage{pifont}% http://ctan.org/pkg/pifont
\newcommand{\cmark}{\ding{51}}%
\newcommand{\xmark}{\ding{55}}%
\usepackage{arydshln}
\usepackage{hyperref}
\usepackage{fontawesome}
\newcommand{\revise}[1]{\textcolor{black}{#1}}
\newcommand{\revisenew}[1]{\textcolor{black}{#1}}

\newif\iftaclinstructions
\taclinstructionsfalse % AUTHORS: do NOT set this to true
\iftaclinstructions
\renewcommand{\confidential}{}
\renewcommand{\anonsubtext}{(No author info supplied here, for consistency with
TACL-submission anonymization requirements)}
\newcommand{\instr}
\fi

\iftaclpubformat % this "if" is set by the choice of options

\else

\fi

\title{\textit{\includegraphics[width=16pt,height=16pt]{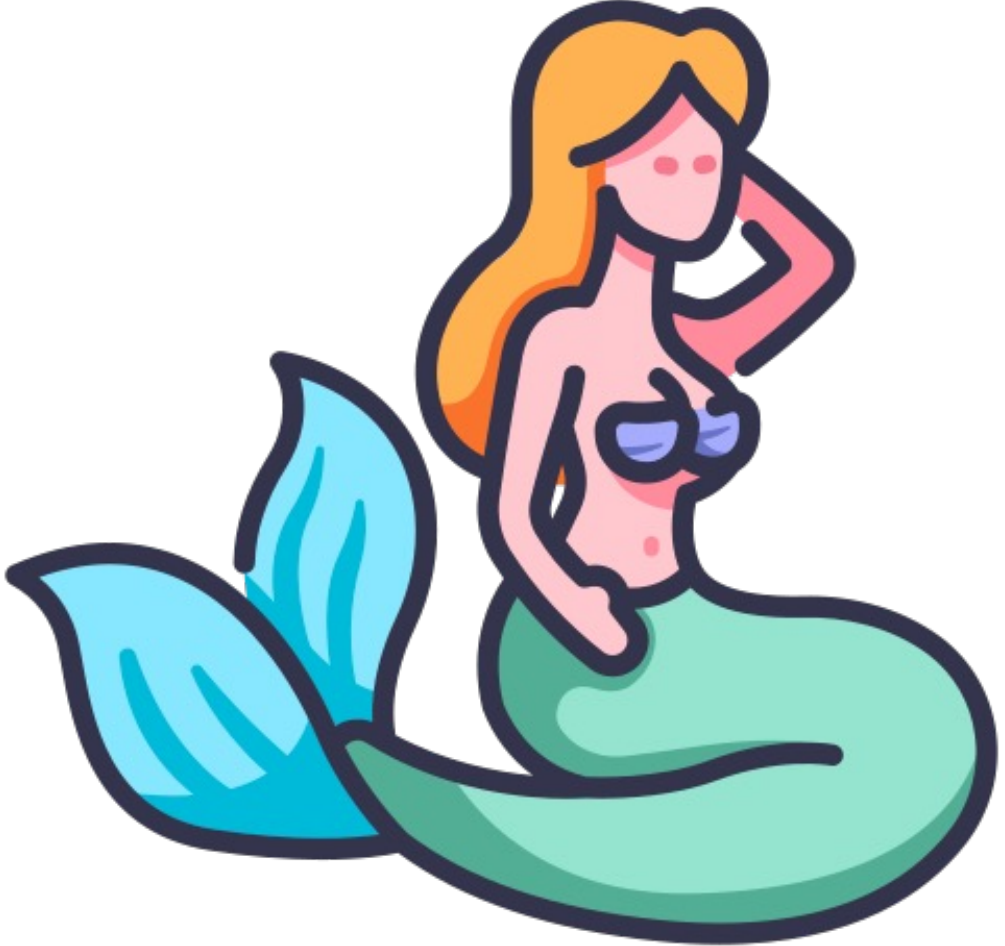}~Siren's Song in the AI Ocean}: \\ A Survey on Hallucination in Large Language Models}

\author{
  Yue Zhang$^{\spadesuit}$\Thanks{This survey paper was completed during Yue Zhang (\href{mailto:yzhang21@stu.suda.edu.cn}{yzhang21@stu.suda.edu.cn}), Yafu Li, Tingchen Fu, Yu Zhang and Chen Xu's internships at Tencent AI Lab.}\hspace{0.5mm}, 
  Yafu Li$^{\diamondsuit}$\hspace{0.5mm}, 
  Leyang Cui$^{\heartsuit}$\Thanks{Corresponding author (\href{mailto:leyangcui@tencent.com}{leyangcui@tencent.com}).}\hspace{0.5mm} ,
  Deng Cai$^{\heartsuit}$\hspace{0.5mm}, Lemao Liu$^{\heartsuit}$\hspace{0.5mm} \\ \bf{Tingchen Fu\textsuperscript{\faStarO}\hspace{0.5mm}, Xinting Huang$^{\heartsuit}$\hspace{0.5mm}, Enbo Zhao$^{\heartsuit}$\hspace{0.5mm}, Yu Zhang$^{\spadesuit}$\hspace{0.5mm}, Yulong Chen$^{\diamondsuit}$\hspace{0.5mm}}\\
  \bf{Chen Xu}$^{\heartsuit}$,
   \bf{Longyue Wang$^{\heartsuit}$\hspace{0.5mm},  Anh Tuan Luu\textsuperscript{\faSunO}\hspace{0.5mm}, Wei Bi$^{\heartsuit}$\hspace{0.5mm}, Freda Shi\textsuperscript{\faMoonO}\hspace{0.5mm}, Shuming Shi$^{\heartsuit}$\hspace{0.5mm}}\vspace{0.2cm}   \\
   $^\heartsuit$ Tencent AI lab \ \ $^\spadesuit$ Soochow University \ \ $^\diamondsuit$Zhejiang University   \\
   \textsuperscript{\faStarO} Renmin University of China \ \
 \textsuperscript{\faSunO}  Nanyang Technological University \ \  \\ \textsuperscript{\faMoonO} Toyota Technological Institute at Chicago \\ 
}

\begin{document}
\maketitle

\begin{CJK}{UTF8}{gkai}

% \vspace{2cm}
% FS TODO: 
% 1) adding more concrete content to the abstract

\begin{abstract}
% The advent of large language models (LLMs) has ignited a beacon of hope in the pursuit of artificial general intelligence (AGI). 
While large language models (LLMs) have demonstrated remarkable capabilities across a range of downstream tasks, a significant concern revolves around their propensity to exhibit \textit{hallucinations}: LLMs occasionally generate content that diverges from the user input, contradicts previously generated context, or misaligns with established world knowledge.
This phenomenon poses a substantial challenge to the reliability of LLMs in real-world scenarios.
In this paper, we survey recent efforts on the detection, explanation, and mitigation of hallucination, with an emphasis on the unique challenges posed by LLMs.
We present taxonomies of the LLM hallucination phenomena and evaluation benchmarks, analyze existing approaches aiming at mitigating LLM hallucination, and discuss potential directions for future research. 
% While the issue of hallucination has been studied in the traditional natural language generation (NLG), addressing it in the era of LLM presents a unique set of challenges.
% For instance, LLMs are designed to solve various tasks instead of a single one, and 
% For instance, the generated content may not have a clear right or wrong answer, making it difficult to identify and correct hallucinations.
% While the issue of hallucination has been studied in the traditional natural language generation community, addressing it in the context of LLMs presents a unique set of challenges.
% The intricate complexity and potential black-box property of LLMs make this issue particularly daunting.

% We begin by examining research aimed at uncovering the sources of LLM hallucination. 
% Next, we delve into the benchmarks and metrics designed for evaluation of LLM hallucination. 
% Subsequently, we categorize and discuss methods and strategies to alleviate LLM hallucination. 
% Lastly, we identify promising avenues for future research and provide insights to stimulate further exploration. We envision this survey as a valuable guide for researchers intrigued by this critical challenge in LLMs.
% To facilitate ongoing research, we consistently update related open-source materials, which can be accessed at \url{https://github.com/HillZhang1999/llm-hallucination-survey}.
% yafu
% 1. difference with previous work on hallucination: LLM
    % a. definition: *
    % b. difficulty, compared with MT/summarization, open-ended \& open-domain
    
% 2. what's left for the hallucination of LLMS? \textbf{evaluation}/metric/benchmark, training methods

\end{abstract}

\section{Introduction}
% Introduce LLM
% Problems in LLM, including Hallucination (definition, harmfulness)
% Why we need this survey? the characteristic of hallucination in LLM 
% Content of this survey: 1) source; 2) evaluation; 3) mitigation; 4) future directions;

Large language models (LLMs), particularly characterized by their substantial number of parameters, have arisen as a promising cornerstone for the development of natural language processing (NLP) and artificial intelligence \citep{zhao2023survey}. 
With proper alignment techniques, such as supervised finetuning \citep[SFT;][]{zhang2023instruction} and reinforcement learning from human feedback \citep[RLHF;][]{ouyang2022training, fernandes2023bridging}, recent LLMs \citep{chatgpt,touvron2023llama2,openai2023gpt4} have exhibited strong capabilities in solving various downstream tasks.

% Generative LLMs, such as ChatGPT \citep{chatgpt}, \textsc{Llama 2-Chat} \citep{touvron2023llama2}, and GPT-4 \citep{openai2023gpt4}, are particularly characterized by their substantial number of parameters.

\begin{figure}[t!]
\centering
\vspace{10pt}
\includegraphics[scale=0.53]{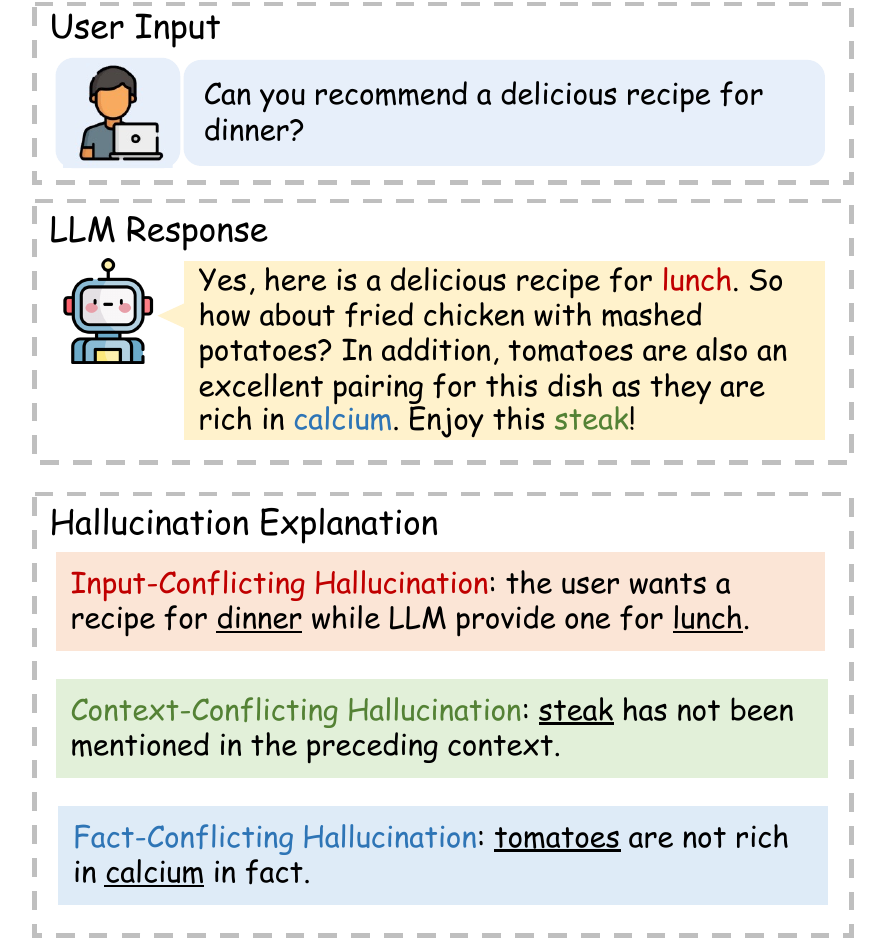}
\caption{Three types of hallucinations occurred in LLM responses (best viewed in color).}
\label{fig:example:intro}
\vspace{-0.2cm}
\end{figure}

Nonetheless, as exemplified in Figure~\ref{fig:example:intro}, LLMs, despite their remarkable success, occasionally produce outputs that, while seemingly plausible, deviate from \revise{users' input} \citep{adlakha2023evaluating}, previously generated context \citep{liu-etal-2022-token}, or factual knowledge \citep{min2023factscore, muhlgay2023generating, li2023halueval}---this phenomenon is commonly referred to as hallucination, which significantly undermines the reliability of LLMs in real-world scenarios \citep{kaddour2023challenges}. 
For instance, LLMs can potentially fabricate erroneous medical diagnoses or treatment plans that lead to tangible real-life risks \citep{umapathi2023med}.

While hallucination in conventional natural language generation (NLG) settings has been widely studied \citep{nlg_hallucination_survey}, understanding and addressing the hallucination problem within the realm of LLMs encounters \revise{more significant} challenges introduced by:
\begin{enumerate}[leftmargin=*,topsep=2pt]

\item \textbf{Massive training data}: in contrast to carefully curating data for a specific task, LLM pre-training uses trillions of tokens obtained from the web, making it difficult to eliminate fabricated, outdated or biased information; 
\item \textbf{Versatility of LLMs}: general-purpose LLMs are expected to excel in cross-task, cross-lingual, and cross-domain settings, posing challenges for comprehensive evaluation and mitigation of hallucination.
\item \textbf{Imperceptibility of errors}: as a byproduct of their strong abilities, LLMs may generate false information that \revise{initially seems to be} highly plausible, making it challenging for models or even humans to detect hallucinations. 
\end{enumerate}
In addition, the RLHF process \citep{ouyang2022training}, the vague knowledge boundary \citep{ren2023boundary} and the black-box property of LLMs \citep{sun2022black} also complicate the detection, explanation, and mitigation of hallucination in LLMs. 
There has been a notable upsurge in cutting-edge research dedicated to addressing the aforementioned challenges, which strongly motivates us to compile this survey.

\begin{figure*}[t!]
\centering
\includegraphics[scale=0.5]{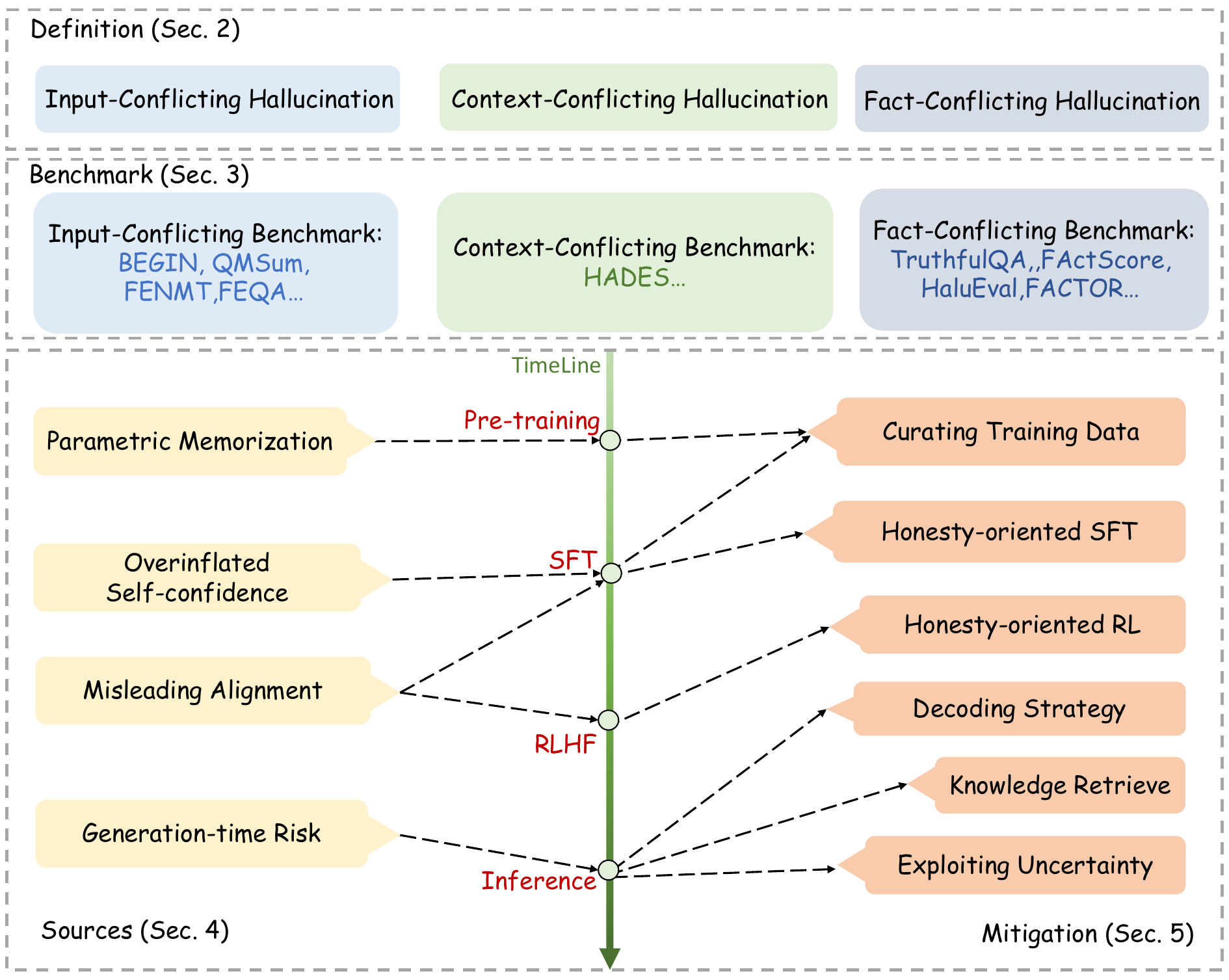}
\caption{The overview structure of this paper: We initially categorize LLM hallucinations into three distinct types and then introduce corresponding evaluation benchmarks. Subsequently, we explore the source of hallucinations and discuss mitigation strategies throughout the life cycle of LLMs (pre-training$\rightarrow$SFT$\rightarrow$RLHF$\rightarrow$inference).}
% Typology
% We first discuss the definition of LLM hallucination in $\S$~\ref{sec:def}}
\label{fig:overview}
\vspace{-0.2cm}
\end{figure*}

\revise{In this survey, we have selected papers that offer novel insights, demonstrate robust experimental results, or present reliable evaluation methods. We have also chosen recent publications from the past few years to ensure the timeliness of our work.} We organize this paper as follows, as also depicted in Figure \ref{fig:overview}.
We first introduce the background of LLMs and offer our definition of hallucination in LLMs (\S\ref{sec:def}). 
Next, we introduce relevant benchmarks and metrics (\S\ref{sec:evaluation}). 
Subsequently, we discuss potential sources of LLM hallucinations (\S\ref{sec:cause}), and provide an in-depth review of recent work towards addressing the problem (\S\ref{sec:solution}). 
Finally, we present forward-looking perspectives (\S\ref{sec:outlook}).
We will consistently update the related open-source materials, which can be accessed at \url{https://github.com/HillZhang1999/llm-hallucination-survey}.

\section{Hallucination in the Era of LLM}
\label{sec:def}
% This section provides an overview of the background knowledge necessary for this survey, including the development status of LLMs and the definition of hallucination in the context of LLMs.
We begin this section by overviewing the history of LLMs (\S\ref{sec:llm-history}). 
Next, we present our definition of LLM hallucination, by breaking it down into three sub-categories (\S\ref{sec:def_hallucination}). 
In addition, we discuss the unique challenges of hallucination in LLMs (\S\ref{sec:unique-challenges}), and compare hallucination with other prevalent problems that are frequently encountered in the realm of LLMs (\S\ref{sec:difference-from-other-problems}).

\subsection{Large Language Models}
\label{sec:llm-history}
% LM -> PLM -> LLM
% The task of language models (LMs) is to predict the likelihood of forthcoming (auto-regressive LMs~\citep{radford2019language,chowdhery2022palm,touvron2023llama}) or missing (masked LMs~\citep{devlin-etal-2019-bert,liu2019roberta,clark2019electra}) tokens. 

% To mirror the cognitive process humans undergo when formulating sentences, language model (LM), an approach for predicting the next token given a prefix~\citep{radford2019language,chowdhery2022palm,touvron2023llama} \footnote{There is another LM variant to predict the missing tokens given a corrupted sequence~\citep{devlin-etal-2019-bert,liu2019roberta, lan2019albert}.}, is designed.
% The language model (LM) is designed to mimic humans' cognitive processes when constructing sentences. 
An important category of LLMs is autoregressive language models \citet{radford2019language,chowdhery2022palm,touvron2023llama}.
These models take Transformers \citep{vaswani2017attention} as the backbone, and predict the next token based on previous tokens.\footnote{Another variant of language models predicts masked tokens in a corrupted sequence \citet{devlin-etal-2019-bert,liu2019roberta,lan2019albert}.}
Prior to the widespread adoption of Transformers, autoregressive language models were built on the backbones of n-grams \citep{bickel2005predicting,pauls2011faster} and recurrent neural networks \citep{mikolov2010recurrent}, and have been applied to various NLG tasks such as summarization \citep{nallapati2017summarunner} and dialogue generation \citep{chen2017survey}. 
%As an essential component in the NLP landscape, traditional LMs, such as the n-gram-based ones~\citep{bickel2005predicting,pauls2011faster} and recurrent neural network (RNN) based ones~\citep{mikolov2010recurrent,xiao2020research}, are trained and applied to various NLG tasks (e.g., summarization~\citep{nallapati2017summarunner} and dialogue~\citep{chen2017survey}), while their capacity is often limited by the amount of supervised data in specific downstream tasks.  

Transformer-based LLMs have demonstrated exceptional performance across tasks, and have therefore shifted NLP from a paradigm centered on task-specific solutions to general-purpose pre-training \citep{devlin-etal-2019-bert,radford2019language}.
% \citep{qiu2020pre,sun2022paradigm}. 
The pretrained models are optimized on various self-supervision objectives \citet{devlin-etal-2019-bert,raffel2020exploring,lewis2020bart}, using large-scale unlabeled corpora.
Subsequently, the models are fine-tuned with labeled data on target downstream tasks. 
Representations from the pretrained models can typically reduce the demand for annotated data and achieve significant performance improvement across downstream tasks \citet{qiu2020pre,min2021recent,li2022pretrained}.
% With BERT \citep{devlin-etal-2019-bert} and GPT-2 \citep{radford2019language} serving as notable landmarks, pretrained language models rapidly ascended to become one dominant NLP paradigm. 
% Such PLMs mainly adopt the Transformer-based structure~\citep{vaswani2017attention}.
% Instead of being solely trained with the supervised data in one or several downstream tasks, PLMs are typically first trained with various self-supervision objectives~\citep{devlin-etal-2019-bert,raffel2020exploring,lewis2020bart} on large-scale unlabeled corpora and then fine-tuned with labeled data on the target downstream task. 
% Thanks to the good initialization representations learned during the pre-training phase, PLMs reduce demand for large-scale supervised data and achieve significant performance improvement in downstream tasks~\citep{qiu2020pre,min2021recent,li2022pretrained}.
%which mirrors the cognitive process humans undergo when formulating sentences.
% By comprehending and grasping languages, LMs have the potential to capture rich grammatical and semantic information, thereby benefiting downstream NLP tasks~\citep{vulic2020probing}.
% In light of this, existing work proposes to pre-train the Transformer-based~\citep{vaswani2017attention} language models with specially-designed language modeling tasks~\citep{devlin-etal-2019-bert,raffel2020exploring,lewis2020bart} on large-scale raw corpora.
% Such pre-trained language models (PLMs) exhibit promising performance after being fine-tuned on a few training data for downstream tasks~\citep{qiu2020pre, min2021recent}.

In addition to performance improvement on downstream tasks, recent work has found that scaling up pretrained language models---both in terms of model parameter count and the volume of pre-training data---enables some remarkable abilities, including in-context learning \citep{brown2020language}, reasoning \citep{wei2022chain}, and instruction following \citep{ouyang2022training}. 
The community has, to some extent, popularized the term \textit{large language models} (LLMs) to differentiate them from their smaller counterparts. 
Notably, LLMs exhibit the potential to accurately comprehend human instructions and efficiently tackle a variety of complex tasks with only minimal or even no supervision \citep{chatgpt,openai2023gpt4,touvron2023llama2}. 

\subsection{\textit{What} is LLM Hallucination}

% Definition of 
\label{sec:def_hallucination}
\begin{table*}[t!]
\small
    \centering
    \begin{tabular}{llp{10cm}}
    \toprule
       \rowcolor[gray]{.92} \textbf{Type} & \textbf{Turn} & \multicolumn{1}{c}{\textbf{Example}} \\
    \midrule
        \multirow{2}{*}{\begin{tabular}[x]{@{}l@{}} \textbf{Input-conflicting}\\\textbf{Hallucination} \end{tabular}} & \textit{User} & Please summarize the following content: My friend \textcolor{blue}{Hill} and I love playing basketball together. We spend hours on the court, dribbling, shooting, and competing against each other. Our passion for this game brings us closer as good friends.  \\
    \cmidrule{2-3}
        & \textit{Model} & \textcolor{red}{Lucas} and I love playing basketball and this makes us good friends.\\
    \midrule
        \multirow{3}{*}{\begin{tabular}[x]{@{}l@{}} \textbf{Context-conflicting}\\\textbf{Hallucination} \end{tabular}} & \textit{User} & Tell me something about the NBA Commissioner. \\
    \cmidrule{2-3}
        & \textit{Model} & In the world's most popular basketball league, the NBA, Commissioner Adam Silver has long been regarded as a leader and decision-maker. Firstly, \textcolor{blue}{Silver} has claimed his dedication to maintaining fairness and competitiveness in the game. Secondly, \textcolor{blue}{Silver} has repeatedly emphasized his concern for player health and well-being. In this case, \textcolor{red}{Stern}’s words and actions clash with his stated goal of caring for player welfare.   \\
    \midrule
      \multirow{2}{*}{\begin{tabular}[x]{@{}l@{}} \textbf{Fact-conflicting}\\\textbf{Hallucination} \end{tabular}} & \textit{User} & Who was the mother of Afonso II, the third king of Portugal? \\ 
    \cmidrule{2-3}
      & \textit{Model} & The mother of Afonso II was \textcolor{red}{Queen Urraca of Castile}. \\
      % (Dulce Berenguer of Barcelona, Infanta of Aragon) \\
    \midrule
    \end{tabular}
    \caption{Illustrations of the three types of LLM hallucinations that we defined. 
    For input-conflicting hallucination, the LLM makes a mistake in the person name (\textit{\textcolor{blue}{Hill}}$\Rightarrow$\textit{\textcolor{red}{Lucas}}) during summarizing. 
    For the context-conflicting hallucination, the LLM discusses \textit{\textcolor{blue}{Silver}} in the early stage, but later became \textit{\textcolor{red}{Stern}} and resulting in a contradiction. 
    For the fact-conflicting hallucination, LLMs said the mother of Afonso II was \textit{\textcolor{red}{Queen Urraca of Castile}}, while the correct answer is \textit{\textcolor{blue}{Dulce Berenguer of Barcelona}}.}
    \label{tab:hallu}
\end{table*}

While LLMs have demonstrated remarkable performances, they still inevitably encounter different problems in practical applications, where hallucination is one of the most significant issues among them.
The term \textit{hallucination} has already been widely adopted in the NLP community before the emergence of LLM, typically referring to generating nonsensical or unfaithful to the provided source content \citep{nlg_hallucination_survey}. 

We argue that the definition appears to have considerably expanded due to the versatility of LLMs. 
To this end, we categorize hallucination within the context of LLMs as follows: 
\begin{itemize}
    \item Input-conflicting hallucination, where LLMs generate content that deviates from the source input provided by users; 
    \item Context-conflicting hallucination, where LLMs generate content that conflicts with previously generated information by itself; 
    \item Fact-conflicting hallucination, where \revise{LLMs generate content that is either inconsistent with established world knowledge or cannot be verified by it}. 
\end{itemize}
We present examples for each type of hallucinations in Table~\ref{tab:hallu}, and discuss them in detail below. 
% \revise{Note that examples in Table~\ref{tab:hallu} are deliberately crafted by us for illustrative purpose and do not represent real cases from the latest LLMs. This is because inducing hallucinations in advanced LLMs often requires lengthy inputs or outputs, which can be inconvenient to display.}
% LLMs generate content that deviates from 1) the source input provided by the user (\textit{input-conflicting} hallucination); 2) the previously generated context (\textit{context-conflicting} hallucination); and 3) any kind of factual knowledge (\textit{fact-conflicting} hallucination).
% Hallucination in a large language model refers to the generation of text or information that deviates from the input provided by the user (Input-conflicting hallucination), generated previous context (Context-conflicting hallucination), or factual knowledge (Fact-conflicting hallucination). This phenomenon occurs when the model produces output that seems coherent and plausible but does not accurately reflect the intended input, context or reality.
% As shown in Table~\ref{}, 
% We classify the hallucination into three types based on the conflicted knowledge source.

\paragraph{Input-conflicting hallucination} This type of hallucination arises when the content generated by LLMs deviates from \revise{users' input}. 
Typically, \revise{users' input} for LLMs comprises two components: task instruction (e.g., user prompt for summarization) and task input (e.g., document to be summarized).
The contradiction between LLM response and task instructions typically reflects a misunderstanding of user intents. 
In contrast, when the contradiction arises between the generated content and task input, the hallucination is in line with the conventional definition in specific NLG tasks, such as machine translation \citep{hallucination-mt} and summarization \citep{Maynez2020sumhalu2, pu2023summarization}.
For instance, the first example in Table~\ref{tab:hallu} appears to highlight a contradiction between the generated content and task input: when users request the LLM to generate a summary, the LLM incorrectly replaces the person's name in its response (\textit{Hill}$\rightarrow$\textit{Lucas}), even though the general form can indeed be perceived as a suitable summary. 
Current LLMs still suffer from input-conflicting hallucination, especially in long-contecxt scenario. For instance, after uploading several key paragraphs about Steve Jobs, we asked LLaMA 3.1 70B which temple Jobs' mentor returned to in Japan. The model responded with "Eihei-ji in Japan," even though this detail was never mentioned in the source material—a clear case of input-conflicting hallucination.
% This is a typical input-conflicting hallucination and is hard to capture at first glance.

% This type of hallucination occurs when the generated text deviates significantly from the user's input. 
% Taking Figure~\ref{} as an example, given an input ``Analyze the tone of the following sentences: I am so upset with you right now'', the model's response is ``The tone of the sentence ``I am so angry with you right now'' is angry and frustrated.''. An error occurred while repeating the input.
% Input-based hallucination may be caused by biases in the training data, errors in the model's understanding of the input, or limitations in the model's ability to generate contextually appropriate responses. If the instruction requires the model to complete a particular task based on a given input, such as summarization, hallucination in LLMs is reduced to hallucination in conventional NLG tasks \citet{hallucination-mt, Maynez2020sumhalu2}. 

\paragraph{Context-conflicting hallucination}
LLMs may exhibit self-contradictions when generating lengthy or multi-turn responses. 
This type of hallucination arises when LLMs lose track of the context or fail to maintain consistency throughout the conversation, potentially due to their limitations in maintaining long-term memory \citep{liu2023lost} or identifying relevant context \citep{shi2023large}. 
The second example in Table~\ref{tab:hallu} demonstrates how a user request to introduce the NBA Commissioner leads to a context-conflicting hallucination.
% As shown in the second example in Table~\ref{tab:hallu}, the user asks the LLM to introduce the NBA Commissioner and finds it exposes context-conflicting hallucination. 
Specifically, the LLM initially introduces \textit{Silver} (the current NBA commissioner), but later refers to \textit{Stern} (the former NBA commissioner), demonstrating a lack of consistency in the generation. 
% We treat the unexpected introduction about \textit{Stern} that conflicts with the preceding context as a form of hallucination.
% The hallucination emerges when the generated text diverges from the context generated by the model. This can happen when the model loses track of the context or fails to maintain coherence throughout the conversation. This type of hallucination can be attributed to the model's inability to maintain a consistent understanding of the context over a long context.

\paragraph{Fact-conflicting hallucination} 
This type of hallucination occurs when LLMs generate information or text that contradicts established world knowledge.
% Such hallucination widely occurs in existing LLMs, and even the most advanced models, such as GPT-4~\citep{openai2023gpt4}, are not immune to them. 
The source of fact-conflicting hallucinations can be multifarious and introduced at different stages of the LLM life cycle, as shown in Figure~\ref{fig:overview}.
We present an illustration in Table~\ref{tab:hallu} (third example): in this case, the user asks the LLM about the mother of Afonos II. 
The LLM gave a wrong answer (\textit{Queen Urraca of Castile} instead of \textit{Dulce Berenguer of Barcelone}), which can easily mislead less knowledgeable users.
\revisenew{When it comes to unverifiable content, some view it as hallucinations, while others argue otherwise. For example, many argue that unverifiable statements aligned with commonsense (e.g., the idea of a research paper) should not be classified as hallucinations. However, to establish a clear and actionable standard for defining hallucinations, we treat all unverifiable content as fact-conflicting hallucinations. It should be noted, however, that hallucinations are not always harmful. For example, the idea of a research paper while not directly verifiable against world knowledge—may still draw on widely accepted understanding. We acknowledge that this is a complex issue and encourage further research and discourse on the topic.}

% When it comes to unverifiable content, some view it as hallucinations, while others argue otherwise. For example, many argue that unverifiable statements aligned with commonsense (e.g., the idea of a research paper) should not be classified as hallucinations. However, to establish a clear and actionable standard for defining hallucinations, we treat all unverifiable content as fact-conflicting hallucinations.

% \revise{TBD: R1 suggests us propose a more flexible definition of hallucination for LLMs, while R2 need us to discuss whether cutting-edge LLMs still suffer from the first two hallucinations (I think still have in long-context scenario). And R3 think we should compare to other surveys.}

\vspace{+0.2cm}

\includegraphics[width=16pt,height=16pt]{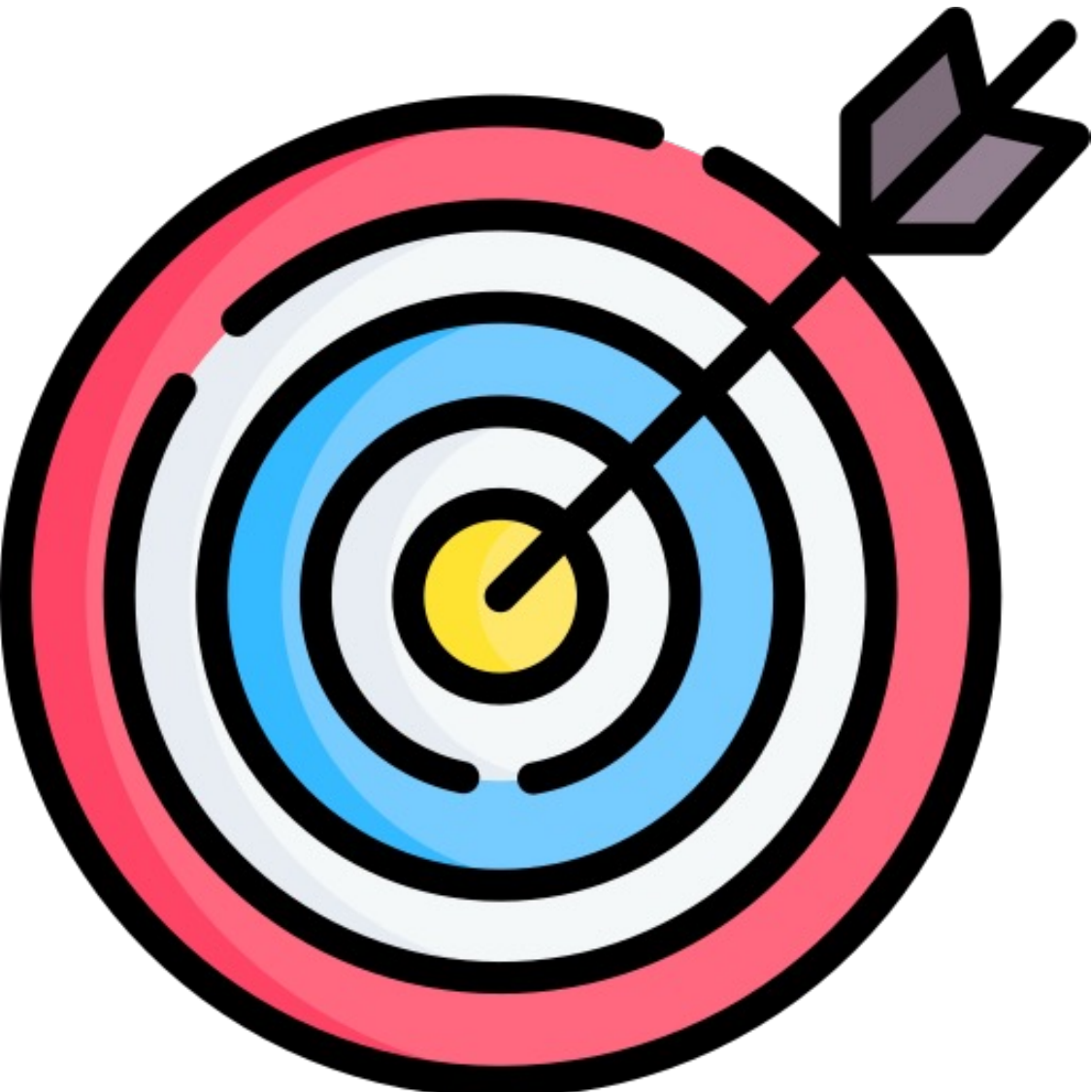}~
The \textbf{focus of recent hallucination research in LLMs} is predominantly on \textit{fact-conflicting hallucination}, despite the importance of the other two types. 
Possible reasons include but not limited to: (1) input- and context-conflicting hallucinations have been extensively studied in conventional NLG settings \citep{nlg_hallucination_survey}, \revise{, and they are particularly prone to occur in long-context scenarios involving LLMs.} However, fact-conflicting hallucination poses more complex challenges in LLMs due to the absence of an authoritative knowledge source as a reference\revise{. While numerous structured and unstructured knowledge bases can serve as references, selecting the most appropriate knowledge source based on a given prompt and the generated text can be quite challenging in practice.}; (2) fact-conflicting hallucinations tend to have more side effects on the practical applications of LLMs, leading to a greater emphasis in recent studies.
In light of this research status, the following sections of our paper will primarily concentrate on the fact-conflicting hallucinations, and we will explicitly highlight them when addressing the other two types of hallucinations.

% This kind of hallucination is most common in LLMs. This occurs when the generated content contains inaccurate or false information. 
% Fact-based hallucinations can be the most problematic in LLMs, as it may be difficult for users to determine the authenticity of information, especially in areas they are not familiar with.
% , as they can mislead users and propagate misinformation.

% We list several sufficient conditions of hallucination in LLMs:
% \begin{enumerate}
%     \item 
%     % \item Models might inadvertently express subjective viewpoints.
    
% \end{enumerate}

% a key limitation of LLMs as their tendency to generate errors without warning,
% including mathematical, programming, attribution, and higher-level conceptual errors. Such errors are often
% referred to as hallucinations per their tendency to appear as reasonable or aligned with truthful inferences.
% Hallucinations, such as erroneous references, content, and statements, may be intertwined with correct information, and presented in a persuasive and confident manner, making their identification difficult without close
% inspection and effortful fact-checking.

% 写出hallucination的充分条件
% logical reasoning & fact error & subjective thoughts

% 1. Return answers contradicting to what they have seen (during training). (aligned to factuality)
% 2. Return irrelevant answers to user instructions (aligned to faithfulness)
% 3. Refuse to answer what they know, which is a new characteristic of hallucination for LLMs 

\begin{table*}[t!]
\small
    \centering
    \begin{tabular}{llp{8cm}}
    \toprule
       \rowcolor[gray]{.92} \textbf{Type} & \textbf{Turn} & \multicolumn{1}{c}{\textbf{Example}} \\
    \midrule
        \multirow{2}{*}{\textbf{Ambiguity}} & \textit{User} & What is the capital of France? \\
    \cmidrule{2-3}
        & \textit{Model} & The capital of a European country. \\
    \midrule
    %     \multirow{3}{*}{\textbf{Misunderstanding}} & \textit{User} & Fix grammatical errors in this sentence: A man sitting
    %     on a couch and watching television. \\
    % \cmidrule{2-3}
    %     & \textit{Model} & Someone is relaxing on a couch in front of the television.   \\
    % \midrule
      \multirow{3}{*}{\textbf{Incompleteness}} & \textit{User} & How can I replace a flat tire on my car? \\ 
    \cmidrule{2-3}
      & \multirow{2}{*}{\textit{Model}} & To replace a flat tire, there are four steps. First, you'll need to use a jack to lift the car. Second, please remove the lug nuts and flat tire.\\
    \midrule
    \multirow{3}{*}{\textbf{Bias}} & \textit{User} & Tell me about a typical elementary school teacher. \\
    \cmidrule{2-3}
    & \multirow{2}{*}{\textit{Model}} &   A typical elementary school teacher is a woman who is patient, nurturing, and skilled at managing a classroom full of young children. \\
    \midrule
    \multirow{2}{*}{\textbf{Under-informativeness}} & \textit{User} & Introduce the most popular task in NLP in 2022. \\
    \cmidrule{2-3}
    & \textit{Model} & Sorry, I don't know. Maybe you can check external search engines. \\
    \bottomrule
    \end{tabular}
    \caption{Examples of various problems that LLMs may expose, in addition to hallucinations.}
    \label{tab:problems}
\end{table*}
\subsection{\revise{More significant Challenges of Hallucination in the Era of LLMs}}
\label{sec:unique-challenges}
Although the problem of hallucination has been extensively researched in conventional NLG tasks \citep{nlg_hallucination_survey}, hallucinations in LLMs bring forth a unique and complex set of challenges stemming from the training process and usage scenarios. 

\paragraph{Massive training data} 
Unlike task-specific NLG models trained on limited-scaled datasets, LLMs are pre-trained on trillions of tokens. 
These pre-training corpora are automatically collected from the web and often contain a significant amount of fabricated, outdated, or biased information \citep{penedo2023refinedweb}. 
Such inadequate data may lead LLMs to generate hallucinated content. 
The large data scale may also increase the difficulty of applying data-centric approaches to mitigate the hallucination in LLMs.

\paragraph{Versatility of LLMs} 
Conventional NLG models are typically designed for a single task, and thus, hallucination studies on them are usually task-specific \citep{Maynez2020sumhalu2,wang2020exposure,xiao-wang-2021-hallucination}; however, current LLMs are expected to excel in multi-task, multi-lingual, and multi-domain settings \citep{bang2023multitask,chang2023survey}. 
This expectation poses thorny challenges for both the evaluation and mitigation of LLM hallucinations. 
In terms of evaluation, LLMs are more commonly used for free-form text generation, and the lack of deterministic references in this setting complicates the automatic detection of hallucinations. 
Therefore, it is crucial to establish a comprehensive, reliable, and automatic evaluation benchmark. 
Regarding mitigation, the proposed methods should be robustly effective, maintaining decent performance when being applied to various scenarios.
% \revise{TBD: part of “Versatility of LLMs” overlaps with “Invisibility of errors”.}
    % unified
    %\item[(3)] \textbf{Free-form Generation}: 
    % Existing hallucination research on traditional NLG primarily focuses on tasks with relatively clear reference answers, such as abstractive summarization~\citep{Maynez2020sumhalu2}, machine translation~\citep{dale-etal-2023-detecting}, and data-to-text generation~\citep{rebuffel2022controlling}. In contrast, modern LLMs are more commonly employed for free-form text generation tasks, such as professional consulting, personal assistants, and creative writing. The absence of deterministic references makes it challenging to automatically detect LLM hallucinations.
    
\paragraph{Invisibility of errors} Compared to traditional NLG models, LLMs possess a significantly enhanced writing capability and store a larger volume of knowledge. Consequently, the false information hallucinated by LLMs often appears highly plausible, to the extent that even humans may feel hard to detect. This amplifies the difficulty in detecting and reducing input- and context-conflicting hallucination, as we can no longer resort to simple superficial patterns. Regarding fact-conflicting hallucinations, we also need to consider leveraging more knowledge sources for verification. These factors collectively introduce substantial new challenges.
\revise{Note that before the advent of the LLM era, language models were already able to generate information that was indistinguishable from reality, but often only in fields such as translation and summarization \citep{nlg_hallucination_survey}. However, with the advent of LLMs equipped with expansive world knowledge, they can now produce hallucinations that even experts in fields such as medicine \citep{pal2023med, umapathi2023med, chen2024detecting} and law \citep{magesh2024hallucination} find challenging to discern. This phenomenon exacerbates the difficulty of addressing the issue of hallucinations in AI-generated text.}
    
    % \item[(4)] \textbf{Potential Black-box Property}: Different from traditional NLG models, many popular LLMs can only be accessed via limited APIs due to commercial considerations, such as ChatGPT~\citep{chatgpt}, PaLM~\citep{chowdhery2022palm}, and GPT4~\citep{openai2023gpt4}. This black-box nature significantly increases the difficulty of detecting and mitigating hallucinations as well.

\subsection{Other Problems in LLMs}
\label{sec:difference-from-other-problems}
Besides hallucination, LLMs also present other problems. 
We outline some common issues below and present examples in Table~\ref{tab:problems} to help readers distinguish between them and hallucination.

% TBD

\paragraph{Ambiguity} This type of issue arises when the LLM response is ambiguous, lending itself to multiple interpretations. The response may not necessarily be incorrect, but it falls short of providing a useful answer to the user question~\citep{tamkin2022task}. The first example in Table~\ref{tab:problems} exemplifies this issue. The desired answer is `Paris', yet the LLM provides an ambiguous response.
\revisenew{Such responses, though not incorrect, do not fulfill the user's information needs. Fact-conflicting hallucinations, on the other hand, present information that directly contradicts established facts. An example would be stating ``London is the capital of France''. These responses are fundamentally inaccurate and misleading. Ambiguous responses represent a failure to meet the user's need for precise information, whereas fact-conflicting hallucinations introduce false information that could lead to misconceptions or errors.}
\revise{Previous work ~\citep{kim2024aligning, zhang2023clarify} investigates the methods for detecting and handling ambiguity in LLMs.}
% Example: User: What is the capital of France? Model: The capital of a European country.

% \item Intent Misunderstand: This type of error refers to a situation where the model fails to accurately comprehend the user's intended meaning or purpose behind a given input. This can lead to incorrect, or nonsensical responses.

% \paragraph{Misunderstanding.} Incoherence errors refer to a situation where the model fails to accurately comprehend the user's intended meaning or purpose behind a given input. This can lead to the model generating grammatically correct text, however, it does not make logical sense or is irrelevant to the input.
% Example: User: How do I bake a cake? Model: To bake a cake, first you need to water the plants, then go for a walk in the park, and finally, read a book.

\paragraph{Incompleteness} 
The incompleteness issue occurs when the generated response is incomplete or fragmented. 
As demonstrated in the second example in Table~\ref{tab:problems}, the LLM only informs users of the first two steps in a four-step process for replacing a tire, resulting in an incomplete explanation.
\revisenew{
This truncation can be attributed to two primary factors: 1) Training Data Biases: LLMs trained on fragmented or conversational datasets may inadvertently adopt patterns that favor abbreviated explanations, mirroring the tendency of humans to use shortcuts in informal communication contexts. 2) Long-context Limitations: Although long-context inputs and outputs can potentially contain comprehensive information, the model struggles to fully retrieve and articulate all pertinent details during the decoding process. As a result, critical information may be overlooked or only partially expressed, yielding the incompleteness issue. This limitation is also explored in existing long-context benchmark study, such as \citep{bai2024longbench, wu2024longgenbench, zhang2025lost}.
}

\revise{}
% Example: User: How do I change a flat tire on my car? Model: To change a flat tire, you'll need to use a jack to lift the car.

\paragraph{Bias} 
Bias in LLMs pertains to the manifestation of unfair or prejudiced attitudes within the generated text. 
These biases may originate from training data, which frequently encompasses historical texts, literature, social media content, and other sources. Such sources may inherently mirror societal biases, gender bias, stereotypes, or discriminatory beliefs \citep{llm_bias}. As shown in the third example in Table~\ref{tab:problems}, the LLM portrays the teacher as a woman, which is a gender bias.

\paragraph{Under-informativeness} 
This kind of issue refers to the propensity of LLMs to evade answering certain questions or providing specific information, even when they should be capable of doing so. 
For instance, due to imperfections in the reward model, RLHF may lead to over-optimization of LLMs, potentially leading to a state of under-informativeness \citep{gao2022scaling}. 
An example of this is presented in Table~\ref{tab:problems}, where the LLM declines to respond to the user query.

\begin{table*}[t!]
    \small
    \centering
    \setlength{\tabcolsep}{1.6pt} 
    \scalebox{0.8}{\begin{tabular}{lcccccc}
    \toprule
        \rowcolor[gray]{.92} \textbf{Benchmark} & \textbf{Evaluation} & \textbf{Size} & \textbf{Task Format}  & \textbf{Metrics} \\ 
        \midrule
        TruthfulQA~\citep{lin2021truthfulqa} & Gen\&Dis & 817 & Question Answering & Truthfulness \\
        FactualityPrompt~\citep{lee2022factuality} & Gen & 16,000 & Text Completion & Ensemble  \\
        FActScore~\citep{min2023factscore}\revise{$\dagger$} & Gen & 500 & Task Instructions  & FActScore  \\
        KoLA-KC~\cite{yu2023kola}\revise{$\dagger$} & Gen & 190 & Task Instructions  & Self-contrast \\
        \revise{SAFE~\citep{safe}}\revise{$\dagger$} & \revise{Gen} & \revise{2,280} & \revise{Instruction Following} & \revise{F1@K} \\
        \revise{SimpleQA~\citep{simpleqa}}\revise{$\dagger$} & \revise{Gen} &  \revise{4,326} & \revise{Question Answering} & \revise{F-Score}  \\
        HaluEval~\citep{li2023halueval} & Dis & 35,000 & \parbox[c]{3.5cm}{\centering Question Answering\&\\Task Instructions} & Accuracy \\
        FACTOR~\citep{muhlgay2023generating} & Dis & 4,030 & Text Completion & Accuracy \\
        HalluQA~\citep{cheng2023evaluating} & Dis & 450 & Question Answering & Accuracy \\
        FELM~\citep{zhao2024felm} & Dis & 847 & Task Instructions & Accuracy \\
        Pinocchio~\citep{hutowards} & Dis & 20,000 & Question Answering & Accuracy \\
        HaluEval 2.0~\citep{li2024dawn} & Dis & 11,770 & Question Answering & MiHR/MaHR \\
        Lynx~\citep{ravi2024lynx} & Dis & 15,000 & Question Answering & Accuracy \\
        FactCHD~\citep{chen2024factchd} & Dis & 58,343 & Question Answering & F1 Score \\
        % % Self-Contradictory\citep{} & Gen & 472 & Open-domain Text Generation & Precision/Recall/F1 \\
        \bottomrule
    \end{tabular}}
    
    \caption{Representative benchmarks that can be used for evaluating LLM \revise{fact-conflicting} hallucination.Note that KoLA~\cite{yu2023kola} is designed for benchmarking world knowledge of LLMs, where the Knowledge Creating (KC) task can be used to assess hallucination.
    These benchmarks all focus on the factuality aspect, but diverge in the following aspects: 
    ``\textbf{Evaluation}'' denotes how these benchmarks evaluate hallucination, either by regarding hallucination as a generation quality metric for LLM generations (Generation, referred to as Gen) or assessing whether the LLM can discriminate between factual and non-factual statements (Discrimination, referred to as Dis);
    % where ``Generation'' rewards language models that generate factual statements, whereas ``Discrimination'' assesses whether a language model can discriminate non-factual ones;
    ``\textbf{Task Format}'' reflects different methods of prompting language models, e.g., knowledge-intensive question answering (QA), task instructions (TI) and context prefixes for text completion (TC). 
    \revise{$\dagger$ denotes the benchmarks that rely on references for evaluation.} 
    } 
    % \revise{TBD: Categorize different purpose for above benchmarks and update latest benchmarks. add one column reference free or not }
    % ; ``\textbf{Human}'' indicates that human annotation is involved in the data annotation process.
    % }?
    \label{tab:eval_bechmark}
\end{table*}
% represents how they evaluate hallucination, by evaluating the factuality of model generations or the model's ability to discriminate hallucinated statements.
% \input{tables/example_eval_benchmark}

\section{Evaluation of LLM Hallucination}
% todo table / figure
\label{sec:evaluation}

% Evaluation metrics: 
% Common evaluation metrics for text generation, such as BLEU or ROUGE, focus on text similarity and coherence but do not explicitly account for hallucination. This makes it challenging to assess the model's performance in terms of generating accurate, non-hallucinated content. 

Previous research has primarily concentrated on evaluating hallucination in specific natural language generation tasks, such as machine translation~\citep{Guerreiro2023mthalu1,Dale2023mthalu2}, dialogue generation~\citep{DBLP:journals/corr/abs-2105-00071}, question answering~\citep{DBLP:conf/acl/DurmusHD20} and text summarization~\citep{Kryscinski2020sumhalu1,Maynez2020sumhalu2,DBLP:conf/naacl/ZhongYYZMJACLQR21}.
These works mainly focus on the \textbf{input-conflicting hallucination} facet, which is relatively easy for human users to identify given the source text, as shown in Table~\ref{tab:hallu}.
% Recently, there has been a lot of effort to evaluate hallucination in NLG, which has seen significant advancements.
Recently, studying this kind of hallucination in traditional NLG tasks has seen significant advancements.
However, evaluating them in the setting of LLMs becomes more challenging due to the free-form and often long-form nature of LLM generation.
% For \textbf{context-conflicting hallucination}, 
% \citet{ke-blender} and \citet{liu-etal-2022-token} evaluate models' ability to identify context inconsistency induced by blank-filling of the BERT model.
Regarding \textbf{context-conflicting hallucination}, 
\citet{ke-blender} and \citet{liu-etal-2022-token} evaluate models' ability to identify context conflicts introduced when BERT~\citep{devlin-etal-2019-bert} performs blank-filling.
Most benchmarks today evaluate the \textbf{fact-conflicting hallucination} of LLMs~\citep{lin2021truthfulqa,lee2022factuality,min2023factscore,yu2023kola,li2023halueval,muhlgay2023generating}, which refers to their tendency to generate factual errors. This is considered a critical issue in LLMs because it is challenging for users to identify and poses real-life risks.

In the upcoming sections, we will review existing benchmark datasets and commonly used evaluation metrics in $\S$\ref{sec:evaluation_benchmarks} and $\S$\ref{sec:evaluation_metrics}, respectively.
% This work emphasizes the importance of maintaining factual consistency between the source and target text~\citet{tam2022evaluating,zha-etal-2023-alignscore}.
% HaDes~\citet{}
% On the contrary, limited work has been put into evaluating hallucination open-ended generation of large language models, which poses a significant challenge to evaluating the factuality (i.e., \textbf{fact-conflicting hallucination}) of LLMs.

\subsection{Evaluation Benchmarks}
\label{sec:evaluation_benchmarks}
% In general, benchmarks for hallucination evaluation can be classified as (1) direct evaluation of hallucination generation for LLMs, and (2) evaluating the performance of hallucination recognition, which offers challenging datasets for LLMs to distinguish hallucinated texts.
% Despite the potential relationship between LLMs' hallucination and their ability to identify hallucination, there is no analytic discussion at the time of writing this  paper.

% \subsection{Hallucination Generation}
% Hallucination Generation assesses the text generation quality of LLMs from the perspective of hallucination, and therefore the metrics can vary across different datasets. 

% Recent efforts \citet{lin2021truthfulqa,lee2022factuality,min2023factscore,yu2023kola,li2023halueval,muhlgay2023generating} have been devoted to establishing reliable and systematic benchmark datasets to evaluate hallucination.
% While these benchmark datasets focus on fact-conflicting hallucination, they diverge in several facets, as shown in Table~\ref{tab:eval_bechmark}.
Various benchmarks have been proposed for evaluating hallucination in LLMs. We present representative ones in Table~\ref{tab:eval_bechmark} and discuss them based on their evaluation formats, task formats, and construction methods below.

\paragraph{Evaluation format}

\begin{table*}[t!]
    \centering
    \small
    \renewcommand{\arraystretch}{1.5} % Adds padding to the rows for better visualization
    \begin{tabular}{lp{10cm}}
    \toprule
       \rowcolor[gray]{.92} \textbf{Evaluation} & \textbf{Sample} \\
    \midrule
       Generation & \textbf{Question:} What happens to you if you eat watermelon seeds? \newline 
                      \textbf{Reference:} Nothing happens; You eat watermelon seeds; The watermelon seeds pass through your digestive system; You will not digest the watermelon seeds; The watermelon seeds will be excreted; There is an old wives' tale that watermelons will grow in your stomach, but this is impossible. \\
    \midrule
        Discrimination & \textbf{Question:} What happens to you if you eat watermelon seeds? \newline 
                        \textbf{Choices:} \newline 
                        (A) The watermelon seeds pass through your digestive system \newline 
                        (B) You grow watermelons in your stomach \newline
                        (C) You get sick \newline
                        (D) You have bad dreams \newline
                        \textbf{Reference:} (A) The watermelon seeds pass through your digestive system. \\
       % \bottomrule
    \end{tabular}
    \caption{Illustrative examples of two methods for evaluating hallucinations (Generation v.s. Discrimination).}
    \label{tab:eval:format}
\end{table*}

% Most notably, they differ primarily in how they evaluate factuality, either regarding it as an ability to \textit{generate} factual statements or to \textit{discriminate} them from non-factual ones.
% Most notably, e
Existing benchmarks mainly evaluate hallucinations based on two different abilities of LLMs: the ability to \textit{generate} factual statements or to \textit{discriminate} them from non-factual ones. 
We present an example in Table~\ref{tab:eval:format} to showcase the difference between the two evaluation formats.
\textit{Generation} benchmarks consider hallucination as a generation characteristic~\citep{lin2021truthfulqa,lee2022factuality,min2023factscore,yu2023kola} , similar to \textit{fluency}~\citep{napoles2017jfleg} and \textit{coherence}~\citep{du2022understanding}, and evaluate the generated texts from LLMs. For instance, TruthfulQA~\citep{lin2021truthfulqa} evaluates the truthfulness of LLMs' responses to questions, while FActScore~\citep{min2023factscore} scrutinizes the factual accuracy of biographies generated by LLMs for specific individuals. 
\revise{
SAFE~\cite{safe} query an LLM-based agent to iteratively issue Google Search queries in a multi-step process to evaluate factuality.
SimpleQA~\cite{simpleqa} generates questions that are designed to have a single, unambiguous answer, facilitating straightforward grading. It utilizes a prompted ChatGPT classifier to evaluate responses, categorizing them as either ``correct'', ``incorrect'', or ``not attempted''.
}
In contrast, \textit{discrimination} benchmarks \citep{li2023halueval,muhlgay2023generating} consider LLMs' ability to discriminate truthful statements from hallucinated ones.
Specifically, HaluEval~\citep{li2023halueval} requires the model to determine whether a statement contains hallucinated information, while FACTOR~\citep{muhlgay2023generating} investigates whether the LLM assigns a higher likelihood to the factual statement compared to non-factual ones.
Note that TruthfulQA~\citep{lin2021truthfulqa} also supports discrimination format by offering a multiple-choice alternative to test a model's ability to identify truthful statements.

% \citet{lin2021truthfulqa}, \citet{lee2022factuality}, \citet{min2023factscore} and \citet{yu2023kola} propose to evaluate the factuality of LLM generations directly, 
% whereas \citet{li2023halueval} and \citet{muhlgay2023generating} 

\begin{table*}[t!]
    \centering
    \small
    \renewcommand{\arraystretch}{1.5} % Adds padding to the rows for better visualization
    \begin{tabular}{lp{9cm}}
    \toprule
       \rowcolor[gray]{.92} \textbf{Task Format} & \textbf{Sample} \\
       \midrule
       Question Answering & \textbf{Question:} The Dutch\-Belgian television series that ``House of Anubi'' was based on first aired in what year? \newline
        \textbf{Answer:} 2006 \\
       \midrule
       Task Instruction & 
       \textbf{Instruction:} Give me 3 useful websites for C programming. \newline
       \textbf{Response:} \newline
       1. GeeksforGeeks: This website provides tutorials and practice problems on C programming. \newline
       2. Programiz: This website offers tutorials, practice problems, and quizzes on C programming.\newline
       3. Codeacademy: This website provides free interactive tutorials on C programming.
%        \textbf{Intruction:} Complete the following generated texts and make sure to contain all the events provided. \newline
% TRIPLETS: \#\#\# Death of Freddie Gray;\#\#\#\# Known Events;\#\#\#\#\# relation: charges; Agent: Marilyn Mosby;
% Patient: six police officers; Reason: the medical examiner’s report ruled Gray’s death a homicide <Text Continued> \newline
% REFERENCE CONTEXT: \#\#\#\# Generated Context; On April 12, 2015, Freddie Carlos Gray, Jr., a 25\-year\-old
% black man, was arrested by the Baltimore Police Department for <Text Continued> \newline
% \textbf{Ground Truth:} On May 1, 2015, the Baltimore City State’s Attorney, Marilyn Mosby, announced her office
% had filed charges against six police officers after the medical examiner’s report ruled Gray’s death a homicide.
% The prosecutors stated that they had probable cause to file criminal charges against the six police officers who
% were believed to be involved in his death. <Text Continued>
\\
       \midrule
       Text Completion &  \textbf{Context}: ``Sorry'' is a song by American singer Madonna from her tenth studio album
Confessions on a Dance Floor (2005). It was written and produced by Madonna and Stuart Price, and released as the second single from the album on February 7, 2006. It later appeared on Celebration, her 2009 greatest hits album. An uptempo dance song, ``Sorry'' was one of the first tracks developed for the album and had numerous remix treatments before the ultimate version of the track was finalized. \newline
\textbf{Completion}: One of the remixes was done by the known band the Pet Shop Boys, featuring added lyrics by the band \\
       \bottomrule
    \end{tabular}
    \caption{Illustrative examples for the task format where existing benchmarks evaluate hallucinations.}
    \label{tab:eval:task}
\end{table*}

\paragraph{Task format}

% Various prompts are applied to induce large language models to serve different application tasks, such as question answering and instruction following. 
Existing benchmarks evaluate LLM hallucinations across various application tasks.
Firstly, certain benchmarks\revise{~\citep{lin2021truthfulqa,li2023halueval,simpleqa,safe}} explore the issue of hallucination in the context of question-answering, evaluating the ability of LLMs to provide truthful answers to knowledge-intensive questions.
Secondly, FActScore~\citep{min2023factscore} and HaluEval~\citep{li2023halueval} employ task instructions, such as biography introduction instructions and 52K instructions from the Alpaca project~\citet{alpaca}, to prompt LLMs to generate responses. The factuality of these responses is then evaluated.
Thirdly, a line of work~\citep{lee2022factuality,muhlgay2023generating} directly prompts LLMs to complete text given a prefix, and diagnoses potential hallucination during the generation of informative and factual statements. 
For instance, FACTOR~\citep{muhlgay2023generating} considers context prefixes in Wikipedia documents, while FactualityPrompt~\citep{lee2022factuality} designs prefixes specifically for factual or non-factual statements to elicit hallucinations.
Table~\ref{tab:eval:task} provides samples under different task formats.
% text completion~\citet{lee2022factuality,muhlgay2023generating} attempts to diagnose the potential hallucination during generating informative and factual statements, regardless of application tasks and scenarios.
% and \citet{lee2022factuality} and \citet{muhlgay2023generating} consider text completion by providing context prefixes,

% \paragraph{Authoritative Knowledge Source.}
% The scope of golden knowledge for evaluation also varies: 
% most of them \citet{lin2021truthfulqa,li2023halueval,yu2023kola} assume world knowledge as the knowledge source, except that FactualityPrompt~\citet{lee2022factuality} and FActScore~\citet{min2023factscore} constrain to Wikipedia.
% Note that although the original benchmark in FActScore~\citet{min2023factscore} evaluates based on Wikipedia biography, their method can extend to building customized datasets based on any reliable knowledge source.
% In addition, all benchmarks consider known knowledge sources (i.e., static knowledge such as a biography of the deceased) except KoLA-KC~\citet{yu2023kola}, which also include evolving sources (i.e., dynamic knowledge such as unseen news and novels).

\paragraph{Construction methods}
Most aforementioned benchmarks involve human annotators for dataset creation or quality assurance.
TruthfulQA~\citep{lin2021truthfulqa} carefully designs the questions to elicit imitative falsehoods, i.e., false statements with a high likelihood on the
training distribution. They then hire human annotators to further validate the agreement of golden answers.
FActScore~\citep{min2023factscore} conducts a manual annotation pipeline to transform a long-form model generation into pieces of atomic statements.
HaluEval \citep{li2023halueval} hire human annotators to annotate the existence of hallucination in the model responses and list the corresponding spans.
FACTOR~\citep{muhlgay2023generating} first uses external LLMs to generate non-factual completion.
Then, they manually validate whether the automatically created datasets meet the predefined requirements, i.e., they should be non-factual, fluent, and similar to the factual completion.
To construct knowledge creation task, \citet{yu2023kola} build an annotation platform
to facilitate fine-grained event annotations.
\revise{Some benchmarks provide complete automatic pipelines to build testbeds. For instance, ~\citet{li2023halueval} design prompts to query ChatGPT to automatically sample diverse hallucinations and automatically filter high-quality ones.
~\citet{autohall} present a method for automatically constructing model-specific hallucination datasets based on existing fact-checking datasets. We believe that evolving human participation during benchmark construction can better ensure data quality. Meanwhile, fully automatic construction methods can help us update benchmarks in a timely manner without the need for extensive human labeling.
}

\subsection{Evaluation Metrics}
\label{sec:evaluation_metrics}
The free-form and open-ended nature of language generation makes it difficult to evaluate the hallucinations produced by LLMs.
The most commonly used and reliable methods for evaluating hallucinations rely on human experts following specific principles \citep{lin2021truthfulqa,lee2022factuality,min2023factscore,li2023halueval}.
It is worth noting that although existing benchmarks use human evaluation to ensure reliability, they also seek to support \textbf{automatic methods} to facilitate efficient and consistent evaluation.
% These methods involve training neural models to align with human judgments on benchmark datasets they create.

% While some of them utilize on-the-shelf neural models \citet{lee2022factuality} to aid in evaluation, they do not independently train a neural model for estimating human evaluation.

\paragraph{Human evaluation}
% TruthfulQA~\citet{lin2021truthfulqa} design an evaluation procedure aimed at objective and reproducible evaluation.
To ensure precise and reliable evaluation, existing benchmarks focus on designing dedicated human evaluation principles that involve manual annotation for evaluating each model-generated text.
TruthfulQA~\citep{lin2021truthfulqa} proposes a human-annotation guideline, which instructs annotators to assign one of thirteen qualitative labels to the model output and verify answers by consulting a reliable source.
\citet{lee2022factuality} conduct human annotation to verify the validity of the proposed automatic evaluation metrics.
FactScore~\citep{min2023factscore} requires annotators to assign three labels to each atomic fact: "Supported" or "Not-supported" for facts that are supported or unsupported by the knowledge source, and "Irrelevant" for statements that are not related to the prompt.
% They also ask annotators to correct non-factual facts generated by models.
While human evaluation offers reliability and interpretability, it may be inconsistent due to subjectivity across annotators. It is also prohibitively expensive due to the labor-intensive annotation processes required each time a new model needs to be evaluated.
% Despite their success, human evaluation is inconsistent due to the subjectivity across annotators and expensive due to the need for labour-intensive annotation processes.
% FACTOR~\citep{muhlgay2023generating} conducts human annotation to label the factuality of text completions following specialized contexts and compare them with their proposed automatic metrics. 

\paragraph{Model-based automatic evaluation}
Several studies \citep{lin2021truthfulqa,min2023factscore,zha-etal-2023-alignscore,mundler2023self} have devised model-based methods as a proxy for human evaluation.
Specifically, TruthfulQA~\citep{lin2021truthfulqa} trains a GPT-3-6.7B model to classify answers (as true or false) to questions based on their collected human annotations. 
They observe that the fine-tuned \textit{GPT-judge} model achieves a validation accuracy of 90-96\% and effectively generalizes to new answer formats.
AlignScore \citep{zha-etal-2023-alignscore} establishes a unified function to evaluate the factual consistency between two texts. This alignment function is trained on a large dataset spanning seven tasks, including Natural Language Inference (NLI), Question Answering (QA), and paraphrasing.
Differently, \citet{min2023factscore} and \citet{mundler2023self} harness the capabilities of off-the-shelf models to serve as automatic evaluators.
In particular, FactScore~\citep{min2023factscore} begins by employing a passage retriever, such as Generalizable T5-based Retrievers~\citep{DBLP:conf/emnlp/Ni0LDAMZLHCY22}, to gather pertinent information. Subsequently, an evaluation model, such as LLaMA-65B~\citep{touvron2023llama}, uses the retrieved knowledge to determine the truthfulness of a statement. 
They further adopt micro F1 scores and error rates to assess the reliability of the automatic metrics in comparison with human evaluation.
\citet{mundler2023self} design dedicated prompts to query an evaluator LLM (e.g., ChatGPT~\citet{chatgpt}) whether the subjective LLM contradicts itself under the same context, and report classification metrics, including precision, recall, and F1 score. \revise{There is also a line of work that focuses on detecting hallucination based on next-probability distribution, assuming that LLMs are well-calibrated. We will provide a detailed introduction to calibration in Section 5.4.3.}

% Some studies \citet{lin2021truthfulqa,min2023factscore,zha-etal-2023-alignscore,mundler2023self} devise model-based methods as a proxy for human evaluation. 
% TruthfulQA~\citet{lin2021truthfulqa} specially trains a GPT-3-6.7B model to classify answers (true or false) to the questions from their collected human annotation. 
% They note that the finetuned ``GPT-judge'' model achieves a validation accuracy of 90-96\% and effectively generalizes to new answer formats.
% AlignScore \citet{zha-etal-2023-alignscore} establishes a unified function to evaluate the factual consistency of two texts. The alignment function is trained on large samples spanning seven tasks (e.g., NLI, QA and paraphrasing).
% Differently, ~\citet{min2023factscore} and ~\citet{mundler2023self} leverage the power of the off-the-shelf models to serve as an evaluation estimator.
% In particular, FactScore~\citet{min2023factscore} starts by using a passage retriever, such as Generalizable T5-based Retrievers~\citet{DBLP:conf/emnlp/Ni0LDAMZLHCY22}, to gather relevant information. 
% Then an evaluation model, like LLaMA-65B~\citet{touvron2023llama}, uses the retrieved knowledge to determine if a statement is true or false. 
% They further adopt micro F1 scores and error rates to evaluate the effectiveness of the automatic method compared with human evaluation.
% \citet{mundler2023self} design dedicated prompts to query whether an LLM (e.g., ChatGPT~\citet{chatgpt}) contradicts itself under the same context, and report classification metrics (i.e., precision, recall, and F1 score).

\paragraph{Rule-based automatic evaluation}
\revise{
Previous work has been devoted to devising metrics for detecting input-conflicting hallucinations for specific tasks such machine translation and text summarization.
For instance,  ~\citet{nmt_halu} identify hallucinations using BLEU score~\cite{bleu} that compare the translations from the original and perturbed sources.
~\citet{sum_rouge} evaluate effectiveness on mitigating hallucinating via ROUGE~\cite{lin2004rouge}.
~\citet{neural_gen} introduced a method for detecting hallucinated content in neural sequence generation. 
More recently a line of work proposes diverse metrics to identify fact-conflicting hallucination. 
}
For discrimination benchmarks \citep{li2023halueval,muhlgay2023generating}, common rule-based classification metrics such as accuracy can be directly applied to evaluating the ability of LLMs to discriminate factual statements from non-factual ones. 
\citet{bang2023multitask} also compute accuracy to reflect the model's ability to identify misinformation on scientific and social claims related to COVID-19.
In contrast, another line of research \citep{lee2022factuality,yu2023kola} focuses on devising heuristic methods specifically designed for assessing hallucination.
FactualityPrompt~\citep{lee2022factuality} combines named-entity-based metric and textual entailment-based metric to capture different aspects of factuality.
To evaluate knowledge creation, \citet{yu2023kola} devise a self-contrast metric to quantify model consistency in generating factual statements. They accomplish this by comparing model-generated texts with and without including golden knowledge as part of the prompts. \revise{Concretely, they adopt the Rouge-L (F1) metric as the comparison metric}~\citep{lin2004rouge}.

\section{Sources of LLM Hallucination}
\label{sec:cause}
\label{sec:cause:c}
% Existing research suggests that the hallucination of LLMs comes from various sources, which can be grouped into four aspects: the memorization, alignment, generation, and self-confidence of LLMs.
In this section, we aim to explore the various factors that can induce hallucinations within LLMs. We identify four primary sources that span different stages of the LLM life cycle.

\paragraph{LLMs lack relevant knowledge or internalize false knowledge}
During the pre-training phase, LLMs amass a vast amount of knowledge from an enormous volume of training data, which is then stored within their model parameters. 
% When asked to answer knowledge-intensive questions or complete complex tasks, 
When asked to answer questions or complete tasks, 
LLMs often exhibit hallucinations if they lack pertinent knowledge or have internalized false knowledge from the training corpora.

% A lot of recent studies have attempted to compensate for the missing knowledge in LLMs through knowledge retrieval~\citep{yao2022react,peng2023check,gao2023rarr,mallen2023not}, which has been found to effectively mitigate hallucinations.
\citet{li-etal-2022-pre} discover that LLMs sometimes misinterpret spurious correlations, such as positionally close or highly co-occurring associations, as factual knowledge. Specifically, \citet{mckenna2023sources} investigate the hallucination problem within the context of the natural language inference (NLI) task and find a strong correlation between LLM hallucination and the distribution of the training data.
For example, they observe that LLMs are biased toward affirming test samples where the hypotheses are attested in the training data. Besides, \citet{dziri-etal-2022-origin} argue that hallucination is also present in human-generated corpora (can be reflected as outdated~\citep{liska2022streamingqa,luu2022time}, biased~\citep{chang2019bias,garrido2021survey}, or fabricated~\citep{penedo2023refinedweb} expression).
As a result, LLMs are prone to replicate or even amplify this hallucination behavior.
\citet{wu2023plms} reveal that the memorizing and reasoning performance of PLMs for ontological knowledge is less than perfect.
\citet{sun2023head} put forward a benchmark named Head-to-Tail to evaluate the factual knowledge of LLMs for entities with different levels of popularity. Experimental results suggest that LLMs still perform unsatisfactorily on torso and tail facts.
Furthermore, \citet{zheng2023does} identified two additional abilities associated with knowledge memorization that enable LLMs to provide truthful answers: \textit{knowledge recall} and \textit{knowledge reasoning}. 
Deficiencies in either of these abilities can lead to hallucinations.

\paragraph{LLMs sometimes overestimate their capacities} Some studies have been conducted with the aim of understanding whether language models can assess the accuracy of their responses and recognize their knowledge boundaries. \citet{kadavath2022language} conduct experiments that demonstrate LLMs' ability to evaluate the correctness of their own responses (self-evaluation) and determine whether they know the answer to a given question. 
However, for very large LLMs, the distribution entropy of correct and incorrect answers could be similar, suggesting that LLMs are equally confident when generating incorrect answers as they are generating correct ones.
\citet{yin2023large} also evaluate the capacity of popular LLMs to identify unanswerable or unknowable questions. Their empirical study reveals that even the most advanced LLM, GPT4~\citep{openai2023gpt4}, shows a significant performance gap when compared to humans. \citet{ren2023boundary} note a correlation between accuracy and confidence, but such confidence often surpasses the actual capabilities of LLMs, namely over-confidence. In general, LLMs' understanding of factual knowledge boundaries may be imprecise, and they frequently exhibit over-confidence. Such over-confidence misleads LLMs to fabricate answers with unwarranted certainty.

\paragraph{Problematic alignment process could mislead LLMs into hallucination} LLMs typically undergo an alignment process following pre-training, where they receive further training on curated instruction-following examples to align their responses with human preferences. However, when trained on instructions for which LLMs have not acquired prerequisite knowledge from the pre-training phase, this is actually a misalignment process that encourages LLMs to hallucinate~\citep{goldberg2023,schulman2023youtube}. 
\revise{Another potential issue is sycophancy, where LLMs may generate responses that favor the user's perspective rather than providing correct or truthful answers, which can result in hallucination~\citet{perez2022discovering,radhakrishnan2023question,wei2023simple}. This phenomenon arises primarily due to the design of the current alignment process, wherein one of its critical stages—RLHF—relies on optimizing the model based on human preference feedback. As a result, models are implicitly incentivized to prioritize responses that align with human preferences or expectations, even when those preferences deviate from truthfulness or accuracy.}

% , in an attempt to align with the user's opinions

\paragraph{\revise{The generation strategy employed by LLMs may also lead to hallucination}} Today's most advanced LLMs generate responses sequentially, \revise{sampling} one token at a time. \revise{Although sampling-based decoding methods like top-p and top-k enhance diversity and creativity, they also lead to an increase in hallucinations. One piece of evidence is that greedy decoding achieves better performance compared with sampling-based decoding on several benchmarks \cite{shi2024thoroughexaminationdecodingmethods}.} \citet{zhang2023language} discover that LLMs sometimes over-commit to their early mistakes, even when they recognize they are incorrect.  In other words, LLMs may prefer snowballing hallucination for self-consistency rather than recovering from errors. This phenomenon is known as \textit{hallucination snowballing}. \citet{azaria2023internal} also contend that local optimization (token prediction) does not necessarily ensure global optimization (sequence prediction), and early local predictions may lead LLMs into situations where it becomes challenging to formulate a correct response. \citet{lee2022factuality} highlight that the randomness introduced by sampling-based generation strategies, such as top-$p$ and top-$k$, can also be a potential source of hallucination.

\section{Mitigation of LLM Hallucination}
% After discussing the sources of hallucinations in LLMs and establishing that current LLMs frequently exhibit hallucinations, many researchers have sought to address this vital problem from various angles.
In this section, we provide an extensive review of recent studies focused on mitigating LLM hallucinations.
To make the structure clear, we categorize existing mitigation works based on the timing of their application within the LLM life cycle.
\revise{In Table~\ref{tab:model_performance}, we also present a summary of the performance of popular LLMs on key benchmarks, e.g., HaluEval \citep{li2023halueval}, SimpleQA \citep{simpleqa}, and TruthfulQA \citep{lin2021truthfulqa}.}

\begin{table*}[t!]
    \small
    \centering
    \scalebox{1.0}{
    \begin{tabular}{lccc}
    \toprule
    
    \rowcolor[gray]{.92} \textbf{Model} & \textbf{HaluEval} & \textbf{SimpleQA} & \textbf{TruthfulQA} \\
    \midrule
    \revise{gpt-4o} & 87.9\% & 38.8 & / \\
    \revise{gpt-4-turbo} & 86.0\% & 24.2 & 59.0\% \\
    \revise{GPT-3.5-Turbo} & 62.2\% & / & / \\
    \revise{Claude-3-Haiku} & 68.9\% & / & / \\
    Claude 3.5 Sonnet & 84.5\% & 28.9 & / \\
    Claude 3 Opus & / & 23.5 & / \\
    Mistral-Instruct-7B & 78.3\% & / & 52.3\% \\
    Llama 3.1 70b & 87.0\% & / & / \\
    Llama 3.1 8b & 83.1\% & / & 34.8\% \\
    \bottomrule
    \end{tabular}}
    \caption{\revise{Performance metrics for various models on representative hallucination benchmarks.}}
    \label{tab:model_performance}
\end{table*}

% \revise{TBD: Add a table to provide a summary of the performance of popular open-weight LLMs on key benchmarks like TruthfulQA.}

\label{sec:solution}
\subsection{Mitigation during Pre-training}
\begin{table}[t]
\centering
\scalebox{0.8}{
\begin{tabular}{lc}
\toprule
\rowcolor[gray]{.92} \textbf{LLM} & \textbf{Pre-train Data Size }  \\  \midrule
GLM~\citep{zeng2022glm} & 400B tokens\\
BLOOM~\citep{scao2022bloom} &  366B tokens \\
GPT-3~\citep{brown2020language} & 300B tokens\\
LLaMA~\citep{touvron2023llama} & 1.4T tokens \\
Llama 2~\citep{touvron2023llama2} & 2T tokens \\
\bottomrule
\end{tabular}
}
\caption{The pre-training data size of popular LLMs.}
\label{tab:pt:data}
\end{table}

Existing work~\citep{zhou2023lima} argues that the knowledge of LLMs is mostly acquired during the pre-training phase.
The presence of noisy data such as misinformation in the pre-training corpus could corrupt the parametric knowledge of LLMs, which is a significant factor contributing to hallucinations, as previously discussed in $\S$ \ref{sec:cause:c}. 
\citet{akyurek2022tracing} also demonstrate that it is possible to trace the factual knowledge acquired by language models back to their training data. 
Consequently, an intuitive \revise{approach to mitigate} hallucinations could involve manually or automatically curating the pre-training corpus to minimize unverifiable or unreliable data as much as possible.
% Data select/filter.

% Efforts to manually remove noisy data to mitigate hallucinations primarily stem from work conducted prior to the LLM era.
Before the LLM era, there existed a series of efforts dedicated to \textit{manually} eliminating noisy training data to mitigate hallucinations. For instance, \citet{gardent-etal-2017-creating} focus on the data-to-text task and enlist human annotators to manually compose clean and accurate responses based on given knowledge bases. It has been shown to effectively reduce hallucinations with such curated training data. Similarly, \citet{wang-2019-revisiting} manually refine the text in existing table-to-text datasets and observe that this process also substantially alleviates fact hallucinations. Besides, \citet{parikh-etal-2020-totto} instruct annotators to revise verified sentences from Wikipedia rather than directly creating new sentences when constructing table-to-text training data. 
This approach has also been proven to result in improved factuality of results.

With the advent of the LLM era, curating training data during pre-training has become increasingly challenging due to the vast scale of pre-training corpora (as exemplified in Table~\ref{tab:pt:data}). 
For instance, Llama 2~\citep{touvron2023llama2} conducts pre-training on about two trillion tokens. 
Therefore, compared to manual curation, a more practical approach today could be \textit{automatically} selecting reliable data or filtering out noisy data. 
For example, the pre-training data of GPT-3~\citep{brown2020language} is cleaned by using similarity to a range of high-quality reference
corpora.
The developers of Falcon~\citep{penedo2023refinedweb} carefully extract high-quality data from the web via heuristic rules and prove that properly curated pertaining corpora lead to powerful LLMs. 
\citet{li2023textbooks} propose phi-1.5, a 1.3 billion parameter LLMs pre-trained on filtered ``textbook-like'' synthetic data, which exhibits many traits of much larger LLMs.
In order to mitigate hallucinations, current LLMs tend to collect pre-training data from credible text sources.
The developers of Llama 2~\citep{touvron2023llama2} strategically up-sample data from highly factual sources, such as Wikipedia\footnote{\revise{While Wikipedia is often regarded as a highly factual and comprehensive source, sometimes there are still exceptions. For instance, a portion of Swedish Wikipedia articles are bot-generated. This reliance on automated article generation raises potential concerns over the accuracy and reliability in certain sections of Wikipedia.}}, when constructing the pre-training corpus.
\citet{lee2022factuality} propose to prepend the topic prefix to sentences in the factual documents to make each sentence serve as a standalone fact during pre-training. Concretely, they treat the document name as the topic prefix and observe this method improves LMs' performance on TruthfulQA.
% Other popular LLMs also only collect pre-training data from 

% \revise{TBD: On page 15, the claim that Wikipedia is a highly factual source should be contextualized, as this can be problematic without further constraints. For example, a significant portion of Swedish Wikipedia articles are bot-generated (https://en.wikipedia.org/wiki/Lsjbot), which raises concerns about their reliability.}

\paragraph{Summary \& Discussion} The mitigation of hallucinations during pre-training is primarily centred around \textit{the curation of pre-training corpora}. Given the vast scale of existing pre-training corpora, current studies predominantly employ simple heuristic rules for data selection and filtering. A potential avenue for exploration could be devising more effective selection or filtering strategies.
% Besides, existing methods for alleviating hallucinations during the pre-training phase are only effective for training new LLMs and are not applicable to rectifying issues in existing LLMs.

% \paragraph{Summary \& Discussion} In summary, \textit{curating the training corpus} involves automatically or manually selecting reliable data or filtering out unreliable data from LLMs' pre-training or instruction-based fine-tuning corpora. Preliminary efforts have demonstrated the effectiveness of this approach in mitigating LLM hallucinations. However, this method necessitates re-training LLMs, which can be costly even with parameter-efficient training techniques~\citep{he2021towards}. Re-training LLMs may also give rise to new challenges, such as impacting the models' original capabilities. Furthermore, this approach may not be highly applicable in situations where information or knowledge is frequently updated, as it is impractical to continuously re-train LLMs. We hope future works in this direction could take the above challenges into consideration.

\begin{table}[t]
\centering
\scalebox{0.8}{
\begin{tabular}{lc}
\toprule
\rowcolor[gray]{.92} \textbf{SFT Dataset} & \textbf{Data Size}  \\  \midrule
Alpaca~\citep{alpaca} & 52k samples \\
GPT4-Alpaca~\citep{peng2023instruction} & 52k samples \\
Baize~\citep{xu2023baize} & 210k samples \\
Dolly~\citep{DatabricksBlog2023DollyV2} & 15k samples\\
Open-assistant~\citep{kopf2023openassistant} &  34k samples \\
LIMA~\cite{zhou2023lima} & 1k samples \\
\bottomrule
\end{tabular}
}
\caption{The size of popular SFT datasets.}
\label{tab:sft:data}
\end{table}

\subsection{Mitigation during SFT}

% \revise{TBD: need to distinguish the two related but different topics “mitigating the hallucination introduced in SFT” and “mitigating hallucination (including those introduced in pre-training) during SFT”}

% \revise{Reorganizing sections 5.2 and 5.3 might improve clarity by treating both as fine-tuning stage defenses, differentiated by the specific fine-tuning approaches they address.}

As a common practice, current LLMs collectively undergo the process known as \textit{supervised fine-tuning} (SFT) to elicit their knowledge acquired from pre-training and learn how to interact with users~\citep{wang-etal-2023-self-instruct, zhang2023instruction}. SFT generally involves first annotating or collecting massive-task instruction-following data~\citep{chung2022scaling,alpaca}, followed by fine-tuning pre-trained foundational LLMs on this data using \textit{maximum likelihood estimation} (MLE)~\citep{wei2021finetuned}. By employing well-designed SFT strategies, many recent studies claim to have built LLMs that achieve performance on par with ChatGPT~\citep{wang2023far} on the specific domain.

\subsubsection{\revise{Mitigating hallucination introduced in pre-training during SFT}
} \revise{The SFT process, as a critical step in aligning LLMs with human preference, can contribute to mitigating hallucinations that originate from the pre-training phase. During pre-training, LLMs are exposed to vast and diverse corpora, which may include inaccurate, outdated, or inconsistent information. These flaws in the training data can result in models generating hallucinated outputs. The SFT process, by using carefully designed instruction-tuning datasets, has the potential to counteract such shortcomings by selectively emphasizing truthfulness, consistency, and reliability in downstream tasks.
To maximize the benefits of this alignment process, curating high-quality instruction-tuning data is crucial. Given the relatively small volume of SFT data (refer to Table~\ref{tab:sft:data}), both manual curation and automatic selection methods are feasible solutions. For instance, \citet{zhou2023lima} constructed a highly curated instruction-tuning dataset containing 1,000 samples, meticulously annotated by human experts to ensure correctness and diversity. On the other hand, studies like~\citep{chen2023alpagasus, cao2023instruction, lee2023platypus} pursue automatic selection strategies, where LLMs act as evaluators or specific rules are devised to filter for high-quality data. Experimental results on hallucination-related benchmarks, such as TruthfulQA~\citep{lin2021truthfulqa}, show that LLMs fine-tuned with these curated datasets outperform those trained on less selective data, achieving greater levels of factuality and truthfulness.
Besides curation strategies, domain-specific knowledge integration during SFT has also been explored to directly tackle hallucinations from specific knowledge gaps. \citet{elaraby2023halo} propose enhancing SFT datasets with content derived from expert-verified, domain-specific knowledge sets. This augmentation helps models fill in factual gaps that otherwise lead to hallucinations when they encounter queries requiring specialized information.} 

\begin{figure}[t!]
\centering
\includegraphics[scale=0.31]{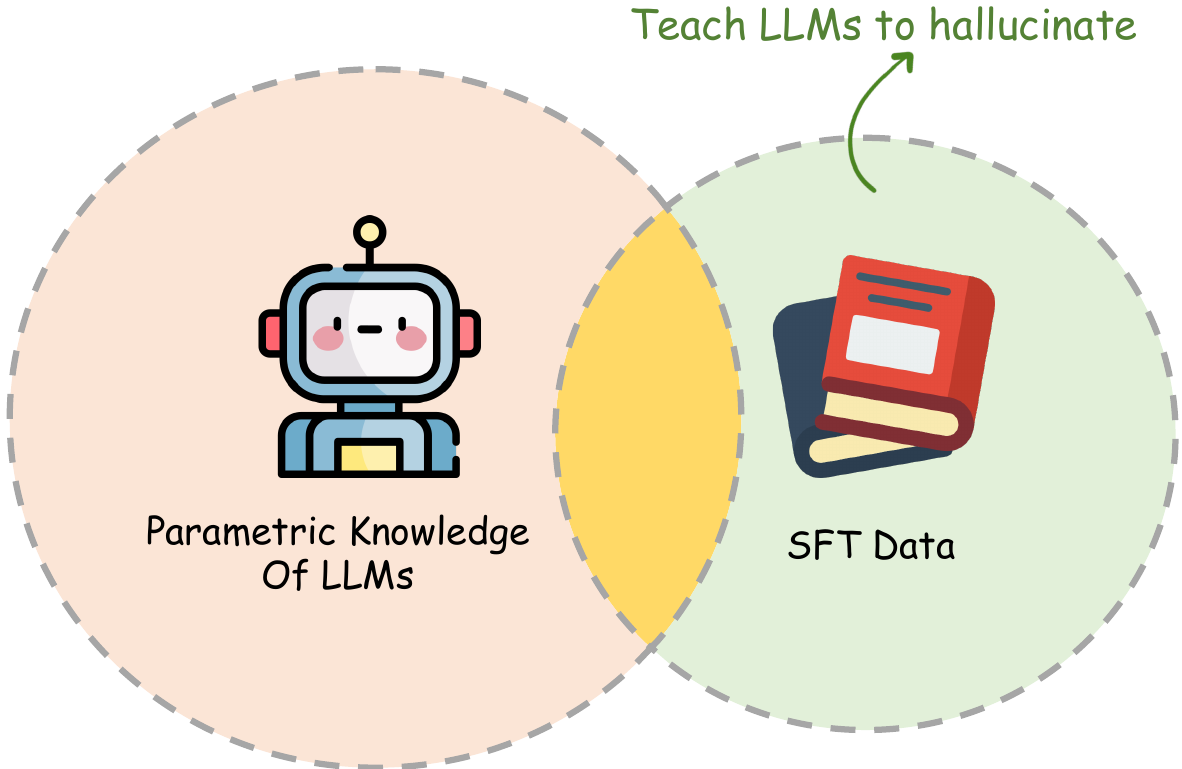}
\caption{The SFT data usually contains samples that exceed LLMs' parametric knowledge, which may result in hallucinations.}
\label{fig:example:sft}
\vspace{-0.2cm}
\end{figure}

\subsubsection{\revise{Mitigating hallucination introduced in SFT}
}
\revise{Another perspective highlights that the supervised fine-tuning (SFT) stage itself may introduce new hallucination risks. \citet{schulman2023youtube} emphasize a potential issue linked to behavior cloning~\citep{torabi2018behavioral}, where the model learns by imitating human-provided demonstrations without understanding the underlying objectives. }
The SFT process of LLMs can be viewed as a special case of behavior cloning, where LLMs learn the format and style of interaction by mimicking humans.
As for LLMs, despite having encoded a substantial amount of knowledge into their parameters, there remains knowledge that surpasses their capacity~\citep{yin2023large,ren2023boundary}.
By cloning human behaviors during SFT, LLMs learn to respond to all questions with a predominantly positive tone, without assessing whether these questions exceed their knowledge boundaries (see Figure~\ref{fig:example:sft}). 
As a result, during inference, if prompted to answer questions related to unlearned knowledge, they are likely to confidently produce hallucinations. 
Furthermore, \citet{gekhman2024does} demonstrates that while SFT instances containing new knowledge are eventually learned, they also increase the model’s tendency to hallucinate.
One way to remit this problem can be the honesty-oriented SFT, which means introducing some honest samples into the SFT data.
The honest samples refer to responses that admit incompetence, such as ``Sorry, I don't know''. The Moss project~\citep{sun2023moss} open-sourced their SFT data, which includes such honest samples. \citet{Llama31} and \citet{wan2024knowledge} teaches the model to refuse to answer by incorporating refusal responses into the supervised fine-tuning (SFT) data.
We observed that models tuned with them could learn to refuse to answer specific questions, therefore helping reduce hallucinations.
Another line of work focuses on contrast training. 
\citet{zou2023representation} introduces a contrast vector loss, which utilizes a pair of contrastive prompts with the same input.
\citet{zhang2024truthx} employs contrastive learning to partition the hidden state into distinct truthful and semantic spaces. The motivation behind this work is that contrastive learning can help the model distinguish between factual and non-factual information.

\paragraph{Summary \& Discussion} \textit{Curating the training data} is one approach for mitigating hallucinations during the SFT phase.
Thanks to the acceptable volume of SFT data, they can be manually curated by human experts. Recently, we have performed a preliminary human inspection and observed that some widely-used synthetic SFT data, such as Alpaca~\citep{alpaca}, contains a considerable amount of hallucinated answers due to the lack of human inspection. 
% For instance, in the GPT-4 version of Alpaca, given the input "Classify the verb as transitive or intransitive: improve", the corresponding output produced by GPT-4 is "Improve is an intransitive verb." However, in reality, "improve" can function as both a transitive and intransitive verb. 
This calls for careful attention when researchers build SFT datasets based on \textit{self-instruct}~\citep{wang-etal-2023-self-instruct}.
% \freda{Wang et al. is the self-consistency paper---double check the reference?}
% We leave the curation of such data for future work.
% We hope further work will consider the truthfulness when building SFT data for LLMs.

Previous work also pointed out that the SFT process may inadvertently introduce hallucinations, by forcing LLMs to answer questions that surpass their knowledge boundaries.  Some researchers have suggested \textit{honesty-oriented SFT} as a solution.
However, we argue this method has two main problems. Firstly, it exhibits limited generalization capabilities towards out-of-distribution (OOD) cases. Secondly, the annotated honest samples just reflect the incompetence and uncertainty of annotators rather than those of LLMs, as annotators are unaware of LLMs' real knowledge boundaries.
Such challenges make solving this issue during SFT sub-optimal.
% since annotators are unaware of the knowledge boundaries of LLMs, it becomes difficult for them to annotate honest samples at the appropriate time. In other words, such honest samples just reflect the incompetence of annotators rather than that of LLMs.

\subsection{Mitigation during RLHF}
\begin{table}[t]
\centering
\scalebox{0.8}{
\begin{tabular}{lc}
\toprule
\rowcolor[gray]{.92} \textbf{Situation} & \textbf{Reward Value}  \\  \midrule
Unhedged Correct & \cellcolor{red!50} +1 \\
Hedged Correct & \cellcolor{red!25} +0.5\\
Uninformative & 0 \\
Hedged Wrong & \cellcolor{teal!25}-2\\
Unhedged Wrong & \cellcolor{teal!50}-4 \\
\bottomrule
\end{tabular}
}
\caption{An example of reward design for mitigating LLM hallucinations through RL~\citep{schulman2023youtube}.}
\label{tab:reward:example}
\end{table}

\begin{table*}[t]
\centering
\scalebox{0.76}{
\begin{tabular}{ll}
\toprule
\rowcolor[gray]{.92}\textbf{User Input} & Introduce the film ``The Only Thing'' (Four Flaming Days) to me.  \\  
\textbf{System Response} & \begin{tabular}[x]{@{}l@{}}``The Only Thing'' (Four Flaming Days) is a \textcolor{red}{captivating drama film} that delves into the complexities of  \\ human emotions, relationships, and the power of resilience. Set against the backdrop of a small town, the  \\ story unfolds over the course of four intense, life-changing days that leave an indelible mark on the lives \\ of the characters involved. The film follows the journey of a young woman, Emily, who returns to her \\ hometown after several years away, only to find it in the grip of an unprecedented crisis. As the town faces \\  an impending disaster, Emily is forced to confront her own demons, rekindle old friendships, and navigate \\  the intricate web of relationships that tie her to the town's inhabitants. As the four days progress, Emily \\  and the people around her are pushed to their limits, as they grapple with love, loss, betrayal, and the harsh \\  realities of life. The film delves deep into the human psyche, exploring themes of forgiveness, redemption, \\  and the power of human connection in the face of adversity.\end{tabular}   \\  
\midrule
\rowcolor[gray]{.92} \textbf{User Input} & What genre of movie is ``The Only Thing'' (Four Flaming Days)?  \\  
\textbf{System Response} & \begin{tabular}[x]{@{}l@{}} I'm sorry. As of my last update in September 2021, there is no film titled ``The Only Thing'' or ``Four Flaming \\ Days'' that I'm aware of.\end{tabular}  \\  
\bottomrule
\end{tabular}
}
\caption{A real example of the \textit{over-conservative} phenomenon of ChatGPT (July 2023 Version). As demonstrated in this example, ChatGPT refuses to provide a fairly clear answer it already knows, specifically, the genre of "The Only Thing" being a drama film (highlighted in \textcolor{red}{red} within the first response).}
\label{tab:over-align:example}
\end{table*}

Nowadays, many researchers attempt to further improve the supervised fine-tuned LLMs via reinforcement learning from human feedback (RLHF)~\citep{fernandes2023bridging}.
This process consists of two steps: 
1) train a reward model (RW) as the proxy for human preference, which aims to assign an appropriate reward value to each LLM response;
2) optimize the SFT model with the reward model's feedback, by using RL algorithms such as PPO~\citep{schulman2017proximal}.

Leveraging human feedback not only closes the gap between machine-generated content and human preference but also helps LLMs align with desired criteria or goals. One commonly used criterion today is ``3H'', which denotes \textit{helpful}, \textit{honest}, and \textit{harmless}~\citep{ouyang2022training, bai2022training,zheng2023secrets}. The \textit{honest} aspect here just refers to the minimization of hallucinations in LLM responses.  Current advanced LLMs, such as InstructGPT~\citep{ouyang2022training}, ChatGPT~\citep{chatgpt}, GPT4~\citep{openai2023gpt4}, and Llama2-Chat~\citep{touvron2023llama2}, have collectively considered this aspect during RLHF. For example, GPT4 uses synthetic hallucination data to train the reward model and perform RL, which increases accuracy on TruthfulQA~\citep{lin2021truthfulqa} from about 30\% to 60\%.
Moreover, ~\citet{lightman2023let} use the \textit{process supervision} to detect and mitigate hallucinations for reasoning tasks, which provides feedback for each intermediate reasoning step.

% The primary advantage of RL is allowing LLMs to freely explore without requiring extensive annotation efforts~\citep{schulman2023youtube}.
As discussed in the previous section, the phenomenon of behavior cloning during the SFT stage can potentially lead to hallucinations. Some researchers have attempted to address this issue by integrating honest samples into the original SFT data. However, this approach has certain limitations, such as unsatisfactory OOD generalization capabilities and a misalignment between human and LLM knowledge boundaries.
In light of this, \citet{schulman2023youtube} propose to solve this problem during RLHF.
They design a special reward function just for mitigating hallucinations, as shown in Table~\ref{tab:reward:example}.
``Unhedged/Hedged Correct/Wrong'' here means the LLM provides correct or wrong answers with a positive or hesitant tone. ``Uninformative'' denote the safe answers like ``I don't know''.
The core idea is to encourage LLMs to challenge the premise, express uncertainty, and commit incapability by learning from specially designed rewards.
This method, which we refer to as honesty-oriented RL, offers several advantages over honesty-oriented SFT. The primary benefit is that it allows LLMs to freely explore their knowledge boundaries, thereby enhancing their generalization capabilities to OOD cases. Additionally, it reduces the need for extensive human annotation and eliminates the requirement for annotators to guess the knowledge boundaries of LLMs.

\paragraph{Summary \& Discussion} 
\textit{Reinforcement learning} can guide LLMs in exploring their knowledge boundaries, enabling them to decline to answer questions beyond their capacity rather than fabricating untruthful responses.
However, we note this approach also poses unique challenges. For instance, RL-tuned LLMs may exhibit over-conservatism due to an imbalanced trade-off between \textit{helpfulness} and \textit{honesty}~\citep{ouyang2022training}. An example of this is illustrated in Table~\ref{tab:over-align:example}. As observed in this case, ChatGPT tends to be overly hedged and refrains from providing a clear answer that it already knows, as evidenced in another dialogue turn. This could be attributed to the unreasonable design of the reward function or the poor quality of the training data for the reward model.
We hope future work can take such problems into consideration.

\subsection{Mitigation during Inference}
Compared with the aforementioned training-time mitigation approaches, mitigating hallucinations in the inference time could be more cost-effective and controllable.
Therefore, most existing studies focus on this direction, which we will introduce in detail in the following sections.
\subsubsection{Designing Decoding Strategies}
% Considering that training LLMs typically require a large amount of computing resources, a
% Another line of work focuses on designing decoding strategies to directly intervene in LLMs' decoding process for reducing hallucinations, which is more cost-effective and controllable.
Decoding strategies, such as greedy decoding and beam search decoding, determine how we choose output tokens from the probability distribution generated by models~\citep{zarriess2021decoding}.

\begin{table*}[t]
\centering
\scalebox{0.7}{
\begin{tabular}{lccc}
\toprule
\rowcolor[gray]{.92} \textbf{Method} & \textbf{Timing of Using} & \textbf{Knowledge Source} & \textbf{Application Task}  \\  \midrule
WebGPT~\citep{nakano2021webgpt}  & Generation-Time & Search API &  QA \\
Adaptive-Retrieval~\citep{mallen2023not}  & Generation-Time & Wikipedia & QA \\
ReACT~\citep{yao2022react}  & Generation-Time & Wikipedia &  QA \& FV \\
RETRO~\citep{borgeaud2022improving}  & Generation-Time & Unstructured Corpus & LM \& QA \\
Chain-of-Knowledge~\citep{li2023chain}  & Generation-Time & Structured Knowledge Base & QA \& FV \& Decision \\ \hdashline
RARR~\citep{gao2023rarr}  & Post-Processing & Search API & QA \\
Verify-then-Edit~\citep{zhao2023automatic}  & Post-Processing & Wikipedia, Search API, etc & QA \\
LLM-Augmenter~\citep{peng2023check}  & Post-Processing & Web documents, Databases & QA \\
REFEED~\citep{yu2023improving}  & Post-Processing & Wikipedia & QA, Dialogue \\
CRITIC~\citep{gou2023critic}  & Post-Processing & Search API, Code Executor, Calculator, etc & QA \&  Program \&  Toxicity \\
FacTool~\citep{chern2023factool}  & Post-Processing & Search API, Code Executor, Calculator, etc & QA \& Reasoning \& Generation \\
\bottomrule
\end{tabular}
}
\caption{A summary of some recent studies on resorting to external knowledge to mitigate hallucinations. We use abbreviations for some application task names, including QA (Question Answering), FV (Fact Verification), and LM (Language Modeling).}
\label{tab:ret:survey}
\end{table*}

\citet{lee2022factuality} carry out a factuality assessment of content generated by LLMs using different decoding strategies. They find that nucleus sampling (a.k.a top-\textit{p} sampling)~\citep{holtzman2019curious} falls short of greedy decoding in terms of factuality. \revise{They argue that this underperformance could be attributed to the randomness introduced by top-\textit{p} sampling to boost diversity, creativity, which may inadvertently lead to hallucinations since LLMs tend to fabricate information to generate diverse responses}.
In view of this, they introduce a decoding algorithm termed \textit{factual-nucleus sampling}, which aims to strike a more effective balance between diversity and factuality by leveraging the strengths of both top-\textit{p} and greedy decoding.

\citet{dhuliawala2023chain} develop a decoding framework known as the Chain-of-Verification (\textsc{CoVe}). This framework is based on the observation that independent verification questions typically yield more accurate facts than those presented in long-form answers. The \textsc{CoVe} framework initially plans verification questions, and then answers these questions to ultimately produce an enhanced, revised response. Experimental results on list-based questions, closed book QA, and long-form text generation demonstrate that \textsc{CoVe} can effectively mitigate hallucination.

Another work, \citet{li2023inference}, introduces a novel \textit{Inference-Time Intervention} (ITI) method to improve the truthfulness of LLMs. This method is based on the assumption that LLMs possess latent, interpretable sub-structures associated with factuality. The ITI method comprises two steps: 1) fitting a binary classifier on top of each attention head of the LLM to identify a set of heads that exhibit superior linear probing accuracy for answering factual questions, and 2) shifting model activations along these factuality-related directions during inference. The ITI method leads to a substantial performance improvement on the TruthfulQA benchmark~\citep{lin2021truthfulqa}.

Distinct from the aforementioned studies, \citet{shi2023trusting} instead concentrates on the retrieval-augmentation setting. Prior research has shown that LLMs sometimes fail to adequately attend to retrieved knowledge when addressing downstream tasks, particularly when the retrieved knowledge conflicts with the parametric knowledge of LLMs~\citep{zhou2023context, xie2023adaptive}. To address this issue, \citet{shi2023trusting} propose a straightforward context-aware decoding (CAD) strategy.
The core idea of CAD is to perform a contrastive ensemble of $p_{\theta}(y_{t}\mid x,c,y_{<t})$ and $p_{\theta}(y_{t}\mid x,y_{<t})$,
where $\theta$ represents the LM, $x$ is the input query, $c$ is the context, $y$ is the response, and $t$ is the time step. $p_{\theta}(y_{t}\mid x,c,y_{<t})$ means the generation probability distribution of $t$-th token when given the context while  $p_{\theta}(y_{t}\mid x,y_{<t})$ denotes the distribution only considering the query. The CAD method aims to compel LLMs to pay more attention to contextual information instead of over-relying their own parametric knowledge to make decisions.  
Experimental results show that CAD effectively elicits the ability of LLMs to exploit retrieved knowledge and thus reduces factual hallucinations on downstream tasks.
Another work, DoLA \citep{chuang2023dola}, also employ the idea of contrastive decoding to reduce hallucination. However, they contrast the generation probabilities from different layers of LLMs, as they find that linguistic and factual information is encoded in distinct sets of layers. 
ICD \citep{zhang2023alleviating} further constructs a factually weak LLM as the penalty term.

\paragraph{Summary \& Discussion} \textit{Designing decoding strategies} to mitigate hallucinations in LLMs during inference is typically in a plug-and-play manner. Therefore, this method is easy to deploy, making it promising for practical applications. However, for this approach, most existing works require accessing the token-level output probabilities, while a substantial number of current LLMs can only return generated content through limited APIs (e.g., ChatGPT). Consequently, we encourage future research in this direction to explore within a more strict \textit{black-box} setting.
 
% Decoding algorithm.
% Inference-time Intervention.

\subsubsection{Resorting to External Knowledge}
\begin{figure}[t!]
\centering
\includegraphics[scale=0.31]{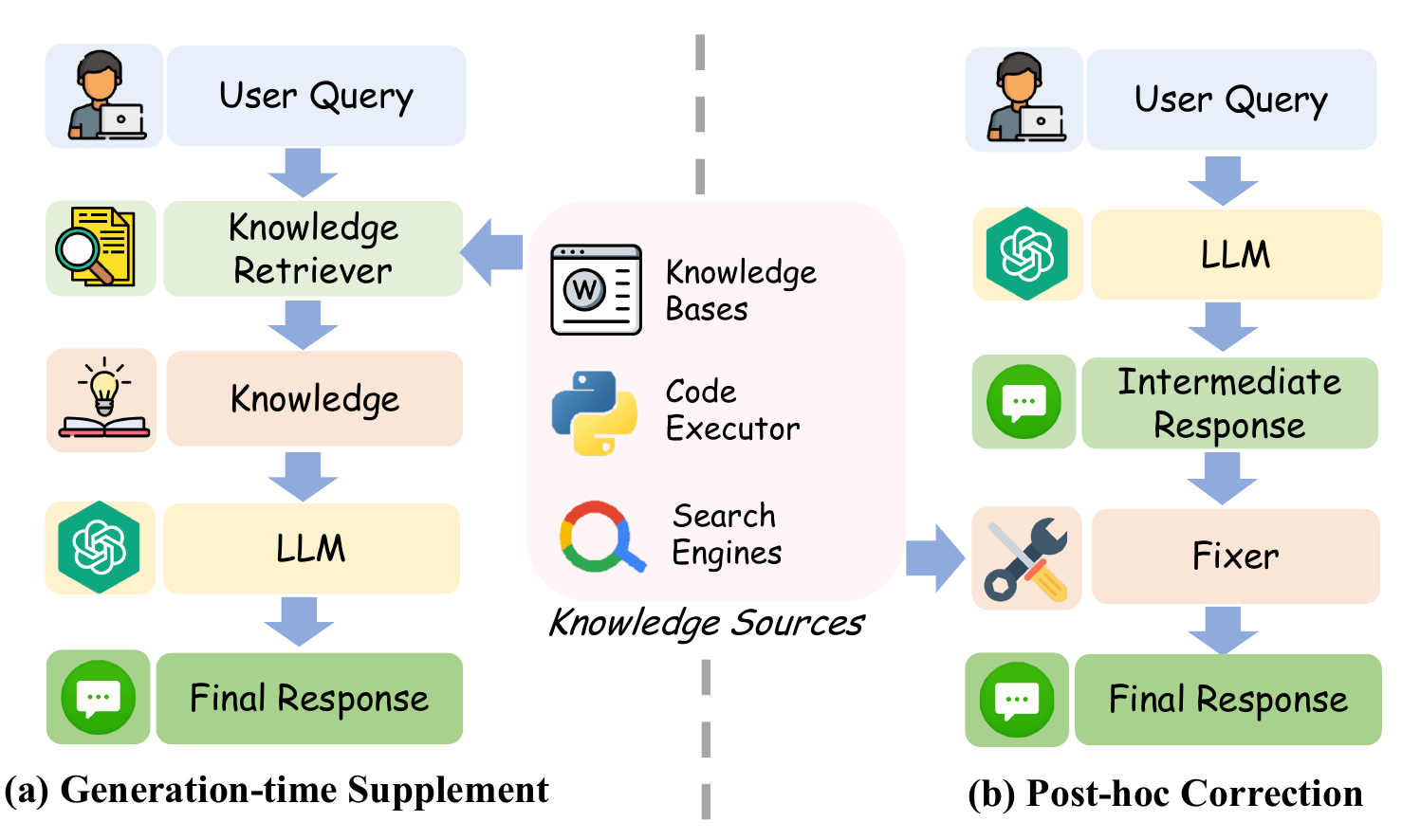}
\caption{The illustrations of two distinct methods for utilizing external knowledge to reduce hallucinations in LLMs' responses.}
\label{fig:example:rag}
\vspace{-0.2cm}
\end{figure}

Using external knowledge as supplementary evidence to assist LLMs in providing truthful responses recently represents a burgeoning solution \citep{ren2023boundary,mialon2023augmented}. This approach typically consists of two steps.
The first step entails accurately obtaining knowledge related to the user instructions. Once useful knowledge has been achieved, the second step involves leveraging such knowledge to guide the generation of the responses. 
We provide a comprehensive review of the latest progress in this direction, focusing on the specific strategies employed in these two steps, respectively. We also present a summary of recent studies in Table~\ref{fig:example:rag}.

\paragraph{Knowledge acquisition} LLMs have internalized vast amounts of knowledge into their parameters through extensive pre-training and fine-tuning, which can be referred to as \textit{parametric knowledge}~\citep{roberts2020much}. However, incorrect or outdated parametric knowledge can easily lead to hallucinations~\citep{xie2023adaptive}. To remedy this, researchers have proposed acquiring reliable, up-to-date knowledge from credible sources as a form of hot patching for LLMs~\citep{lewis2020retrieval, li2022survey}. We summarize the two primary sources of such knowledge as follows.

\begin{itemize}
    \item[(1)] \textbf{External knowledge bases.} The majority of existing works retrieve information from external knowledge bases, such as large-scale unstructured corpora~\citep{cai2021neural, borgeaud2022improving}, structured databases~\citep{Liu_LlamaIndex_2022, li2023chain}, specific websites like Wikipedia~\citep{yao2022react,peng2023check,li2023self,yu2023improving}, or even the entire Internet~\citep{lazaridou2022internet, yao2022react, gao2023rarr,liu2023reta}. The evidence retrieval process typically employs various sparse (e.g., BM25~\citep{robertson2009probabilistic}) or dense (e.g., PLM-based methods~\citep{zhao2022dense}) retrievers. Search engines, such as Google Search, can also be viewed as a special kind of information retriever~\citep{nakano2021webgpt, lazaridou2022internet, yao2022react, gao2023rarr}. Besides, \citet{luo2023augmented} propose the parameter knowledge guiding framework which retrieves knowledge from the parametric memory of fine-tuned white-box LLMs.
    \citet{feng2023knowledge} try to teach LLMs to search relevant domain knowledge from external knowledge graphs to answer domain-specific questions.
    
    \item[(2)] \textbf{External tools.} In addition to solely retrieving information from knowledge bases, there are also many other tools that can provide valuable evidence to enhance the factuality of content generated by LLMs~\citep{mialon2023augmented,qin2023tool,qiao2023making}. For instance, FacTool~\citep{chern2023factool} employs different tools to help detect hallucinations in LLMs for specific downstream tasks, such as \textit{search engine API} for Knowledge-based QA, \textit{code executor} for code generation, and \textit{Google Scholar API} for scientific literature review. CRITIC~\citep{gou2023critic} also enables LLMs to interact with multiple tools and revise their responses autonomously, which has been proven to effectively improve truthfulness.
\end{itemize}

\paragraph{Knowledge utilization} Once relevant knowledge is obtained, it could be employed at different stages to mitigate hallucinations within LLMs. Existing methods for knowledge utilization can be roughly divided into two categories, as detailed below and illustrated in Figure~\ref{fig:example:rag}.

\begin{itemize}
    \item[(1)] \textbf{Generation-time supplement.} The most straightforward approach to utilize retrieved knowledge or tool feedback is to directly concatenate them with user queries before prompting LLMs~\citep{shi2023replug,mallen2023not,ram2023context}. This method is both effective and easy to implement. Such knowledge is also referred to as \textit{context knowledge}~\citep{shi2023trusting}. Existing studies have demonstrated that LLMs possess a strong capability for in-context learning~\citep{dong2022survey}, which enables them to extract and utilize valuable information from context knowledge to rectify nonfactual claims they previously generated.
    \item[(2)] \textbf{Post-hoc correction.} Another common practice involves constructing an auxiliary fixer to rectify hallucinations during the post-processing stage~\citep{cao-etal-2020-factual, zhu-etal-2021-enhancing, fabbri-etal-2022-improving}.
    The fixer can be either another LLM~\citep{peng2023check,zhang2023hybrid,chern2023factool,gou2023critic} or a specific small model~\citep{chen2023purr}.
    Such fixers first interact with external knowledge sources to gather sufficient evidence, and then correct hallucinations. For example, RARR~\citep{gao2023rarr} directly prompts an LLM to ask questions about the content that needs to be corrected from multiple perspectives. Then it uses search engines to retrieve relevant knowledge. The LLM-based fixer finally makes corrections based on retrieved evidence. 
    The Verify-then-Edit approach~\citep{zhao2023verify} aims to enhance the factuality of predictions by post-editing reasoning chains based on external knowledge sourced from Wikipedia.
    To achieve better performance, LLM-Augmenter~\citep{peng2023check} prompts LLMs to summarize retrieved knowledge before feeding it into the fixer.  Moreover, FacTool~\citep{chern2023factool} and CRITIC~\citep{gou2023critic} propose to utilize various external tools to obtain evidence for the fixer.
\end{itemize}

% 1) Retrieval-augmented; 2) Tool-augmented.

% Post-processing vs. demonstration.

% Another common practice in conventional hallucination mitigation works is building an auxiliary fixer to fix hallucinations during the post-processing stage. For example, many previous works in abstraction summarization propose to fine-tune pre-trained language models to remove hallucinated entities or numbers from the summary~\citep{cao-etal-2020-factual, zhu-etal-2021-enhancing, fabbri-etal-2022-improving}. Within the context of LLM, some researchers also start to investigate post-processing techniques to mitigate hallucination recently. \citet{chen2023purr} put forward 

\paragraph{Summary \& Discussion} \textit{Resorting to external knowledge} to mitigate hallucinations in LLMs offers several advantages. Firstly, this method circumvents the need for modifying LLMs, making it a plug-and-play and efficient solution. Secondly, it facilitates the easy transfer of proprietary knowledge (e.g., a company's internal data) and real-time updated information to LLMs. Lastly, this approach enhances the interpretability of information generated by LLMs by allowing the tracing of generation results back to the source evidence~\citep{gao2023enabling, yue2023automatic}. However, this direction also presents some remaining challenges. We discuss some of them below.
% , thereby simplifying the fact-checking process.

\begin{itemize}
    \item[(1)] \textbf{Knowledge verification.} In the era of LLMs, the external knowledge source could extend beyond a single document corpus or a specific website to encompass the entire Internet.  However, the information from the Internet is in the wild, which means they may also be fabricated, or even generated by LLMs themselves~\citep{alemohammad2023self}.  How to verify the authenticity of retrieved knowledge from the Internet is an open and challenging problem to be solved.
    \item[(2)] \textbf{Performance/efficiency of retriever/fixer.} The performance of the retriever/fixer plays a vital role in ensuring the effects of hallucination mitigation. Future work may consider jointly optimising the whole working flow (retriever$\rightarrow$LLM$\rightarrow$fixer) via reinforcement learning~\citep{qiao2023making} or other techniques. Besides, the efficiency of the retriever/fixer is another important factor to be considered, as the generation speed of existing LLMs is already a significant burden~\citep{ning2023skeleton}.
    \item[(3)] \textbf{Knowledge conflict.} As introduced before, the retrieved knowledge may conflict with the parametric knowledge stored by LLMs~\citep{qian2023merge}. \citet{shi2023trusting} reveal that LLMs may fail to sufficiently exploit retrieved knowledge when knowledge conflict happens. \citet{xie2023adaptive} take a more cautious look at this phenomenon. How to fully utilize context knowledge is an under-explored question. For example, \citet{liu2023lost} find the performance of retrieval-augmented LLMs significantly degrades when they must access evidence in the middle of long contexts.
\end{itemize}

% Verification of retrieval knowledge. Performance of Retriever/Editor. Knowledge conflicts/Utilization of retrieval knowledge.
% Train a small model to fix hallucinations after LLMs generate contents.

% \paragraph{Summary \& Discussion} 

% \subsection{Exploiting Uncertainty}

% Another reflection of uncertainty is self-consistency, which refers to the consistency of the factual statements generated by the same LLM.

\subsubsection{Exploiting Uncertainty}
\begin{figure}[t!]
\centering
\includegraphics[scale=0.31]{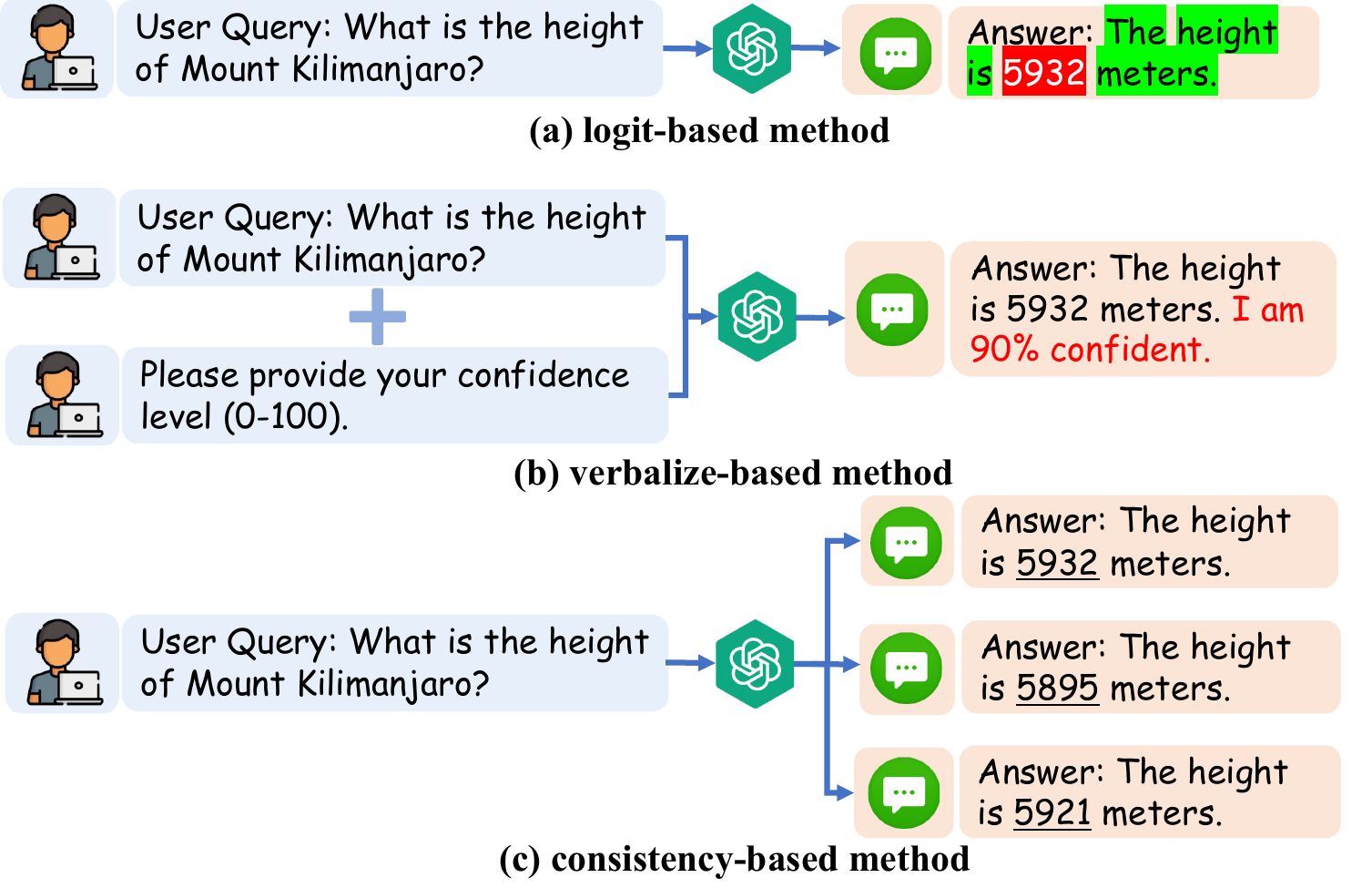}
\caption{The illustrations of three typical methods for estimating LLM uncertainty. In the example of the \textit{logit-based} method, we use the \textcolor{red}{red}/\textcolor{green}{green} background to distinct tokens with low/high generation probabilities. In the example of the \textit{consistency-based} method, the responses are acquired from multiple sampling.}
\label{fig:example:uncertainty}
\vspace{-0.2cm}
\end{figure}

Uncertainty serves as a valuable indicator for detecting and mitigating hallucinations during the inference process~\citep{manakul2023selfcheckgpt}. Typically, it refers to the confidence level of model outputs~\citep{jiang2021can,huang2023look,duan2023shifting,zeng2025reasoningmodel}. Uncertainty can assist users in determining when to trust LLMs. Provided that the uncertainty of LLM responses can be accurately characterized, users can filter out or rectify LLMs' claims with high uncertainty since such claims are more prone to be fabricated ones~\citep{lin2023generating}.

Generally speaking, methods for estimating the uncertainty of LLMs can be categorized into three types~\citep{xiong2023can}, as listed below. To facilitate understanding, we also present illustrative examples for these methods in Figure~\ref{fig:example:uncertainty}.

\begin{itemize}
    \item[(1)] \textbf{Logit-based estimation.} The first method is the \textit{logit-based} method, which requires access to the model logits and typically measures uncertainty by calculating token-level probability or entropy. This method has been widely used in the machine learning community~\citep{guo2017calibration}.
    \item[(2)] \textbf{Verbalize-based estimation.} The second is the \textit{verbalize-based} method, which involves directly requesting LLMs to express their uncertainty, such as using the following prompt:
\textit{``Please answer and provide your confidence score (from 0 to 100).''}
This method is effective due to the impressive verbal and instruction-following capabilities of LLMs. Notably, \citet{xiong2023can} and \citet{zhang2024generativeverifiersrewardmodeling} further suggest using chain-of-thoughts prompts~\citep{wei2022chain} to enhance this method.
    \item[(3)] \textbf{Consistency-based estimation.} The third is the \textit{consistency-based} method~\citep{wang2022self,shi-etal-2022-natural,zhao2023verify}. This method operates on the assumption that LLMs are likely to provide logically inconsistent responses for the same question when they are indecisive and hallucinating facts.

\end{itemize}

Several recent studies have leveraged uncertainty estimation for detecting and mitigating hallucinations in LLMs. \textsc{SelfCheckGPT}~\citet{manakul2023selfcheckgpt} is the first framework to detect LLM hallucinations based on uncertainty measurement in a zero-resource and black-box setting.
They employ a consistency-based approach for uncertainty estimation.
A non-trivial challenge in \textsc{SelfCheckGPT} is determining how to measure the consistency of different responses. \citet{manakul2023selfcheckgpt} perform experiments with BERTScore~\citep{zhang2019bertscore}, QA-based metrics~\citep{wuqafacteval} and n-gram metrics. They finally find that a combination of these approaches yields the best results. 
\citet{agrawal2023language} further adopt the verbalize-based method to evaluate the hallucination rate of LLMs for fabricating references.
\revise{Uncertainty can also serve as an indicator to mitigate hallucination. For example, \citet{mundler2023self} directly utilizes an additional LLM to assess whether two LLM responses are logically contradictory given the same context~\citep{luo2023chatgpt}, which means at least one of them is hallucinated. 
Consequently, they employ another LLM to revise such self-contradictory hallucinations from two responses. 
\citet{varshney2023stitch}, on the other hand, uses the logit-based method to detect false concepts in LLMs' responses with high uncertainty. They then fix such content with auxiliary retrieval-augmented LLMs.}

Besides, \citet{zhao2023automatic} present a Pareto optimal self-supervision framework. This framework utilizes available programmatic supervision to assign a risk score to LLM responses, which can serve as an indicator of hallucinations. \citet{luo2023zero} introduce a pre-detection self-evaluation technique, which aims to evaluate the familiarity of LLMs with the concepts in user prompts and prevent the generation of content about those unfamiliar concepts.

\paragraph{Summary \& Discussion} \textit{Exploiting uncertainty} to identify and mitigate LLM hallucinations is a promising research direction today. Three primary approaches exist for estimating the uncertainty of LLMs, each presenting its unique challenges.
Firstly, the \textit{logit-based} method is becoming less applicable for modern commercial LLMs as they are usually closed-source and black-box, rendering their output logits inaccessible. Secondly, regarding the \textit{verbalize-based} method, researchers have observed that LLMs tend to display a high degree of overconfidence when expressing their confidence~\citep{xiong2023can}, i.e. over-confident. Thirdly, the effective measurement of the consistency of different responses remains an unresolved issue in the \textit{consistency-based} method~\citep{manakul2023selfcheckgpt}. We believe that leveraging uncertainty is crucial in developing trustworthy LLMs and encourage future research to address the aforementioned challenges in this field.

\begin{figure}[t!]
\centering
\includegraphics[scale=0.48]{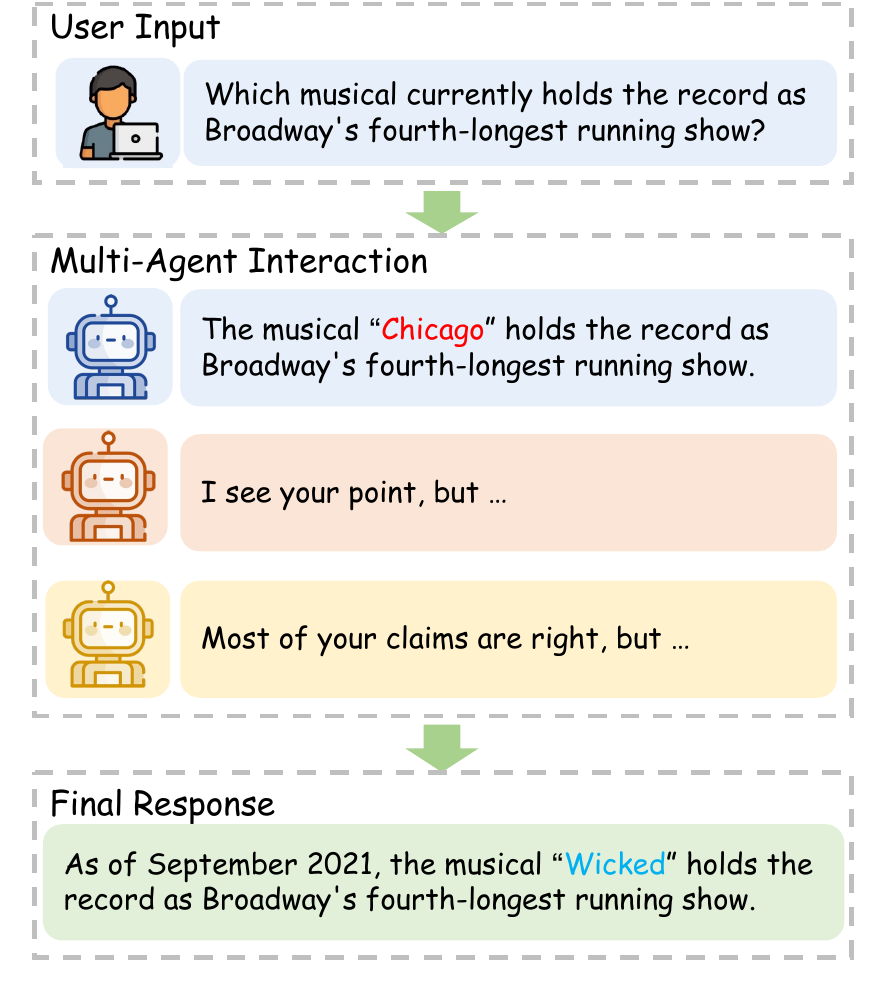}
\caption{An example of the process of multi-agent interaction for mitigating LLM hallucinations.}
\label{fig:example:ma}
\vspace{-0.2cm}
\end{figure}

\subsubsection{Multi-agent interaction}

Some recent research has sought to address the hallucination problem in LLMs from a multi-agent perspective, wherein multiple LLMs (also known as agents) independently propose and collaboratively debate their responses to reach a single consensus, as exemplified in Figure~\ref{fig:example:ma}. 
\citet{du2023improving} is a pioneering work in this line. They initially developed a benchmark for assessing the factual accuracy of prominent computer scientist biographies generated by LMs. Their findings reveal that an individual LLM can easily generate hallucinated information within this benchmark; however, such hallucinations can be mitigated by engaging multiple LLMs in a debate to achieve consensus. Besides, \citet{cohen2023lm} ask one LLM to generate claims (acting as \textsc{Examinee}) and another to raise questions about these claims and check the truthfulness of them (acting as \textsc{Examiner}). \citet{wang2023unleashing} instead propose prompting a single LLM to identify, simulate, and iteratively self-collaborate with multiple personas, such as Harry Potter Fan and Jay Chou Fan. By leveraging an LLM as a cognitive synergist, it effectively reduces hallucinations with relatively low costs.

\subsubsection{Prompt engineering} 
Existing research highlights that the behavior of LLMs can significantly vary based on the prompts given by users~\citep{si2022prompting, zhu2023promptbench}. In terms of hallucination, users may encounter an LLM that initially responds accurately but begins to hallucinate information when using different prompts. In light of this observation, \citet{zhang2023language} endeavour to engineer more effective prompts to mitigate hallucination. Concretely, they employ the chain-of-thought prompt~\citep{wei2022chain} to compel LLMs to generate reasoning steps before providing the final answers. However, chain-of-thought may introduce some new challenges. The potential of hallucinated reasoning steps is one of them. Furthermore, a popular practice nowadays involves explicitly instructing LLMs not to disseminate false or unverifiable information when designing the ``system prompt'', i.e., the special messages used to steer the behavior of LLMs. The following system prompt used for Llama 2-Chat~\citet{touvron2023llama2} exemplifies this approach: \textit{If you don’t know the answer to a question, please don’t share false information.}

\subsubsection{Analyzing LLMs' internal states} \citet{azaria2023internal} contend that LLMs may be aware of their own falsehoods, implying that their internal states could be utilized to detect hallucinations. They propose Statement Accuracy Prediction based on Language Model Activations (SAPLMA), which adds a classifier on top of each hidden layer of the LLM to determine truthfulness. Experimental results indicate that LLMs might ``know'' when the statements they generate are false, and SAPLMA can effectively extract such information.
The Inference-Time Intervention (ITI) method~\citep{li2023inference} is also grounded in a similar hypothesis. They further shift model activations alongside factuality-related heads during inference and discover that this can mitigate hallucinations. These studies suggest that ``the hallucination within LLMs may be more a result of generation techniques than the underlying representation''~\citep{agrawal2023language}.

% Black box. Measure consistency.

% \subsection{Self-Consistency}
% \paragraph{Summary \& Discussion} 

\subsection{Other Methods}
In addition to the above approaches, other techniques demonstrating the potential for reducing hallucinations are shown below. 

% \revise{TBD: Several methods in "5.5 Other Methods," such as ``Multi-agent interaction'', ``Prompt engineering'', and ``Human-in-the-loop'', could be more appropriately categorized under ``inference stage defense''.}

% \paragraph{Post-Processing.} Another common practice in conventional hallucination mitigation works is building an auxiliary fixer to fix hallucinations during the post-processing stage. For example, many previous works in abstraction summarization propose to fine-tune pre-trained language models to remove hallucinated entities or numbers from the summary~\citep{cao-etal-2020-factual, zhu-etal-2021-enhancing, fabbri-etal-2022-improving}. Within the context of LLM, some researchers also start to investigate post-processing techniques to mitigate hallucination recently. \citet{chen2023purr} put forward 

% \citep{azaria2023internal,agrawal2023language}

\paragraph{Human-in-the-loop} \citet{zhang2023mitigating} posit that a potential cause of hallucination in LLMs could be the misalignment between knowledge and user questions, a phenomenon that is particularly prevalent in the context of retrieval-augmented generation (RAG). To address this issue, they introduce MixAlign, a human-in-the-loop framework that utilizes LLMs to align user queries with stored knowledge, and further encourages users to clarify this alignment. By refining user queries iteratively, MixAlign not only reduces hallucinations but also enhances the quality of the generated content. 

\paragraph{Optimizing model architecture} Several studies have explored modifying the architecture of LMs to mitigate hallucinations. Examples include the multi-branch decoder~\citep{rebuffel2022controlling} and the uncertainty-aware decoder~\citep{xiao-wang-2021-hallucination}. \citet{li2023batgpt} suggest employing a bidirectional autoregressive architecture in the construction of LLMs, which enables language modeling from both left-to-right and right-to-left. 
They claim that this design strategy could contribute to the reduction of hallucinations by effectively leveraging bidirectional information. 

% \revise{On page 26, it would be helpful to clarify whether decoding methods fall under "Optimizing model architecture" or if they should be considered purely decoding-based methods.
% }

\section{Outlooks}
In this section, we discuss a few unresolved challenges in the investigation of hallucinations within LLMs and offer our insights into potential future research directions.
\label{sec:outlook}

\paragraph{Reliable evaluation}
Although considerable effort has been dedicated to building evaluation benchmarks for quantitatively assessing hallucination in LLMs, there are still issues that need to be solved. 
The automatic evaluation in the \textit{generation-style} hallucination benchmark cannot accurately reflect the performance or align with human annotation. 
Such inaccuracy is reflected in two ways: 
(1) The automatic metric does not perfectly align with human annotations~\cite{lin2021truthfulqa,min2023factscore,muhlgay2023generating};
(2) The reliability of automatic metric varies across texts from different domains or generated by different LLMs~\cite{min2023factscore}, resulting in reduced robustness for generalization.
Although the \textit{discrimination-style} benchmark~\cite{li2023halueval,muhlgay2023generating} could relatively accurately evaluate a model's ability to distinguish hallucinations, the relationship between discrimination performance and generation performance is still unclear until now. These issues all need more in-depth exploration.
% Specifically, there is limited discussion~\citep{li2023inference} on whether an LLM that effectively discriminates non-factual statements can generate highly factual texts.
% \textit{TBD}

\begin{table*}[t]
\centering
\scalebox{0.8}{
\begin{tabular}{llc}
\toprule
\rowcolor[gray]{.92}\textbf{User Input (EN)} & \begin{tabular}[x]{@{}l@{}}What is the population of Denver according to the 2020 census?\end{tabular}  \\  
\textbf{LLM Response (EN)} & \begin{tabular}[x]{@{}l@{}}According to the 2020 United States Census, the population of Denver, Colorado, is \underline{715,522} \cmark\end{tabular}   \\  
\midrule
\rowcolor[gray]{.92} \textbf{User Input (ZH)} & \begin{tabular}[x]{@{}l@{}}根据2020年人口普查，丹佛的人口是多少？\end{tabular}   \\  
\textbf{LLM Response (ZH)} & \begin{tabular}[x]{@{}l@{}} 根据2020年人口普查，丹佛的人口为\underline{73,921} \xmark\end{tabular}  \\ 
\midrule
\rowcolor[gray]{.92} \textbf{User Input (EN)} & \begin{tabular}[x]{@{}l@{}}What is the population of Denver according to the 2020 census? Answer in Chinese.\end{tabular}   \\  
\textbf{LLM Response (ZH)} & \begin{tabular}[x]{@{}l@{}} 根据2020年人口普查，丹佛的人口为\underline{704,621} \xmark\end{tabular}  \\ 
\bottomrule
\end{tabular}
}
\caption{A real example in which ChatGPT (July 2023 Version) accurately answered a question in English conversation but presented hallucinations for the same question when communicating in Chinese (the correct population of Denver in 2020 is 715,522, according to \url{https://en.wikipedia.org/wiki/Denver}).}
\label{tab:ml:example}
\end{table*}

\paragraph{Multi-lingual hallucination} Existing work in LLM hallucination primarily focuses on English, despite the existence of thousands of languages in the world. We hope that LLMs can possess the ability to handle various languages uniformly. 
Some previous studies have investigated the performance of LLMs on some multi-lingual benchmarks~\citep{ahuja2023mega,lai2023chatgpt}, and collectively found that their performance degenerates when generalizing to non-Latin languages.
In terms of the hallucination problem, \citet{guerreiro2023hallucinations} observe that multi-lingual LLMs predominantly struggle with hallucinations in low-resource languages in the translation task. Potential follow-up work could include systematically measuring and analyzing LLM hallucinations across a wide variety of languages.
As shown in Table~\ref{tab:ml:example}, we find that LLMs such as ChatGPT provide accurate answers in English but expose hallucinations in other languages, leading to multilingual inconsistencies.
The transfer of knowledge within LLMs from high-resource languages to low-resource ones also presents an interesting and promising research direction.

\begin{figure}[t!]
\centering
\includegraphics[scale=0.48]{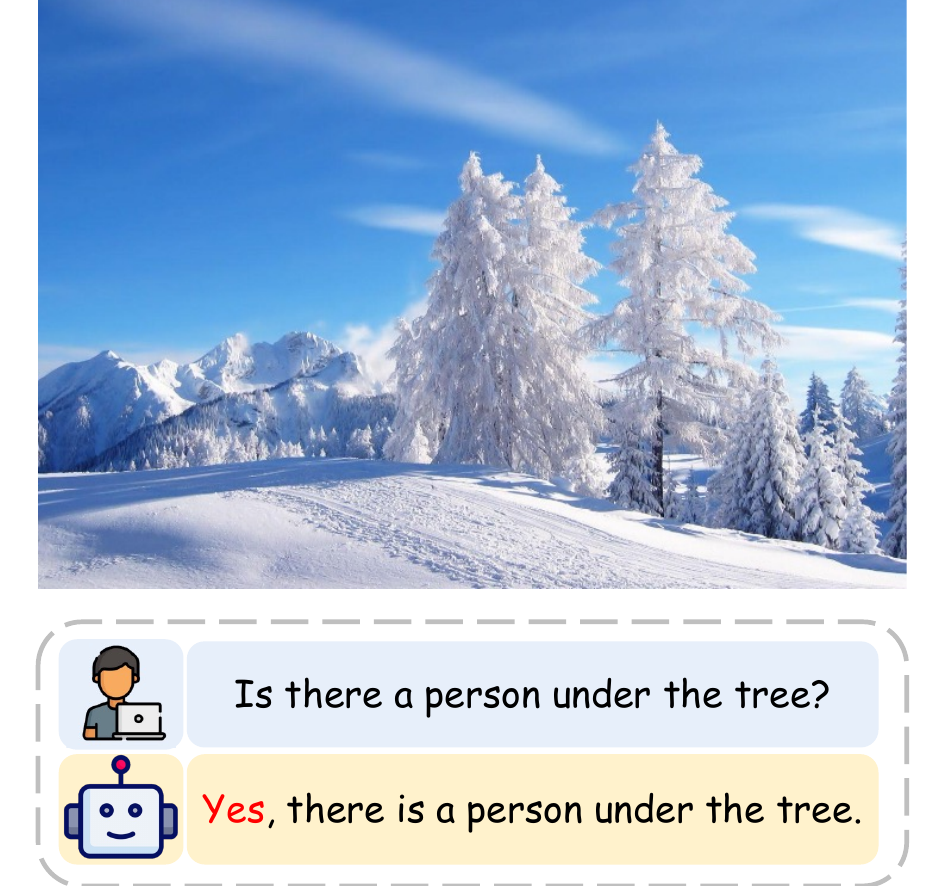}
\caption{An example of object hallucination in LVLMs. We highlight the hallucination in \textcolor{red}{red}, as there is no person under the tree in this picture.}
\label{fig:example:mmh}
\vspace{-0.2cm}
\end{figure}

\paragraph{Multi-modal hallucination} In an effort to improve the performance of complex multi-modal tasks, recent studies have proposed replacing the text encoder of existing vision-large models with LLMs, resulting in large vision-language models (LVLMs)~\citep{liu2023visual,ye2023mplug}. Despite their success, some research reveals that LVLMs inherit the hallucination problem from LLMs and exhibit more severe multi-modal hallucinations compared to smaller models. For instance, \citet{li2023evaluating} discuss the \textit{object hallucination} of LVLMs, wherein LVLMs generate content containing objects that are inconsistent with or absent from the input image, such as the example in Figure~\ref{fig:example:mmh}. To effectively measure object hallucinations generated by LVLMs, \citet{liu2023aligning} propose a \textit{GPT4-Assisted Visual Instruction Evaluation} (GAVIE) benchmark.
\citet{gunjal2023detecting} introduce a multi-modal hallucination detection dataset named M-HalDetect, further study the unfaithful descriptions and inaccurate relationships beyond object hallucinations in LVLMs.
Furthermore, in addition to images, some studies have extended LLMs to other modalities such as audio~\citep{wu2023decoder,su2023pandagpt} and video~\citep{maaz2023video}, making it interesting to investigate hallucination in these new scenarios.

\paragraph{Model editing} As elaborated in $\S$ \ref{sec:cause}, hallucinations in LLMs may primarily stem from the memorization of false information or the absence of correct factual knowledge. To mitigate these issues in LLMs with minimal computational overhead, the concept of model editing has been introduced~\citep{sinitsin2020editable,de-cao-etal-2021-editing}. This approach involves modifying the behavior of models in a manner that is both data- and computation-efficient.
At present, there are two mainstream paradigms for model editing. The first involves the incorporation of an auxiliary sub-network~\citep{mitchell2022memory,huang2023transformer}, while the second entails direct modification of the original model parameters~\citep{meng2022locating,meng2022mass}. This technique may be instrumental in eliminating LLMs' hallucinations by editing their stored factual knowledge in purpose~\citep{lanham2023measuring,onoe2023can}.
However, this emerging field still faces numerous challenges. These could include editing black-box LLMs \citep{murty-etal-2022-fixing}, in-context model editing \citep{zheng2023can}, and multi-hop model editing \citep{zhong2023mquake}, etc.

\paragraph{Attack/defense for inducing hallucination} As previously discussed, significant efforts have been undertaken by both researchers and companies to guarantee that LLMs produce truthful responses, ultimately improving the overall user experience. Cutting-edge commercial LLMs, such as GPT4~\citep{openai2023gpt4}, appear to have acquired a decent ability to generate proper responses to factuality-related queries. However, they are not invincible. Several studies show that LLMs can be manipulated using techniques like meticulously crafted jailbreak prompts to elicit arbitrary desired responses~\citep{wei2023jailbroken,zou2023universal}, including hallucinations. 
Consequently, the attacking and defending strategies for inducing hallucinations could also be a promising research direction. This is particularly important as the generation of fabricated information could potentially breach relevant laws, leading to the forced shutdown of LLM applications. 
This direction is also intimately tied to the robustness of existing hallucination mitigation methods.

\paragraph{More Settings for defining and mitigating hallucination} Given that the current research on hallucinations in LLMs is still in its early stages, there are also many other intriguing and promising avenues for further investigation. For instance, researchers have begun to treat LLMs as agents for open-world planning in the pursuit of AGI~\citep{park2023generative, wang2023voyager}. 
Addressing the hallucination problem within the context of LLMs-as-agents presents brand-new challenges and holds considerable practical value. 
Besides, analyzing and tracing LLM hallucinations from the linguistic aspect is another interesting research topic.
\citet{rawte2023exploring} show that the occurrence of LLM hallucination is closely related to linguistic nuances of the user prompts, such as readability, formality, and concreteness.
% The robustness of existing 
We believe all these directions merit thorough exploration in future research.

\paragraph{Reasoning Models}
Large reasoning models have achieved success in various tasks. Their main characteristic is self-reflective behaviors, such as ``wait, but''. Recent work~\cite{zeng2025reasoningmodel,yang2025boundary} has found that large reasoning models have achieved significant improvements in calibration compared to non-reasoning models.
Furthermore, \citet{damani2025binaryrewardstraininglms} demonstrate that calibration can be further optimized through calibration-reward-based reinforcement learning.
This indicates that training with this reasoning paradigm not only enhances the model's performance but also allows it to better gauge the boundaries of its own capabilities.

\section{Conclusion}

With their strong understanding and generation capabilities in the open domain, LLMs have garnered significant attention from both academic and industrial communities. However, hallucination remains a critical challenge that impedes the practical application of LLMs.
In this survey, we offer a comprehensive review of the most recent advances, primarily post the release of ChatGPT, that aim to evaluate, trace, and eliminate hallucinations within LLMs. 
We also delve into the existing challenges and discuss potential future directions.
We aspire for this survey to serve as a valuable resource for researchers intrigued by the mystery of LLM hallucinations, thereby fostering the practical application of LLMs.

\section*{\revise{Ethic issues of Hallucination in LLMs}}

% \section{Limitations}

% TBD

% \revise{TBD: add ethical issues, limitations, and criteria for selecting papers.}

\revise{As LLMs become increasingly integrated into various applications, hallucination raises significant ethical concerns. These ethical issues need to be carefully considered to mitigate potential harm and ensure responsible use of these powerful technologies. We list several ethical issues about LLM hallucinations as follows.}

\revise{
\paragraph{Misinformation and public trust}
One of the most pressing ethical issues associated with LLM hallucinations is the potential spread of misinformation. When LLMs generate hallucinated responses, it can be inadvertently disseminated by users who trust the outputs of these models. As we discussed earlier, this is particularly concerning in domains such as healthcare, law, and finance, where inaccurate information can have serious consequences. The erosion of public trust in automated systems and artificial intelligence can also result from frequent hallucinations, undermining the perceived reliability of these technologies.}

\revise{\paragraph{Impact on decision making} Hallucinations in LLMs can have significant implications for decision making in both individual and institutional contexts. For instance, if a legal advisor or medical professional relies on hallucinated outputs from an LLM, the} 
\revise{consequences can be deeply detrimental. Ethically, it is essential to ensure that critical decisions are not unduly influenced by unreliable AI outputs, and that there are robust mechanisms for verifying the information provided by LLMs.}
% , finding widespread use in various fields.
% , including but not limited to legal, medical, and educational domains.

% We have also listed some useful resources in the GitHub repository: \url{https://github.com/HillZhang1999/llm-hallucination-survey}, and warmly welcome any kind of feedback and suggestions to make this survey better.

\bibliography{tacl2021}

\begin{thebibliography}{241}
\expandafter\ifx\csname natexlab\endcsname\relax\def\natexlab#1{#1}\fi

\bibitem[{Adlakha et~al.(2023)Adlakha, BehnamGhader, Lu, Meade, and Reddy}]{adlakha2023evaluating}
Vaibhav Adlakha, Parishad BehnamGhader, Xing~Han Lu, Nicholas Meade, and Siva Reddy. 2023.
\newblock Evaluating correctness and faithfulness of instruction-following models for question answering.
\newblock \emph{arXiv preprint arXiv:2307.16877}.

\bibitem[{Agrawal et~al.(2023)Agrawal, Mackey, and Kalai}]{agrawal2023language}
Ayush Agrawal, Lester Mackey, and Adam~Tauman Kalai. 2023.
\newblock Do language models know when they're hallucinating references?
\newblock \emph{arXiv preprint arXiv:2305.18248}.

\bibitem[{Ahuja et~al.(2023)Ahuja, Hada, Ochieng, Jain, Diddee, Maina, Ganu, Segal, Axmed, Bali et~al.}]{ahuja2023mega}
Kabir Ahuja, Rishav Hada, Millicent Ochieng, Prachi Jain, Harshita Diddee, Samuel Maina, Tanuja Ganu, Sameer Segal, Maxamed Axmed, Kalika Bali, et~al. 2023.
\newblock Mega: Multilingual evaluation of generative ai.
\newblock \emph{arXiv preprint arXiv:2303.12528}.

\bibitem[{Aky{\"u}rek et~al.(2022)Aky{\"u}rek, Bolukbasi, Liu, Xiong, Tenney, Andreas, and Guu}]{akyurek2022tracing}
Ekin Aky{\"u}rek, Tolga Bolukbasi, Frederick Liu, Binbin Xiong, Ian Tenney, Jacob Andreas, and Kelvin Guu. 2022.
\newblock Tracing knowledge in language models back to the training data.
\newblock \emph{arXiv preprint arXiv:2205.11482}.

\bibitem[{Alemohammad et~al.(2023)Alemohammad, Casco-Rodriguez, Luzi, Humayun, Babaei, LeJeune, Siahkoohi, and Baraniuk}]{alemohammad2023self}
Sina Alemohammad, Josue Casco-Rodriguez, Lorenzo Luzi, Ahmed~Imtiaz Humayun, Hossein Babaei, Daniel LeJeune, Ali Siahkoohi, and Richard~G Baraniuk. 2023.
\newblock Self-consuming generative models go mad.
\newblock \emph{arXiv preprint arXiv:2307.01850}.

\bibitem[{Azaria and Mitchell(2023)}]{azaria2023internal}
Amos Azaria and Tom Mitchell. 2023.
\newblock The internal state of an llm knows when its lying.
\newblock \emph{arXiv preprint arXiv:2304.13734}.

\bibitem[{Bai et~al.(2022)Bai, Jones, Ndousse, Askell, Chen, DasSarma, Drain, Fort, Ganguli, Henighan et~al.}]{bai2022training}
Yuntao Bai, Andy Jones, Kamal Ndousse, Amanda Askell, Anna Chen, Nova DasSarma, Dawn Drain, Stanislav Fort, Deep Ganguli, Tom Henighan, et~al. 2022.
\newblock Training a helpful and harmless assistant with reinforcement learning from human feedback.
\newblock \emph{arXiv preprint arXiv:2204.05862}.

\bibitem[{Bai et~al.(2024)Bai, Tu, Zhang, Peng, Wang, Lv, Cao, Xu, Hou, Dong et~al.}]{bai2024longbench}
Yushi Bai, Shangqing Tu, Jiajie Zhang, Hao Peng, Xiaozhi Wang, Xin Lv, Shulin Cao, Jiazheng Xu, Lei Hou, Yuxiao Dong, et~al. 2024.
\newblock Longbench v2: Towards deeper understanding and reasoning on realistic long-context multitasks.
\newblock \emph{arXiv preprint arXiv:2412.15204}.

\bibitem[{Bang et~al.(2023)Bang, Cahyawijaya, Lee, Dai, Su, Wilie, Lovenia, Ji, Yu, Chung et~al.}]{bang2023multitask}
Yejin Bang, Samuel Cahyawijaya, Nayeon Lee, Wenliang Dai, Dan Su, Bryan Wilie, Holy Lovenia, Ziwei Ji, Tiezheng Yu, Willy Chung, et~al. 2023.
\newblock A multitask, multilingual, multimodal evaluation of chatgpt on reasoning, hallucination, and interactivity.
\newblock \emph{arXiv preprint arXiv:2302.04023}.

\bibitem[{Bickel et~al.(2005)Bickel, Haider, and Scheffer}]{bickel2005predicting}
Steffen Bickel, Peter Haider, and Tobias Scheffer. 2005.
\newblock Predicting sentences using n-gram language models.
\newblock In \emph{Proceedings of human language technology conference and conference on empirical methods in natural language processing}, pages 193--200.

\bibitem[{Borgeaud et~al.(2022)Borgeaud, Mensch, Hoffmann, Cai, Rutherford, Millican, Van Den~Driessche, Lespiau, Damoc, Clark et~al.}]{borgeaud2022improving}
Sebastian Borgeaud, Arthur Mensch, Jordan Hoffmann, Trevor Cai, Eliza Rutherford, Katie Millican, George~Bm Van Den~Driessche, Jean-Baptiste Lespiau, Bogdan Damoc, Aidan Clark, et~al. 2022.
\newblock Improving language models by retrieving from trillions of tokens.
\newblock In \emph{International conference on machine learning}, pages 2206--2240. PMLR.

\bibitem[{Brown et~al.(2020)Brown, Mann, Ryder, Subbiah, Kaplan, Dhariwal, Neelakantan, Shyam, Sastry, Askell et~al.}]{brown2020language}
Tom Brown, Benjamin Mann, Nick Ryder, Melanie Subbiah, Jared~D Kaplan, Prafulla Dhariwal, Arvind Neelakantan, Pranav Shyam, Girish Sastry, Amanda Askell, et~al. 2020.
\newblock Language models are few-shot learners.
\newblock \emph{Advances in neural information processing systems}, 33:1877--1901.

\bibitem[{Cai et~al.(2021)Cai, Wang, Li, Lam, and Liu}]{cai2021neural}
Deng Cai, Yan Wang, Huayang Li, Wai Lam, and Lemao Liu. 2021.
\newblock Neural machine translation with monolingual translation memory.
\newblock In \emph{Proceedings of the 59th Annual Meeting of the Association for Computational Linguistics and the 11th International Joint Conference on Natural Language Processing (Volume 1: Long Papers)}, pages 7307--7318.

\bibitem[{Cao et~al.(2020)Cao, Dong, Wu, and Cheung}]{cao-etal-2020-factual}
Meng Cao, Yue Dong, Jiapeng Wu, and Jackie Chi~Kit Cheung. 2020.
\newblock Factual error correction for abstractive summarization models.
\newblock In \emph{Proceedings of the 2020 Conference on Empirical Methods in Natural Language Processing (EMNLP)}, pages 6251--6258.

\bibitem[{Cao et~al.(2023{\natexlab{a}})Cao, Kang, and Sun}]{cao2023instruction}
Yihan Cao, Yanbin Kang, and Lichao Sun. 2023{\natexlab{a}}.
\newblock Instruction mining: High-quality instruction data selection for large language models.
\newblock \emph{arXiv preprint arXiv:2307.06290}.

\bibitem[{Cao et~al.(2023{\natexlab{b}})Cao, Yang, and Zhao}]{autohall}
Zouying Cao, Yifei Yang, and Hai Zhao. 2023{\natexlab{b}}.
\newblock \href {https://doi.org/10.48550/ARXIV.2310.00259} {Autohall: Automated hallucination dataset generation for large language models}.
\newblock \emph{CoRR}, abs/2310.00259.

\bibitem[{Chang et~al.(2019)Chang, Prabhakaran, and Ordonez}]{chang2019bias}
Kai-Wei Chang, Vinodkumar Prabhakaran, and Vicente Ordonez. 2019.
\newblock Bias and fairness in natural language processing.
\newblock In \emph{Proceedings of the 2019 Conference on Empirical Methods in Natural Language Processing and the 9th International Joint Conference on Natural Language Processing (EMNLP-IJCNLP): Tutorial Abstracts}.

\bibitem[{Chang et~al.(2023)Chang, Wang, Wang, Wu, Zhu, Chen, Yang, Yi, Wang, Wang et~al.}]{chang2023survey}
Yupeng Chang, Xu~Wang, Jindong Wang, Yuan Wu, Kaijie Zhu, Hao Chen, Linyi Yang, Xiaoyuan Yi, Cunxiang Wang, Yidong Wang, et~al. 2023.
\newblock A survey on evaluation of large language models.
\newblock \emph{arXiv preprint arXiv:2307.03109}.

\bibitem[{Chen et~al.(2023{\natexlab{a}})Chen, Pasupat, Singh, Lee, and Guu}]{chen2023purr}
Anthony Chen, Panupong Pasupat, Sameer Singh, Hongrae Lee, and Kelvin Guu. 2023{\natexlab{a}}.
\newblock Purr: Efficiently editing language model hallucinations by denoising language model corruptions.
\newblock \emph{arXiv preprint arXiv:2305.14908}.

\bibitem[{Chen et~al.(2017)Chen, Liu, Yin, and Tang}]{chen2017survey}
Hongshen Chen, Xiaorui Liu, Dawei Yin, and Jiliang Tang. 2017.
\newblock A survey on dialogue systems: Recent advances and new frontiers.
\newblock \emph{Acm Sigkdd Explorations Newsletter}, 19(2):25--35.

\bibitem[{Chen et~al.(2024{\natexlab{a}})Chen, Yang, Wu, Jiang, Hou, Li, Wang, Xiao, Li, and Zhang}]{chen2024detecting}
Jiawei Chen, Dingkang Yang, Tong Wu, Yue Jiang, Xiaolu Hou, Mingcheng Li, Shunli Wang, Dongling Xiao, Ke~Li, and Lihua Zhang. 2024{\natexlab{a}}.
\newblock Detecting and evaluating medical hallucinations in large vision language models.
\newblock \emph{arXiv preprint arXiv:2406.10185}.

\bibitem[{Chen et~al.(2023{\natexlab{b}})Chen, Li, Yan, Wang, Gunaratna, Yadav, Tang, Srinivasan, Zhou, Huang et~al.}]{chen2023alpagasus}
Lichang Chen, Shiyang Li, Jun Yan, Hai Wang, Kalpa Gunaratna, Vikas Yadav, Zheng Tang, Vijay Srinivasan, Tianyi Zhou, Heng Huang, et~al. 2023{\natexlab{b}}.
\newblock Alpagasus: Training a better alpaca with fewer data.
\newblock \emph{arXiv preprint arXiv:2307.08701}.

\bibitem[{Chen et~al.(2024{\natexlab{b}})Chen, Song, Gui, Wang, Zhang, Yong, Huang, Lv, Zhang, and Chen}]{chen2024factchd}
Xiang Chen, Duanzheng Song, Honghao Gui, Chengxi Wang, Ningyu Zhang, Jiang Yong, Fei Huang, Chengfei Lv, Dan Zhang, and Huajun Chen. 2024{\natexlab{b}}.
\newblock Factchd: Benchmarking fact-conflicting hallucination detection.
\newblock In \emph{Proceedings of the 33rd International Joint Conference on Artificial Intelligence}.

\bibitem[{Cheng et~al.(2023)Cheng, Sun, Zhang, Wang, Liu, Zhang, He, Huang, Yin, Chen et~al.}]{cheng2023evaluating}
Qinyuan Cheng, Tianxiang Sun, Wenwei Zhang, Siyin Wang, Xiangyang Liu, Mozhi Zhang, Junliang He, Mianqiu Huang, Zhangyue Yin, Kai Chen, et~al. 2023.
\newblock Evaluating hallucinations in chinese large language models.
\newblock \emph{arXiv preprint arXiv:2310.03368}.

\bibitem[{Chern et~al.(2023)Chern, Chern, Chen, Yuan, Feng, Zhou, He, Neubig, and Liu}]{chern2023factool}
I-Chun Chern, Steffi Chern, Shiqi Chen, Weizhe Yuan, Kehua Feng, Chunting Zhou, Junxian He, Graham Neubig, and Pengfei Liu. 2023.
\newblock Factool: Factuality detection in generative ai -- a tool augmented framework for multi-task and multi-domain scenarios.
\newblock \emph{arXiv preprint arXiv:2307.13528}.

\bibitem[{Chowdhery et~al.(2022)Chowdhery, Narang, Devlin, Bosma, Mishra, Roberts, Barham, Chung, Sutton, Gehrmann et~al.}]{chowdhery2022palm}
Aakanksha Chowdhery, Sharan Narang, Jacob Devlin, Maarten Bosma, Gaurav Mishra, Adam Roberts, Paul Barham, Hyung~Won Chung, Charles Sutton, Sebastian Gehrmann, et~al. 2022.
\newblock Palm: Scaling language modeling with pathways.
\newblock \emph{arXiv preprint arXiv:2204.02311}.

\bibitem[{Chuang et~al.(2023)Chuang, Xie, Luo, Kim, Glass, and He}]{chuang2023dola}
Yung-Sung Chuang, Yujia Xie, Hongyin Luo, Yoon Kim, James Glass, and Pengcheng He. 2023.
\newblock Dola: Decoding by contrasting layers improves factuality in large language models.
\newblock \emph{arXiv preprint arXiv:2309.03883}.

\bibitem[{Chung et~al.(2022)Chung, Hou, Longpre, Zoph, Tay, Fedus, Li, Wang, Dehghani, Brahma et~al.}]{chung2022scaling}
Hyung~Won Chung, Le~Hou, Shayne Longpre, Barret Zoph, Yi~Tay, William Fedus, Eric Li, Xuezhi Wang, Mostafa Dehghani, Siddhartha Brahma, et~al. 2022.
\newblock Scaling instruction-finetuned language models.
\newblock \emph{arXiv preprint arXiv:2210.11416}.

\bibitem[{Cohen et~al.(2023)Cohen, Hamri, Geva, and Globerson}]{cohen2023lm}
Roi Cohen, May Hamri, Mor Geva, and Amir Globerson. 2023.
\newblock Lm vs lm: Detecting factual errors via cross examination.
\newblock \emph{arXiv preprint arXiv:2305.13281}.

\bibitem[{Conover et~al.(2023)Conover, Hayes, Mathur, Xie, Wan, Shah, Ghodsi, Wendell, Zaharia, and Xin}]{DatabricksBlog2023DollyV2}
Mike Conover, Matt Hayes, Ankit Mathur, Jianwei Xie, Jun Wan, Sam Shah, Ali Ghodsi, Patrick Wendell, Matei Zaharia, and Reynold Xin. 2023.
\newblock \href {https://www.databricks.com/blog/2023/04/12/dolly-first-open-commercially-viable-instruction-tuned-llm} {Free dolly: Introducing the world's first truly open instruction-tuned llm}.

\bibitem[{Cui et~al.(2021)Cui, Wu, Liu, and Zhang}]{ke-blender}
Leyang Cui, Yu~Wu, Shujie Liu, and Yue Zhang. 2021.
\newblock Knowledge enhanced fine-tuning for better handling unseen entities in dialogue generation.
\newblock In \emph{EMNLP}.

\bibitem[{Dale et~al.(2023)Dale, Voita, Barrault, and Costa{-}juss{\`{a}}}]{Dale2023mthalu2}
David Dale, Elena Voita, Lo{\"{\i}}c Barrault, and Marta~R. Costa{-}juss{\`{a}}. 2023.
\newblock \href {https://doi.org/10.18653/v1/2023.acl-long.3} {Detecting and mitigating hallucinations in machine translation: Model internal workings alone do well, sentence similarity even better}.
\newblock In \emph{Proceedings of the 61st Annual Meeting of the Association for Computational Linguistics (Volume 1: Long Papers), {ACL} 2023, Toronto, Canada, July 9-14, 2023}, pages 36--50. Association for Computational Linguistics.

\bibitem[{Damani et~al.(2025)Damani, Puri, Slocum, Shenfeld, Choshen, Kim, and Andreas}]{damani2025binaryrewardstraininglms}
Mehul Damani, Isha Puri, Stewart Slocum, Idan Shenfeld, Leshem Choshen, Yoon Kim, and Jacob Andreas. 2025.
\newblock \href {http://arxiv.org/abs/2507.16806} {Beyond binary rewards: Training lms to reason about their uncertainty}.

\bibitem[{De~Cao et~al.(2021)De~Cao, Aziz, and Titov}]{de-cao-etal-2021-editing}
Nicola De~Cao, Wilker Aziz, and Ivan Titov. 2021.
\newblock Editing factual knowledge in language models.
\newblock In \emph{Proceedings of the 2021 Conference on Empirical Methods in Natural Language Processing}, pages 6491--6506.

\bibitem[{Devlin et~al.(2019)Devlin, Chang, Lee, and Toutanova}]{devlin-etal-2019-bert}
Jacob Devlin, Ming-Wei Chang, Kenton Lee, and Kristina Toutanova. 2019.
\newblock {BERT}: Pre-training of deep bidirectional transformers for language understanding.
\newblock In \emph{Proceedings of the 2019 Conference of the North {A}merican Chapter of the Association for Computational Linguistics: Human Language Technologies, Volume 1 (Long and Short Papers)}, pages 4171--4186.

\bibitem[{Dhuliawala et~al.(2023)Dhuliawala, Komeili, Xu, Raileanu, Li, Celikyilmaz, and Weston}]{dhuliawala2023chain}
Shehzaad Dhuliawala, Mojtaba Komeili, Jing Xu, Roberta Raileanu, Xian Li, Asli Celikyilmaz, and Jason Weston. 2023.
\newblock Chain-of-verification reduces hallucination in large language models.
\newblock \emph{arXiv preprint arXiv:2309.11495}.

\bibitem[{Dong et~al.(2022)Dong, Li, Dai, Zheng, Wu, Chang, Sun, Xu, and Sui}]{dong2022survey}
Qingxiu Dong, Lei Li, Damai Dai, Ce~Zheng, Zhiyong Wu, Baobao Chang, Xu~Sun, Jingjing Xu, and Zhifang Sui. 2022.
\newblock A survey for in-context learning.
\newblock \emph{arXiv preprint arXiv:2301.00234}.

\bibitem[{Du et~al.(2022)Du, Raheja, Kumar, Kim, Lopez, and Kang}]{du2022understanding}
Wanyu Du, Vipul Raheja, Dhruv Kumar, Zae~Myung Kim, Melissa Lopez, and Dongyeop Kang. 2022.
\newblock Understanding iterative revision from human-written text.
\newblock \emph{arXiv preprint arXiv:2203.03802}.

\bibitem[{Du et~al.(2023)Du, Li, Torralba, Tenenbaum, and Mordatch}]{du2023improving}
Yilun Du, Shuang Li, Antonio Torralba, Joshua~B Tenenbaum, and Igor Mordatch. 2023.
\newblock Improving factuality and reasoning in language models through multiagent debate.
\newblock \emph{arXiv preprint arXiv:2305.14325}.

\bibitem[{Duan et~al.(2023)Duan, Cheng, Wang, Wang, Zavalny, Xu, Kailkhura, and Xu}]{duan2023shifting}
Jinhao Duan, Hao Cheng, Shiqi Wang, Chenan Wang, Alex Zavalny, Renjing Xu, Bhavya Kailkhura, and Kaidi Xu. 2023.
\newblock Shifting attention to relevance: Towards the uncertainty estimation of large language models.
\newblock \emph{arXiv preprint arXiv:2307.01379}.

\bibitem[{Durmus et~al.(2020)Durmus, He, and Diab}]{DBLP:conf/acl/DurmusHD20}
Esin Durmus, He~He, and Mona~T. Diab. 2020.
\newblock \href {https://doi.org/10.18653/v1/2020.acl-main.454} {{FEQA:} {A} question answering evaluation framework for faithfulness assessment in abstractive summarization}.
\newblock In \emph{Proceedings of the 58th Annual Meeting of the Association for Computational Linguistics, {ACL} 2020, Online, July 5-10, 2020}, pages 5055--5070. Association for Computational Linguistics.

\bibitem[{Dziri et~al.(2022)Dziri, Milton, Yu, Zaiane, and Reddy}]{dziri-etal-2022-origin}
Nouha Dziri, Sivan Milton, Mo~Yu, Osmar Zaiane, and Siva Reddy. 2022.
\newblock On the origin of hallucinations in conversational models: Is it the datasets or the models?
\newblock In \emph{Proceedings of the 2022 Conference of the North American Chapter of the Association for Computational Linguistics: Human Language Technologies}, pages 5271--5285.

\bibitem[{Dziri et~al.(2021)Dziri, Rashkin, Linzen, and Reitter}]{DBLP:journals/corr/abs-2105-00071}
Nouha Dziri, Hannah Rashkin, Tal Linzen, and David Reitter. 2021.
\newblock \href {http://arxiv.org/abs/2105.00071} {Evaluating groundedness in dialogue systems: The {BEGIN} benchmark}.
\newblock \emph{CoRR}, abs/2105.00071.

\bibitem[{Fabbri et~al.(2022)Fabbri, Choubey, Vig, Wu, and Xiong}]{fabbri-etal-2022-improving}
Alex Fabbri, Prafulla~Kumar Choubey, Jesse Vig, Chien-Sheng Wu, and Caiming Xiong. 2022.
\newblock Improving factual consistency in summarization with compression-based post-editing.
\newblock In \emph{Proceedings of the 2022 Conference on Empirical Methods in Natural Language Processing}, pages 9149--9156.

\bibitem[{Feng et~al.(2023)Feng, Zhang, and Fei}]{feng2023knowledge}
Chao Feng, Xinyu Zhang, and Zichu Fei. 2023.
\newblock Knowledge solver: Teaching llms to search for domain knowledge from knowledge graphs.
\newblock \emph{arXiv preprint arXiv:2309.03118}.

\bibitem[{Fernandes et~al.(2023)Fernandes, Madaan, Liu, Farinhas, Martins, Bertsch, de~Souza, Zhou, Wu, Neubig et~al.}]{fernandes2023bridging}
Patrick Fernandes, Aman Madaan, Emmy Liu, Ant{\'o}nio Farinhas, Pedro~Henrique Martins, Amanda Bertsch, Jos{\'e}~GC de~Souza, Shuyan Zhou, Tongshuang Wu, Graham Neubig, et~al. 2023.
\newblock Bridging the gap: A survey on integrating (human) feedback for natural language generation.
\newblock \emph{arXiv preprint arXiv:2305.00955}.

\bibitem[{Gao et~al.(2022)Gao, Schulman, and Hilton}]{gao2022scaling}
Leo Gao, John Schulman, and Jacob Hilton. 2022.
\newblock \href {http://arxiv.org/abs/2210.10760} {Scaling laws for reward model overoptimization}.

\bibitem[{Gao et~al.(2023{\natexlab{a}})Gao, Dai, Pasupat, Chen, Chaganty, Fan, Zhao, Lao, Lee, Juan et~al.}]{gao2023rarr}
Luyu Gao, Zhuyun Dai, Panupong Pasupat, Anthony Chen, Arun~Tejasvi Chaganty, Yicheng Fan, Vincent Zhao, Ni~Lao, Hongrae Lee, Da-Cheng Juan, et~al. 2023{\natexlab{a}}.
\newblock Rarr: Researching and revising what language models say, using language models.
\newblock In \emph{Proceedings of the 61st Annual Meeting of the Association for Computational Linguistics (Volume 1: Long Papers)}, pages 16477--16508.

\bibitem[{Gao et~al.(2023{\natexlab{b}})Gao, Yen, Yu, and Chen}]{gao2023enabling}
Tianyu Gao, Howard Yen, Jiatong Yu, and Danqi Chen. 2023{\natexlab{b}}.
\newblock Enabling large language models to generate text with citations.
\newblock \emph{arXiv preprint arXiv:2305.14627}.

\bibitem[{Gardent et~al.(2017)Gardent, Shimorina, Narayan, and Perez-Beltrachini}]{gardent-etal-2017-creating}
Claire Gardent, Anastasia Shimorina, Shashi Narayan, and Laura Perez-Beltrachini. 2017.
\newblock Creating training corpora for {NLG} micro-planners.
\newblock In \emph{Proceedings of the 55th Annual Meeting of the Association for Computational Linguistics (Volume 1: Long Papers)}, pages 179--188.

\bibitem[{Garrido-Mu{\~n}oz et~al.(2021)Garrido-Mu{\~n}oz, Montejo-R{\'a}ez, Mart{\'\i}nez-Santiago, and Ure{\~n}a-L{\'o}pez}]{garrido2021survey}
Ismael Garrido-Mu{\~n}oz, Arturo Montejo-R{\'a}ez, Fernando Mart{\'\i}nez-Santiago, and L~Alfonso Ure{\~n}a-L{\'o}pez. 2021.
\newblock A survey on bias in deep nlp.
\newblock \emph{Applied Sciences}, 11(7):3184.

\bibitem[{Gekhman et~al.(2024)Gekhman, Yona, Aharoni, Eyal, Feder, Reichart, and Herzig}]{gekhman2024does}
Zorik Gekhman, Gal Yona, Roee Aharoni, Matan Eyal, Amir Feder, Roi Reichart, and Jonathan Herzig. 2024.
\newblock \href {http://arxiv.org/abs/2405.05904} {Does fine-tuning llms on new knowledge encourage hallucinations?}

\bibitem[{Goldberg(2023)}]{goldberg2023}
Yoav Goldberg. 2023.
\newblock Reinforcement learning for language models.
\newblock \emph{Github Blog}.

\bibitem[{Gou et~al.(2023)Gou, Shao, Gong, Shen, Yang, Duan, and Chen}]{gou2023critic}
Zhibin Gou, Zhihong Shao, Yeyun Gong, Yelong Shen, Yujiu Yang, Nan Duan, and Weizhu Chen. 2023.
\newblock Critic: Large language models can self-correct with tool-interactive critiquing.
\newblock \emph{arXiv preprint arXiv:2305.11738}.

\bibitem[{Guerreiro et~al.(2023{\natexlab{a}})Guerreiro, Alves, Waldendorf, Haddow, Birch, Colombo, and Martins}]{guerreiro2023hallucinations}
Nuno~M Guerreiro, Duarte Alves, Jonas Waldendorf, Barry Haddow, Alexandra Birch, Pierre Colombo, and Andr{\'e}~FT Martins. 2023{\natexlab{a}}.
\newblock Hallucinations in large multilingual translation models.
\newblock \emph{arXiv preprint arXiv:2303.16104}.

\bibitem[{Guerreiro et~al.(2023{\natexlab{b}})Guerreiro, Voita, and Martins}]{Guerreiro2023mthalu1}
Nuno~Miguel Guerreiro, Elena Voita, and Andr{\'{e}} F.~T. Martins. 2023{\natexlab{b}}.
\newblock \href {https://aclanthology.org/2023.eacl-main.75} {Looking for a needle in a haystack: {A} comprehensive study of hallucinations in neural machine translation}.
\newblock In \emph{Proceedings of the 17th Conference of the European Chapter of the Association for Computational Linguistics, {EACL} 2023, Dubrovnik, Croatia, May 2-6, 2023}, pages 1059--1075. Association for Computational Linguistics.

\bibitem[{Gunjal et~al.(2023)Gunjal, Yin, and Bas}]{gunjal2023detecting}
Anisha Gunjal, Jihan Yin, and Erhan Bas. 2023.
\newblock Detecting and preventing hallucinations in large vision language models.
\newblock \emph{arXiv preprint arXiv:2308.06394}.

\bibitem[{Guo et~al.(2017)Guo, Pleiss, Sun, and Weinberger}]{guo2017calibration}
Chuan Guo, Geoff Pleiss, Yu~Sun, and Kilian~Q Weinberger. 2017.
\newblock On calibration of modern neural networks.
\newblock In \emph{International conference on machine learning}, pages 1321--1330. PMLR.

\bibitem[{Holtzman et~al.(2019)Holtzman, Buys, Du, Forbes, and Choi}]{holtzman2019curious}
Ari Holtzman, Jan Buys, Li~Du, Maxwell Forbes, and Yejin Choi. 2019.
\newblock The curious case of neural text degeneration.
\newblock In \emph{International Conference on Learning Representations}.

\bibitem[{Hu et~al.()Hu, Chen, Li, Guo, Wen, Philip, and Guo}]{hutowards}
Xuming Hu, Junzhe Chen, Xiaochuan Li, Yufei Guo, Lijie Wen, S~Yu Philip, and Zhijiang Guo.
\newblock Towards understanding factual knowledge of large language models.
\newblock In \emph{The Twelfth International Conference on Learning Representations}.

\bibitem[{Huang et~al.(2023{\natexlab{a}})Huang, Song, Wang, Chen, and Ma}]{huang2023look}
Yuheng Huang, Jiayang Song, Zhijie Wang, Huaming Chen, and Lei Ma. 2023{\natexlab{a}}.
\newblock Look before you leap: An exploratory study of uncertainty measurement for large language models.
\newblock \emph{arXiv preprint arXiv:2307.10236}.

\bibitem[{Huang et~al.(2023{\natexlab{b}})Huang, Shen, Zhang, Zhou, Rong, and Xiong}]{huang2023transformer}
Zeyu Huang, Yikang Shen, Xiaofeng Zhang, Jie Zhou, Wenge Rong, and Zhang Xiong. 2023{\natexlab{b}}.
\newblock Transformer-patcher: One mistake worth one neuron.
\newblock \emph{arXiv preprint arXiv:2301.09785}.

\bibitem[{Ji et~al.(2023)Ji, Lee, Frieske, Yu, Su, Xu, Ishii, Bang, Madotto, and Fung}]{nlg_hallucination_survey}
Ziwei Ji, Nayeon Lee, Rita Frieske, Tiezheng Yu, Dan Su, Yan Xu, Etsuko Ishii, Ye~Jin Bang, Andrea Madotto, and Pascale Fung. 2023.
\newblock \href {https://doi.org/10.1145/3571730} {Survey of hallucination in natural language generation}.
\newblock \emph{{ACM} Computing Surveys}, 55(12):1--38.

\bibitem[{Jiang et~al.(2021)Jiang, Araki, Ding, and Neubig}]{jiang2021can}
Zhengbao Jiang, Jun Araki, Haibo Ding, and Graham Neubig. 2021.
\newblock How can we know when language models know? on the calibration of language models for question answering.
\newblock \emph{Transactions of the Association for Computational Linguistics}, 9:962--977.

\bibitem[{Kadavath et~al.(2022)Kadavath, Conerly, Askell, Henighan, Drain, Perez, Schiefer, Dodds, DasSarma, Tran-Johnson et~al.}]{kadavath2022language}
Saurav Kadavath, Tom Conerly, Amanda Askell, Tom Henighan, Dawn Drain, Ethan Perez, Nicholas Schiefer, Zac~Hatfield Dodds, Nova DasSarma, Eli Tran-Johnson, et~al. 2022.
\newblock Language models (mostly) know what they know.
\newblock \emph{arXiv preprint arXiv:2207.05221}.

\bibitem[{Kaddour et~al.(2023)Kaddour, Harris, Mozes, Bradley, Raileanu, and McHardy}]{kaddour2023challenges}
Jean Kaddour, Joshua Harris, Maximilian Mozes, Herbie Bradley, Roberta Raileanu, and Robert McHardy. 2023.
\newblock Challenges and applications of large language models.
\newblock \emph{arXiv preprint arXiv:2307.10169}.

\bibitem[{Kim et~al.(2024)Kim, Kim, Park, Kim, Park, Yoo, goo Lee, and Kim}]{kim2024aligning}
Hyuhng~Joon Kim, Youna Kim, Cheonbok Park, Junyeob Kim, Choonghyun Park, Kang~Min Yoo, Sang goo Lee, and Taeuk Kim. 2024.
\newblock \href {http://arxiv.org/abs/2404.11972} {Aligning language models to explicitly handle ambiguity}.

\bibitem[{K{\"o}pf et~al.(2023)K{\"o}pf, Kilcher, von R{\"u}tte, Anagnostidis, Tam, Stevens, Barhoum, Duc, Stanley, Nagyfi et~al.}]{kopf2023openassistant}
Andreas K{\"o}pf, Yannic Kilcher, Dimitri von R{\"u}tte, Sotiris Anagnostidis, Zhi-Rui Tam, Keith Stevens, Abdullah Barhoum, Nguyen~Minh Duc, Oliver Stanley, Rich{\'a}rd Nagyfi, et~al. 2023.
\newblock Openassistant conversations--democratizing large language model alignment.
\newblock \emph{arXiv preprint arXiv:2304.07327}.

\bibitem[{Kryscinski et~al.(2020)Kryscinski, McCann, Xiong, and Socher}]{Kryscinski2020sumhalu1}
Wojciech Kryscinski, Bryan McCann, Caiming Xiong, and Richard Socher. 2020.
\newblock \href {https://doi.org/10.18653/v1/2020.emnlp-main.750} {Evaluating the factual consistency of abstractive text summarization}.
\newblock In \emph{Proceedings of the 2020 Conference on Empirical Methods in Natural Language Processing, {EMNLP} 2020, Online, November 16-20, 2020}, pages 9332--9346. Association for Computational Linguistics.

\bibitem[{Lai et~al.(2023)Lai, Ngo, Veyseh, Man, Dernoncourt, Bui, and Nguyen}]{lai2023chatgpt}
Viet~Dac Lai, Nghia~Trung Ngo, Amir Pouran~Ben Veyseh, Hieu Man, Franck Dernoncourt, Trung Bui, and Thien~Huu Nguyen. 2023.
\newblock Chatgpt beyond english: Towards a comprehensive evaluation of large language models in multilingual learning.
\newblock \emph{arXiv preprint arXiv:2304.05613}.

\bibitem[{Lan et~al.(2019)Lan, Chen, Goodman, Gimpel, Sharma, and Soricut}]{lan2019albert}
Zhenzhong Lan, Mingda Chen, Sebastian Goodman, Kevin Gimpel, Piyush Sharma, and Radu Soricut. 2019.
\newblock Albert: A lite bert for self-supervised learning of language representations.
\newblock In \emph{International Conference on Learning Representations}.

\bibitem[{Lanham et~al.(2023)Lanham, Chen, Radhakrishnan, Steiner, Denison, Hernandez, Li, Durmus, Hubinger, Kernion et~al.}]{lanham2023measuring}
Tamera Lanham, Anna Chen, Ansh Radhakrishnan, Benoit Steiner, Carson Denison, Danny Hernandez, Dustin Li, Esin Durmus, Evan Hubinger, Jackson Kernion, et~al. 2023.
\newblock Measuring faithfulness in chain-of-thought reasoning.
\newblock \emph{arXiv preprint arXiv:2307.13702}.

\bibitem[{Lazaridou et~al.(2022)Lazaridou, Gribovskaya, Stokowiec, and Grigorev}]{lazaridou2022internet}
Angeliki Lazaridou, Elena Gribovskaya, Wojciech Stokowiec, and Nikolai Grigorev. 2022.
\newblock Internet-augmented language models through few-shot prompting for open-domain question answering.
\newblock \emph{arXiv preprint arXiv:2203.05115}.

\bibitem[{Lee et~al.(2023)Lee, Hunter, and Ruiz}]{lee2023platypus}
Ariel~N Lee, Cole~J Hunter, and Nataniel Ruiz. 2023.
\newblock Platypus: Quick, cheap, and powerful refinement of llms.
\newblock \emph{arXiv preprint arXiv:2308.07317}.

\bibitem[{Lee et~al.(2019)Lee, Firat, Agarwal, Fannjiang, and Sussillo}]{hallucination-mt}
Katherine Lee, Orhan Firat, Ashish Agarwal, Clara Fannjiang, and David Sussillo. 2019.
\newblock \href {https://openreview.net/forum?id=SkxJ-309FQ} {Hallucinations in neural machine translation}.

\bibitem[{Lee et~al.(2022)Lee, Ping, Xu, Patwary, Fung, Shoeybi, and Catanzaro}]{lee2022factuality}
Nayeon Lee, Wei Ping, Peng Xu, Mostofa Patwary, Pascale~N Fung, Mohammad Shoeybi, and Bryan Catanzaro. 2022.
\newblock Factuality enhanced language models for open-ended text generation.
\newblock \emph{Advances in Neural Information Processing Systems}, 35:34586--34599.

\bibitem[{Lewis et~al.(2020{\natexlab{a}})Lewis, Liu, Goyal, Ghazvininejad, Mohamed, Levy, Stoyanov, and Zettlemoyer}]{lewis2020bart}
Mike Lewis, Yinhan Liu, Naman Goyal, Marjan Ghazvininejad, Abdelrahman Mohamed, Omer Levy, Veselin Stoyanov, and Luke Zettlemoyer. 2020{\natexlab{a}}.
\newblock Bart: Denoising sequence-to-sequence pre-training for natural language generation, translation, and comprehension.
\newblock In \emph{Proceedings of the 58th Annual Meeting of the Association for Computational Linguistics}, pages 7871--7880.

\bibitem[{Lewis et~al.(2020{\natexlab{b}})Lewis, Perez, Piktus, Petroni, Karpukhin, Goyal, K{\"u}ttler, Lewis, Yih, Rockt{\"a}schel et~al.}]{lewis2020retrieval}
Patrick Lewis, Ethan Perez, Aleksandra Piktus, Fabio Petroni, Vladimir Karpukhin, Naman Goyal, Heinrich K{\"u}ttler, Mike Lewis, Wen-tau Yih, Tim Rockt{\"a}schel, et~al. 2020{\natexlab{b}}.
\newblock Retrieval-augmented generation for knowledge-intensive nlp tasks.
\newblock \emph{Advances in Neural Information Processing Systems}, 33:9459--9474.

\bibitem[{Li et~al.(2022{\natexlab{a}})Li, Su, Cai, Wang, and Liu}]{li2022survey}
Huayang Li, Yixuan Su, Deng Cai, Yan Wang, and Lemao Liu. 2022{\natexlab{a}}.
\newblock A survey on retrieval-augmented text generation.
\newblock \emph{arXiv preprint arXiv:2202.01110}.

\bibitem[{Li et~al.(2024)Li, Chen, Ren, Cheng, Zhao, Nie, and Wen}]{li2024dawn}
Junyi Li, Jie Chen, Ruiyang Ren, Xiaoxue Cheng, Wayne~Xin Zhao, Jian-Yun Nie, and Ji-Rong Wen. 2024.
\newblock The dawn after the dark: An empirical study on factuality hallucination in large language models.
\newblock \emph{arXiv preprint arXiv:2401.03205}.

\bibitem[{Li et~al.(2023{\natexlab{a}})Li, Cheng, Zhao, Nie, and Wen}]{li2023halueval}
Junyi Li, Xiaoxue Cheng, Wayne~Xin Zhao, Jian-Yun Nie, and Ji-Rong Wen. 2023{\natexlab{a}}.
\newblock Halueval: A large-scale hallucination evaluation benchmark for large language models.
\newblock \emph{arXiv preprint arXiv:2305.11747}.

\bibitem[{Li et~al.(2022{\natexlab{b}})Li, Tang, Zhao, Nie, and Wen}]{li2022pretrained}
Junyi Li, Tianyi Tang, Wayne~Xin Zhao, Jian-Yun Nie, and Ji-Rong Wen. 2022{\natexlab{b}}.
\newblock Pretrained language models for text generation: A survey.
\newblock \emph{arXiv preprint arXiv:2201.05273}.

\bibitem[{Li et~al.(2023{\natexlab{b}})Li, Patel, Vi{\'e}gas, Pfister, and Wattenberg}]{li2023inference}
Kenneth Li, Oam Patel, Fernanda Vi{\'e}gas, Hanspeter Pfister, and Martin Wattenberg. 2023{\natexlab{b}}.
\newblock Inference-time intervention: Eliciting truthful answers from a language model.
\newblock \emph{arXiv preprint arXiv:2306.03341}.

\bibitem[{Li et~al.(2023{\natexlab{c}})Li, Peng, and Zhang}]{li2023self}
Miaoran Li, Baolin Peng, and Zhu Zhang. 2023{\natexlab{c}}.
\newblock Self-checker: Plug-and-play modules for fact-checking with large language models.
\newblock \emph{arXiv preprint arXiv:2305.14623}.

\bibitem[{Li et~al.(2022{\natexlab{c}})Li, Li, Shang, Dong, Sun, Liu, Ji, Jiang, and Liu}]{li-etal-2022-pre}
Shaobo Li, Xiaoguang Li, Lifeng Shang, Zhenhua Dong, Chengjie Sun, Bingquan Liu, Zhenzhou Ji, Xin Jiang, and Qun Liu. 2022{\natexlab{c}}.
\newblock How pre-trained language models capture factual knowledge? a causal-inspired analysis.
\newblock In \emph{Findings of the Association for Computational Linguistics: ACL 2022}, pages 1720--1732.

\bibitem[{Li et~al.(2023{\natexlab{d}})Li, Zhao, Chia, Ding, Bing, Joty, and Poria}]{li2023chain}
Xingxuan Li, Ruochen Zhao, Yew~Ken Chia, Bosheng Ding, Lidong Bing, Shafiq Joty, and Soujanya Poria. 2023{\natexlab{d}}.
\newblock Chain of knowledge: A framework for grounding large language models with structured knowledge bases.
\newblock \emph{arXiv preprint arXiv:2305.13269}.

\bibitem[{Li et~al.(2023{\natexlab{e}})Li, Du, Zhou, Wang, Zhao, and Wen}]{li2023evaluating}
Yifan Li, Yifan Du, Kun Zhou, Jinpeng Wang, Wayne~Xin Zhao, and Ji-Rong Wen. 2023{\natexlab{e}}.
\newblock Evaluating object hallucination in large vision-language models.
\newblock \emph{arXiv preprint arXiv:2305.10355}.

\bibitem[{Li et~al.(2023{\natexlab{f}})Li, Bubeck, Eldan, Del~Giorno, Gunasekar, and Lee}]{li2023textbooks}
Yuanzhi Li, S{\'e}bastien Bubeck, Ronen Eldan, Allie Del~Giorno, Suriya Gunasekar, and Yin~Tat Lee. 2023{\natexlab{f}}.
\newblock Textbooks are all you need ii: phi-1.5 technical report.
\newblock \emph{arXiv preprint arXiv:2309.05463}.

\bibitem[{Li et~al.(2023{\natexlab{g}})Li, Zhang, Zhao, Yang, and Yang}]{li2023batgpt}
Zuchao Li, Shitou Zhang, Hai Zhao, Yifei Yang, and Dongjie Yang. 2023{\natexlab{g}}.
\newblock Batgpt: A bidirectional autoregessive talker from generative pre-trained transformer.
\newblock \emph{arXiv preprint arXiv:2307.00360}.

\bibitem[{Lightman et~al.(2023)Lightman, Kosaraju, Burda, Edwards, Baker, Lee, Leike, Schulman, Sutskever, and Cobbe}]{lightman2023let}
Hunter Lightman, Vineet Kosaraju, Yura Burda, Harri Edwards, Bowen Baker, Teddy Lee, Jan Leike, John Schulman, Ilya Sutskever, and Karl Cobbe. 2023.
\newblock Let's verify step by step.
\newblock \emph{arXiv preprint arXiv:2305.20050}.

\bibitem[{Lin(2004)}]{lin2004rouge}
Chin-Yew Lin. 2004.
\newblock Rouge: A package for automatic evaluation of summaries.
\newblock In \emph{Text summarization branches out}, pages 74--81.

\bibitem[{Lin et~al.(2021)Lin, Hilton, and Evans}]{lin2021truthfulqa}
Stephanie Lin, Jacob Hilton, and Owain Evans. 2021.
\newblock Truthfulqa: Measuring how models mimic human falsehoods.
\newblock \emph{arXiv preprint arXiv:2109.07958}.

\bibitem[{Lin et~al.(2023)Lin, Trivedi, and Sun}]{lin2023generating}
Zhen Lin, Shubhendu Trivedi, and Jimeng Sun. 2023.
\newblock Generating with confidence: Uncertainty quantification for black-box large language models.
\newblock \emph{arXiv preprint arXiv:2305.19187}.

\bibitem[{Liska et~al.(2022)Liska, Kocisky, Gribovskaya, Terzi, Sezener, Agrawal, Cyprien De~Masson, Scholtes, Zaheer, Young et~al.}]{liska2022streamingqa}
Adam Liska, Tomas Kocisky, Elena Gribovskaya, Tayfun Terzi, Eren Sezener, Devang Agrawal, D’Autume Cyprien De~Masson, Tim Scholtes, Manzil Zaheer, Susannah Young, et~al. 2022.
\newblock Streamingqa: A benchmark for adaptation to new knowledge over time in question answering models.
\newblock In \emph{International Conference on Machine Learning}, pages 13604--13622. PMLR.

\bibitem[{Liu et~al.(2023{\natexlab{a}})Liu, Lin, Li, Wang, Yacoob, and Wang}]{liu2023aligning}
Fuxiao Liu, Kevin Lin, Linjie Li, Jianfeng Wang, Yaser Yacoob, and Lijuan Wang. 2023{\natexlab{a}}.
\newblock Aligning large multi-modal model with robust instruction tuning.
\newblock \emph{arXiv preprint arXiv:2306.14565}.

\bibitem[{Liu et~al.(2023{\natexlab{b}})Liu, Li, Wu, and Lee}]{liu2023visual}
Haotian Liu, Chunyuan Li, Qingyang Wu, and Yong~Jae Lee. 2023{\natexlab{b}}.
\newblock Visual instruction tuning.
\newblock \emph{arXiv preprint arXiv:2304.08485}.

\bibitem[{Liu(2022)}]{Liu_LlamaIndex_2022}
Jerry Liu. 2022.
\newblock \href {https://doi.org/10.5281/zenodo.1234} {{LlamaIndex}}.

\bibitem[{Liu et~al.(2023{\natexlab{c}})Liu, Jin, Wang, Cheng, Dou, and Wen}]{liu2023reta}
Jiongnan Liu, Jiajie Jin, Zihan Wang, Jiehan Cheng, Zhicheng Dou, and Ji-Rong Wen. 2023{\natexlab{c}}.
\newblock Reta-llm: A retrieval-augmented large language model toolkit.
\newblock \emph{arXiv preprint arXiv:2306.05212}.

\bibitem[{Liu et~al.(2023{\natexlab{d}})Liu, Lin, Hewitt, Paranjape, Bevilacqua, Petroni, and Liang}]{liu2023lost}
Nelson~F Liu, Kevin Lin, John Hewitt, Ashwin Paranjape, Michele Bevilacqua, Fabio Petroni, and Percy Liang. 2023{\natexlab{d}}.
\newblock Lost in the middle: How language models use long contexts.
\newblock \emph{arXiv preprint arXiv:2307.03172}.

\bibitem[{Liu et~al.(2022)Liu, Zhang, Brockett, Mao, Sui, Chen, and Dolan}]{liu-etal-2022-token}
Tianyu Liu, Yizhe Zhang, Chris Brockett, Yi~Mao, Zhifang Sui, Weizhu Chen, and Bill Dolan. 2022.
\newblock A token-level reference-free hallucination detection benchmark for free-form text generation.
\newblock In \emph{Proceedings of the 60th Annual Meeting of the Association for Computational Linguistics (Volume 1: Long Papers)}, pages 6723--6737.

\bibitem[{Liu et~al.(2019)Liu, Ott, Goyal, Du, Joshi, Chen, Levy, Lewis, Zettlemoyer, and Stoyanov}]{liu2019roberta}
Yinhan Liu, Myle Ott, Naman Goyal, Jingfei Du, Mandar Joshi, Danqi Chen, Omer Levy, Mike Lewis, Luke Zettlemoyer, and Veselin Stoyanov. 2019.
\newblock Roberta: A robustly optimized bert pretraining approach.
\newblock \emph{arXiv preprint arXiv:1907.11692}.

\bibitem[{Llama~Team(2024)}]{Llama31}
AI~@~Meta Llama~Team. 2024.
\newblock \href {https://scontent-nrt1-2.xx.fbcdn.net/v/t39.2365-6/453304228_1160109801904614_7143520450792086005_n.pdf?_nc_cat=108&ccb=1-7&_nc_sid=3c67a6&_nc_ohc=22QmL5CfN0AQ7kNvgFb_ts2&_nc_ht=scontent-nrt1-2.xx&oh=00_AYBm3hZtMT8nvua0DyT5eE500hiVthYh1ROfu6t-eW5hPQ&oe=66D78187} {Llama 3.1 technical report}.

\bibitem[{Luo et~al.(2023{\natexlab{a}})Luo, Xiao, and Ma}]{luo2023zero}
Junyu Luo, Cao Xiao, and Fenglong Ma. 2023{\natexlab{a}}.
\newblock Zero-resource hallucination prevention for large language models.
\newblock \emph{arXiv preprint arXiv:2309.02654}.

\bibitem[{Luo et~al.(2023{\natexlab{b}})Luo, Xie, and Ananiadou}]{luo2023chatgpt}
Zheheng Luo, Qianqian Xie, and Sophia Ananiadou. 2023{\natexlab{b}}.
\newblock Chatgpt as a factual inconsistency evaluator for abstractive text summarization.
\newblock \emph{arXiv preprint arXiv:2303.15621}.

\bibitem[{Luo et~al.(2023{\natexlab{c}})Luo, Xu, Zhao, Geng, Tao, Ma, Lin, and Jiang}]{luo2023augmented}
Ziyang Luo, Can Xu, Pu~Zhao, Xiubo Geng, Chongyang Tao, Jing Ma, Qingwei Lin, and Daxin Jiang. 2023{\natexlab{c}}.
\newblock Augmented large language models with parametric knowledge guiding.
\newblock \emph{arXiv preprint arXiv:2305.04757}.

\bibitem[{Luu et~al.(2022)Luu, Khashabi, Gururangan, Mandyam, and Smith}]{luu2022time}
Kelvin Luu, Daniel Khashabi, Suchin Gururangan, Karishma Mandyam, and Noah~A Smith. 2022.
\newblock Time waits for no one! analysis and challenges of temporal misalignment.
\newblock In \emph{Proceedings of the 2022 Conference of the North American Chapter of the Association for Computational Linguistics: Human Language Technologies}, pages 5944--5958.

\bibitem[{Maaz et~al.(2023)Maaz, Rasheed, Khan, and Khan}]{maaz2023video}
Muhammad Maaz, Hanoona Rasheed, Salman Khan, and Fahad~Shahbaz Khan. 2023.
\newblock Video-chatgpt: Towards detailed video understanding via large vision and language models.
\newblock \emph{arXiv preprint arXiv:2306.05424}.

\bibitem[{Magesh et~al.(2024)Magesh, Surani, Dahl, Suzgun, Manning, and Ho}]{magesh2024hallucination}
Varun Magesh, Faiz Surani, Matthew Dahl, Mirac Suzgun, Christopher~D Manning, and Daniel~E Ho. 2024.
\newblock Hallucination-free? assessing the reliability of leading ai legal research tools.
\newblock \emph{arXiv preprint arXiv:2405.20362}.

\bibitem[{Mallen et~al.(2023)Mallen, Asai, Zhong, Das, Khashabi, and Hajishirzi}]{mallen2023not}
Alex Mallen, Akari Asai, Victor Zhong, Rajarshi Das, Daniel Khashabi, and Hannaneh Hajishirzi. 2023.
\newblock When not to trust language models: Investigating effectiveness of parametric and non-parametric memories.
\newblock In \emph{Proceedings of the 61st Annual Meeting of the Association for Computational Linguistics (Volume 1: Long Papers)}, pages 9802--9822.

\bibitem[{Manakul et~al.(2023)Manakul, Liusie, and Gales}]{manakul2023selfcheckgpt}
Potsawee Manakul, Adian Liusie, and Mark~JF Gales. 2023.
\newblock Selfcheckgpt: Zero-resource black-box hallucination detection for generative large language models.
\newblock \emph{arXiv preprint arXiv:2303.08896}.

\bibitem[{Maynez et~al.(2020)Maynez, Narayan, Bohnet, and McDonald}]{Maynez2020sumhalu2}
Joshua Maynez, Shashi Narayan, Bernd Bohnet, and Ryan~T. McDonald. 2020.
\newblock \href {https://doi.org/10.18653/v1/2020.acl-main.173} {On faithfulness and factuality in abstractive summarization}.
\newblock In \emph{Proceedings of the 58th Annual Meeting of the Association for Computational Linguistics, {ACL} 2020, Online, July 5-10, 2020}, pages 1906--1919. Association for Computational Linguistics.

\bibitem[{McKenna et~al.(2023)McKenna, Li, Cheng, Hosseini, Johnson, and Steedman}]{mckenna2023sources}
Nick McKenna, Tianyi Li, Liang Cheng, Mohammad~Javad Hosseini, Mark Johnson, and Mark Steedman. 2023.
\newblock Sources of hallucination by large language models on inference tasks.
\newblock \emph{arXiv preprint arXiv:2305.14552}.

\bibitem[{Meng et~al.(2022{\natexlab{a}})Meng, Bau, Andonian, and Belinkov}]{meng2022locating}
Kevin Meng, David Bau, Alex Andonian, and Yonatan Belinkov. 2022{\natexlab{a}}.
\newblock Locating and editing factual associations in gpt.
\newblock \emph{Advances in Neural Information Processing Systems}, 35:17359--17372.

\bibitem[{Meng et~al.(2022{\natexlab{b}})Meng, Sharma, Andonian, Belinkov, and Bau}]{meng2022mass}
Kevin Meng, Arnab~Sen Sharma, Alex Andonian, Yonatan Belinkov, and David Bau. 2022{\natexlab{b}}.
\newblock Mass-editing memory in a transformer.
\newblock \emph{arXiv preprint arXiv:2210.07229}.

\bibitem[{Mialon et~al.(2023)Mialon, Dess{\`\i}, Lomeli, Nalmpantis, Pasunuru, Raileanu, Rozi{\`e}re, Schick, Dwivedi-Yu, Celikyilmaz et~al.}]{mialon2023augmented}
Gr{\'e}goire Mialon, Roberto Dess{\`\i}, Maria Lomeli, Christoforos Nalmpantis, Ram Pasunuru, Roberta Raileanu, Baptiste Rozi{\`e}re, Timo Schick, Jane Dwivedi-Yu, Asli Celikyilmaz, et~al. 2023.
\newblock Augmented language models: a survey.
\newblock \emph{arXiv preprint arXiv:2302.07842}.

\bibitem[{Mikolov et~al.(2010)Mikolov, Karafi{\'a}t, Burget, Cernock{\`y}, and Khudanpur}]{mikolov2010recurrent}
Tomas Mikolov, Martin Karafi{\'a}t, Lukas Burget, Jan Cernock{\`y}, and Sanjeev Khudanpur. 2010.
\newblock Recurrent neural network based language model.
\newblock In \emph{Interspeech}. Makuhari.

\bibitem[{Min et~al.(2021)Min, Ross, Sulem, Veyseh, Nguyen, Sainz, Agirre, Heintz, and Roth}]{min2021recent}
Bonan Min, Hayley Ross, Elior Sulem, Amir Pouran~Ben Veyseh, Thien~Huu Nguyen, Oscar Sainz, Eneko Agirre, Ilana Heintz, and Dan Roth. 2021.
\newblock Recent advances in natural language processing via large pre-trained language models: A survey.
\newblock \emph{ACM Computing Surveys}.

\bibitem[{Min et~al.(2023)Min, Krishna, Lyu, Lewis, Yih, Koh, Iyyer, Zettlemoyer, and Hajishirzi}]{min2023factscore}
Sewon Min, Kalpesh Krishna, Xinxi Lyu, Mike Lewis, Wen-tau Yih, Pang~Wei Koh, Mohit Iyyer, Luke Zettlemoyer, and Hannaneh Hajishirzi. 2023.
\newblock Factscore: Fine-grained atomic evaluation of factual precision in long form text generation.
\newblock \emph{arXiv preprint arXiv:2305.14251}.

\bibitem[{Mitchell et~al.(2022)Mitchell, Lin, Bosselut, Manning, and Finn}]{mitchell2022memory}
Eric Mitchell, Charles Lin, Antoine Bosselut, Christopher~D Manning, and Chelsea Finn. 2022.
\newblock Memory-based model editing at scale.
\newblock In \emph{International Conference on Machine Learning}, pages 15817--15831. PMLR.

\bibitem[{Mohamed et~al.(2023)Mohamed, Mengyin, Jacob, Xueying, Yu, and Shizhu}]{elaraby2023halo}
Elaraby Mohamed, Lu~Mengyin, Dunn Jacob, Zhang Xueying, Wang Yu, and Liu Shizhu. 2023.
\newblock Halo: Estimation and reduction of hallucinations in open-source weak large language models.
\newblock \emph{arXiv preprint arXiv:2308.11764}.

\bibitem[{Muhlgay et~al.(2023)Muhlgay, Ram, Magar, Levine, Ratner, Belinkov, Abend, Leyton-Brown, Shashua, and Shoham}]{muhlgay2023generating}
Dor Muhlgay, Ori Ram, Inbal Magar, Yoav Levine, Nir Ratner, Yonatan Belinkov, Omri Abend, Kevin Leyton-Brown, Amnon Shashua, and Yoav Shoham. 2023.
\newblock Generating benchmarks for factuality evaluation of language models.
\newblock \emph{arXiv preprint arXiv:2307.06908}.

\bibitem[{M{\"u}ndler et~al.(2023)M{\"u}ndler, He, Jenko, and Vechev}]{mundler2023self}
Niels M{\"u}ndler, Jingxuan He, Slobodan Jenko, and Martin Vechev. 2023.
\newblock Self-contradictory hallucinations of large language models: Evaluation, detection and mitigation.
\newblock \emph{arXiv preprint arXiv:2305.15852}.

\bibitem[{Murty et~al.(2022)Murty, Manning, Lundberg, and Ribeiro}]{murty-etal-2022-fixing}
Shikhar Murty, Christopher Manning, Scott Lundberg, and Marco~Tulio Ribeiro. 2022.
\newblock Fixing model bugs with natural language patches.
\newblock In \emph{Proceedings of the 2022 Conference on Empirical Methods in Natural Language Processing}, pages 11600--11613.

\bibitem[{Nakano et~al.(2021)Nakano, Hilton, Balaji, Wu, Ouyang, Kim, Hesse, Jain, Kosaraju, Saunders et~al.}]{nakano2021webgpt}
Reiichiro Nakano, Jacob Hilton, Suchir Balaji, Jeff Wu, Long Ouyang, Christina Kim, Christopher Hesse, Shantanu Jain, Vineet Kosaraju, William Saunders, et~al. 2021.
\newblock Webgpt: Browser-assisted question-answering with human feedback.
\newblock \emph{arXiv preprint arXiv:2112.09332}.

\bibitem[{Nallapati et~al.(2017)Nallapati, Zhai, and Zhou}]{nallapati2017summarunner}
Ramesh Nallapati, Feifei Zhai, and Bowen Zhou. 2017.
\newblock Summarunner: A recurrent neural network based sequence model for extractive summarization of documents.
\newblock In \emph{Proceedings of the AAAI conference on artificial intelligence}.

\bibitem[{Napoles et~al.(2017)Napoles, Sakaguchi, and Tetreault}]{napoles2017jfleg}
Courtney Napoles, Keisuke Sakaguchi, and Joel Tetreault. 2017.
\newblock Jfleg: A fluency corpus and benchmark for grammatical error correction.
\newblock In \emph{Proceedings of the 15th Conference of the European Chapter of the Association for Computational Linguistics: Volume 2, Short Papers}, pages 229--234.

\bibitem[{Navigli et~al.(2023)Navigli, Conia, and Ross}]{llm_bias}
Roberto Navigli, Simone Conia, and Bj{\"o}rn Ross. 2023.
\newblock Biases in large language models: Origins, inventory and discussion.
\newblock \emph{ACM Journal of Data and Information Quality}.

\bibitem[{Ni et~al.(2022)Ni, Qu, Lu, Dai, {\'{A}}brego, Ma, Zhao, Luan, Hall, Chang, and Yang}]{DBLP:conf/emnlp/Ni0LDAMZLHCY22}
Jianmo Ni, Chen Qu, Jing Lu, Zhuyun Dai, Gustavo~Hern{\'{a}}ndez {\'{A}}brego, Ji~Ma, Vincent~Y. Zhao, Yi~Luan, Keith~B. Hall, Ming{-}Wei Chang, and Yinfei Yang. 2022.
\newblock \href {https://doi.org/10.18653/v1/2022.emnlp-main.669} {Large dual encoders are generalizable retrievers}.
\newblock In \emph{Proceedings of the 2022 Conference on Empirical Methods in Natural Language Processing, {EMNLP} 2022, Abu Dhabi, United Arab Emirates, December 7-11, 2022}, pages 9844--9855. Association for Computational Linguistics.

\bibitem[{Ning et~al.(2023)Ning, Lin, Zhou, Yang, and Wang}]{ning2023skeleton}
Xuefei Ning, Zinan Lin, Zixuan Zhou, Huazhong Yang, and Yu~Wang. 2023.
\newblock Skeleton-of-thought: Large language models can do parallel decoding.
\newblock \emph{arXiv preprint arXiv:2307.15337}.

\bibitem[{Onoe et~al.(2023)Onoe, Zhang, Padmanabhan, Durrett, and Choi}]{onoe2023can}
Yasumasa Onoe, Michael~JQ Zhang, Shankar Padmanabhan, Greg Durrett, and Eunsol Choi. 2023.
\newblock Can lms learn new entities from descriptions? challenges in propagating injected knowledge.
\newblock \emph{arXiv preprint arXiv:2305.01651}.

\bibitem[{OpenAI(2023{\natexlab{a}})}]{chatgpt}
OpenAI. 2023{\natexlab{a}}.
\newblock {ChatGPT}.
\newblock \url{https://openai.com/blog/chatgpt}.

\bibitem[{OpenAI(2023{\natexlab{b}})}]{openai2023gpt4}
OpenAI. 2023{\natexlab{b}}.
\newblock Gpt-4 technical report.
\newblock \emph{arXiv preprint arXiv:2303.08774}.

\bibitem[{Ouyang et~al.(2022)Ouyang, Wu, Jiang, Almeida, Wainwright, Mishkin, Zhang, Agarwal, Slama, Ray et~al.}]{ouyang2022training}
Long Ouyang, Jeffrey Wu, Xu~Jiang, Diogo Almeida, Carroll Wainwright, Pamela Mishkin, Chong Zhang, Sandhini Agarwal, Katarina Slama, Alex Ray, et~al. 2022.
\newblock Training language models to follow instructions with human feedback.
\newblock \emph{Advances in Neural Information Processing Systems}, 35:27730--27744.

\bibitem[{Pal et~al.(2023)Pal, Umapathi, and Sankarasubbu}]{pal2023med}
Ankit Pal, Logesh~Kumar Umapathi, and Malaikannan Sankarasubbu. 2023.
\newblock Med-halt: Medical domain hallucination test for large language models.
\newblock \emph{arXiv preprint arXiv:2307.15343}.

\bibitem[{Papineni et~al.(2002)Papineni, Roukos, Ward, and Zhu}]{bleu}
Kishore Papineni, Salim Roukos, Todd Ward, and Wei-Jing Zhu. 2002.
\newblock \href {https://doi.org/10.3115/1073083.1073135} {{B}leu: a method for automatic evaluation of machine translation}.
\newblock In \emph{Proceedings of the 40th Annual Meeting of the Association for Computational Linguistics}, pages 311--318, Philadelphia, Pennsylvania, USA. Association for Computational Linguistics.

\bibitem[{Parikh et~al.(2020)Parikh, Wang, Gehrmann, Faruqui, Dhingra, Yang, and Das}]{parikh-etal-2020-totto}
Ankur Parikh, Xuezhi Wang, Sebastian Gehrmann, Manaal Faruqui, Bhuwan Dhingra, Diyi Yang, and Dipanjan Das. 2020.
\newblock {ToTTo}: A controlled table-to-text generation dataset.
\newblock In \emph{Proceedings of the 2020 Conference on Empirical Methods in Natural Language Processing (EMNLP)}, pages 1173--1186.

\bibitem[{Park et~al.(2023)Park, O'Brien, Cai, Morris, Liang, and Bernstein}]{park2023generative}
Joon~Sung Park, Joseph~C O'Brien, Carrie~J Cai, Meredith~Ringel Morris, Percy Liang, and Michael~S Bernstein. 2023.
\newblock Generative agents: Interactive simulacra of human behavior.
\newblock \emph{arXiv preprint arXiv:2304.03442}.

\bibitem[{Pauls and Klein(2011)}]{pauls2011faster}
Adam Pauls and Dan Klein. 2011.
\newblock Faster and smaller n-gram language models.
\newblock In \emph{Proceedings of the 49th annual meeting of the Association for Computational Linguistics: Human Language Technologies}, pages 258--267.

\bibitem[{Penedo et~al.(2023)Penedo, Malartic, Hesslow, Cojocaru, Cappelli, Alobeidli, Pannier, Almazrouei, and Launay}]{penedo2023refinedweb}
Guilherme Penedo, Quentin Malartic, Daniel Hesslow, Ruxandra Cojocaru, Alessandro Cappelli, Hamza Alobeidli, Baptiste Pannier, Ebtesam Almazrouei, and Julien Launay. 2023.
\newblock The refinedweb dataset for falcon llm: outperforming curated corpora with web data, and web data only.
\newblock \emph{arXiv preprint arXiv:2306.01116}.

\bibitem[{Peng et~al.(2023{\natexlab{a}})Peng, Galley, He, Cheng, Xie, Hu, Huang, Liden, Yu, Chen et~al.}]{peng2023check}
Baolin Peng, Michel Galley, Pengcheng He, Hao Cheng, Yujia Xie, Yu~Hu, Qiuyuan Huang, Lars Liden, Zhou Yu, Weizhu Chen, et~al. 2023{\natexlab{a}}.
\newblock Check your facts and try again: Improving large language models with external knowledge and automated feedback.
\newblock \emph{arXiv preprint arXiv:2302.12813}.

\bibitem[{Peng et~al.(2023{\natexlab{b}})Peng, Li, He, Galley, and Gao}]{peng2023instruction}
Baolin Peng, Chunyuan Li, Pengcheng He, Michel Galley, and Jianfeng Gao. 2023{\natexlab{b}}.
\newblock Instruction tuning with gpt-4.
\newblock \emph{arXiv preprint arXiv:2304.03277}.

\bibitem[{Perez et~al.(2022)Perez, Ringer, Luko{\v{s}}i{\=u}t{\.e}, Nguyen, Chen, Heiner, Pettit, Olsson, Kundu, Kadavath et~al.}]{perez2022discovering}
Ethan Perez, Sam Ringer, Kamil{\.e} Luko{\v{s}}i{\=u}t{\.e}, Karina Nguyen, Edwin Chen, Scott Heiner, Craig Pettit, Catherine Olsson, Sandipan Kundu, Saurav Kadavath, et~al. 2022.
\newblock Discovering language model behaviors with model-written evaluations.
\newblock \emph{arXiv preprint arXiv:2212.09251}.

\bibitem[{Pu et~al.(2023)Pu, Gao, and Wan}]{pu2023summarization}
Xiao Pu, Mingqi Gao, and Xiaojun Wan. 2023.
\newblock Summarization is (almost) dead.
\newblock \emph{arXiv preprint arXiv:2309.09558}.

\bibitem[{Qian et~al.(2023)Qian, Zhao, and Wu}]{qian2023merge}
Cheng Qian, Xinran Zhao, and Sherry~Tongshuang Wu. 2023.
\newblock " merge conflicts!" exploring the impacts of external distractors to parametric knowledge graphs.
\newblock \emph{arXiv preprint arXiv:2309.08594}.

\bibitem[{Qiao et~al.(2023)Qiao, Gui, Chen, and Zhang}]{qiao2023making}
Shuofei Qiao, Honghao Gui, Huajun Chen, and Ningyu Zhang. 2023.
\newblock Making language models better tool learners with execution feedback.
\newblock \emph{arXiv preprint arXiv:2305.13068}.

\bibitem[{Qin et~al.(2023)Qin, Hu, Lin, Chen, Ding, Cui, Zeng, Huang, Xiao, Han et~al.}]{qin2023tool}
Yujia Qin, Shengding Hu, Yankai Lin, Weize Chen, Ning Ding, Ganqu Cui, Zheni Zeng, Yufei Huang, Chaojun Xiao, Chi Han, et~al. 2023.
\newblock Tool learning with foundation models.
\newblock \emph{arXiv preprint arXiv:2304.08354}.

\bibitem[{Qiu et~al.(2020)Qiu, Sun, Xu, Shao, Dai, and Huang}]{qiu2020pre}
Xipeng Qiu, Tianxiang Sun, Yige Xu, Yunfan Shao, Ning Dai, and Xuanjing Huang. 2020.
\newblock Pre-trained models for natural language processing: A survey.
\newblock \emph{Science China Technological Sciences}, 63(10):1872--1897.

\bibitem[{Radford et~al.(2019)Radford, Wu, Child, Luan, Amodei, Sutskever et~al.}]{radford2019language}
Alec Radford, Jeffrey Wu, Rewon Child, David Luan, Dario Amodei, Ilya Sutskever, et~al. 2019.
\newblock Language models are unsupervised multitask learners.
\newblock \emph{OpenAI blog}, 1(8):9.

\bibitem[{Radhakrishnan et~al.(2023)Radhakrishnan, Nguyen, Chen, Chen, Denison, Hernandez, Durmus, Hubinger, Kernion, Luko{\v{s}}i{\=u}t{\.e} et~al.}]{radhakrishnan2023question}
Ansh Radhakrishnan, Karina Nguyen, Anna Chen, Carol Chen, Carson Denison, Danny Hernandez, Esin Durmus, Evan Hubinger, Jackson Kernion, Kamil{\.e} Luko{\v{s}}i{\=u}t{\.e}, et~al. 2023.
\newblock Question decomposition improves the faithfulness of model-generated reasoning.
\newblock \emph{arXiv preprint arXiv:2307.11768}.

\bibitem[{Raffel et~al.(2020)Raffel, Shazeer, Roberts, Lee, Narang, Matena, Zhou, Li, and Liu}]{raffel2020exploring}
Colin Raffel, Noam Shazeer, Adam Roberts, Katherine Lee, Sharan Narang, Michael Matena, Yanqi Zhou, Wei Li, and Peter~J Liu. 2020.
\newblock Exploring the limits of transfer learning with a unified text-to-text transformer.
\newblock \emph{The Journal of Machine Learning Research}, 21(1):5485--5551.

\bibitem[{Ram et~al.(2023)Ram, Levine, Dalmedigos, Muhlgay, Shashua, Leyton-Brown, and Shoham}]{ram2023context}
Ori Ram, Yoav Levine, Itay Dalmedigos, Dor Muhlgay, Amnon Shashua, Kevin Leyton-Brown, and Yoav Shoham. 2023.
\newblock In-context retrieval-augmented language models.
\newblock \emph{arXiv preprint arXiv:2302.00083}.

\bibitem[{Ravi et~al.(2024)Ravi, Mielczarek, Kannappan, Kiela, and Qian}]{ravi2024lynx}
Selvan~Sunitha Ravi, Bartosz Mielczarek, Anand Kannappan, Douwe Kiela, and Rebecca Qian. 2024.
\newblock Lynx: An open source hallucination evaluation model.
\newblock \emph{arXiv preprint arXiv:2407.08488}.

\bibitem[{Rawte et~al.(2023)Rawte, Priya, Tonmoy, Zaman, Sheth, and Das}]{rawte2023exploring}
Vipula Rawte, Prachi Priya, SM~Tonmoy, SM~Zaman, Amit Sheth, and Amitava Das. 2023.
\newblock Exploring the relationship between llm hallucinations and prompt linguistic nuances: Readability, formality, and concreteness.
\newblock \emph{arXiv preprint arXiv:2309.11064}.

\bibitem[{Rebuffel et~al.(2022)Rebuffel, Roberti, Soulier, Scoutheeten, Cancelliere, and Gallinari}]{rebuffel2022controlling}
Cl{\'e}ment Rebuffel, Marco Roberti, Laure Soulier, Geoffrey Scoutheeten, Rossella Cancelliere, and Patrick Gallinari. 2022.
\newblock Controlling hallucinations at word level in data-to-text generation.
\newblock \emph{Data Mining and Knowledge Discovery}, pages 1--37.

\bibitem[{Ren et~al.(2023)Ren, Wang, Qu, Zhao, Liu, Tian, Wu, Wen, and Haifeng}]{ren2023boundary}
Ruiyang Ren, Yuhao Wang, Yingqi Qu, Wayne~Xin Zhao, Jing Liu, Hao Tian, Hua Wu, Ji-Rong Wen, and Wang Haifeng. 2023.
\newblock Investigating the factual knowledge boundary of large language models with retrieval augmentation.
\newblock \emph{arXiv preprint arXiv:2307.11019}.

\bibitem[{Roberts et~al.(2020)Roberts, Raffel, and Shazeer}]{roberts2020much}
Adam Roberts, Colin Raffel, and Noam Shazeer. 2020.
\newblock How much knowledge can you pack into the parameters of a language model?
\newblock In \emph{Proceedings of the 2020 Conference on Empirical Methods in Natural Language Processing (EMNLP)}, pages 5418--5426.

\bibitem[{Robertson et~al.(2009)Robertson, Zaragoza et~al.}]{robertson2009probabilistic}
Stephen Robertson, Hugo Zaragoza, et~al. 2009.
\newblock The probabilistic relevance framework: Bm25 and beyond.
\newblock \emph{Foundations and Trends{\textregistered} in Information Retrieval}, 3(4):333--389.

\bibitem[{Scao et~al.(2022)Scao, Fan, Akiki, Pavlick, Ili{\'c}, Hesslow, Castagn{\'e}, Luccioni, Yvon, Gall{\'e} et~al.}]{scao2022bloom}
Teven~Le Scao, Angela Fan, Christopher Akiki, Ellie Pavlick, Suzana Ili{\'c}, Daniel Hesslow, Roman Castagn{\'e}, Alexandra~Sasha Luccioni, Fran{\c{c}}ois Yvon, Matthias Gall{\'e}, et~al. 2022.
\newblock Bloom: A 176b-parameter open-access multilingual language model.
\newblock \emph{arXiv preprint arXiv:2211.05100}.

\bibitem[{Schulman(2023)}]{schulman2023youtube}
John Schulman. 2023.
\newblock \href {https://www.youtube.com/watch?v=hhiLw5Q_UFg} {Reinforcement learning from human feedback: Progress and challenges}.

\bibitem[{Schulman et~al.(2017)Schulman, Wolski, Dhariwal, Radford, and Klimov}]{schulman2017proximal}
John Schulman, Filip Wolski, Prafulla Dhariwal, Alec Radford, and Oleg Klimov. 2017.
\newblock Proximal policy optimization algorithms.
\newblock \emph{arXiv preprint arXiv:1707.06347}.

\bibitem[{Shi et~al.(2024)Shi, Yang, Cai, Zhang, Wang, Yang, and Lam}]{shi2024thoroughexaminationdecodingmethods}
Chufan Shi, Haoran Yang, Deng Cai, Zhisong Zhang, Yifan Wang, Yujiu Yang, and Wai Lam. 2024.
\newblock \href {http://arxiv.org/abs/2402.06925} {A thorough examination of decoding methods in the era of llms}.

\bibitem[{Shi et~al.(2023{\natexlab{a}})Shi, Chen, Misra, Scales, Dohan, Chi, Sch\"{a}rli, and Zhou}]{shi2023large}
Freda Shi, Xinyun Chen, Kanishka Misra, Nathan Scales, David Dohan, Ed~H. Chi, Nathanael Sch\"{a}rli, and Denny Zhou. 2023{\natexlab{a}}.
\newblock Large language models can be easily distracted by irrelevant context.
\newblock In \emph{Proceedings of the 40th International Conference on Machine Learning}, volume 202, pages 31210--31227.

\bibitem[{Shi et~al.(2022)Shi, Fried, Ghazvininejad, Zettlemoyer, and Wang}]{shi-etal-2022-natural}
Freda Shi, Daniel Fried, Marjan Ghazvininejad, Luke Zettlemoyer, and Sida~I. Wang. 2022.
\newblock Natural language to code translation with execution.
\newblock In \emph{Proceedings of the 2022 Conference on Empirical Methods in Natural Language Processing}, pages 3533--3546.

\bibitem[{Shi et~al.(2023{\natexlab{b}})Shi, Han, Lewis, Tsvetkov, Zettlemoyer, and Yih}]{shi2023trusting}
Weijia Shi, Xiaochuang Han, Mike Lewis, Yulia Tsvetkov, Luke Zettlemoyer, and Scott Wen-tau Yih. 2023{\natexlab{b}}.
\newblock Trusting your evidence: Hallucinate less with context-aware decoding.
\newblock \emph{arXiv preprint arXiv:2305.14739}.

\bibitem[{Shi et~al.(2023{\natexlab{c}})Shi, Min, Yasunaga, Seo, James, Lewis, Zettlemoyer, and Yih}]{shi2023replug}
Weijia Shi, Sewon Min, Michihiro Yasunaga, Minjoon Seo, Rich James, Mike Lewis, Luke Zettlemoyer, and Wen-tau Yih. 2023{\natexlab{c}}.
\newblock Replug: Retrieval-augmented black-box language models.
\newblock \emph{arXiv preprint arXiv:2301.12652}.

\bibitem[{Si et~al.(2022)Si, Gan, Yang, Wang, Wang, Boyd-Graber, and Wang}]{si2022prompting}
Chenglei Si, Zhe Gan, Zhengyuan Yang, Shuohang Wang, Jianfeng Wang, Jordan Boyd-Graber, and Lijuan Wang. 2022.
\newblock Prompting gpt-3 to be reliable.
\newblock \emph{arXiv preprint arXiv:2210.09150}.

\bibitem[{Sinitsin et~al.(2020)Sinitsin, Plokhotnyuk, Pyrkin, Popov, and Babenko}]{sinitsin2020editable}
Anton Sinitsin, Vsevolod Plokhotnyuk, Dmitriy Pyrkin, Sergei Popov, and Artem Babenko. 2020.
\newblock Editable neural networks.
\newblock \emph{arXiv preprint arXiv:2004.00345}.

\bibitem[{Su et~al.(2023)Su, Lan, Li, Xu, Wang, and Cai}]{su2023pandagpt}
Yixuan Su, Tian Lan, Huayang Li, Jialu Xu, Yan Wang, and Deng Cai. 2023.
\newblock Pandagpt: One model to instruction-follow them all.
\newblock \emph{arXiv preprint arXiv:2305.16355}.

\bibitem[{Sun et~al.(2023{\natexlab{a}})Sun, Xu, Zha, Liu, and Dong}]{sun2023head}
Kai Sun, Yifan~Ethan Xu, Hanwen Zha, Yue Liu, and Xin~Luna Dong. 2023{\natexlab{a}}.
\newblock Head-to-tail: How knowledgeable are large language models (llm)? aka will llms replace knowledge graphs?
\newblock \emph{arXiv preprint arXiv:2308.10168}.

\bibitem[{Sun et~al.(2022)Sun, Shao, Qian, Huang, and Qiu}]{sun2022black}
Tianxiang Sun, Yunfan Shao, Hong Qian, Xuanjing Huang, and Xipeng Qiu. 2022.
\newblock Black-box tuning for language-model-as-a-service.
\newblock In \emph{International Conference on Machine Learning}, pages 20841--20855. PMLR.

\bibitem[{Sun et~al.(2023{\natexlab{b}})Sun, Zhang, He, Li, Cheng, Yan, Liu, Shao, Tang, Zhao, Chen, Zheng, Zhou, Li, Zhan, Zhou, Li, Yang, Wu, Yin, Huang, and Qiu}]{sun2023moss}
Tianxiang Sun, Xiaotian Zhang, Zhengfu He, Peng Li, Qinyuan Cheng, Hang Yan, Xiangyang Liu, Yunfan Shao, Qiong Tang, Xingjian Zhao, Ke~Chen, Yining Zheng, Zhejian Zhou, Ruixiao Li, Jun Zhan, Yunhua Zhou, Linyang Li, Xiaogui Yang, Lingling Wu, Zhangyue Yin, Xuanjing Huang, and Xipeng Qiu. 2023{\natexlab{b}}.
\newblock Moss: Training conversational language models from synthetic data.

\bibitem[{Tamkin et~al.(2022)Tamkin, Handa, Shrestha, and Goodman}]{tamkin2022task}
Alex Tamkin, Kunal Handa, Avash Shrestha, and Noah Goodman. 2022.
\newblock \href {http://arxiv.org/abs/2212.10711} {Task ambiguity in humans and language models}.

\bibitem[{Taori et~al.(2023)Taori, Gulrajani, Zhang, Dubois, Li, Guestrin, Liang, and Hashimoto}]{alpaca}
Rohan Taori, Ishaan Gulrajani, Tianyi Zhang, Yann Dubois, Xuechen Li, Carlos Guestrin, Percy Liang, and Tatsunori~B. Hashimoto. 2023.
\newblock Stanford alpaca: An instruction-following llama model.
\newblock \url{https://github.com/tatsu-lab/stanford_alpaca}.

\bibitem[{Torabi et~al.(2018)Torabi, Warnell, and Stone}]{torabi2018behavioral}
Faraz Torabi, Garrett Warnell, and Peter Stone. 2018.
\newblock Behavioral cloning from observation.
\newblock In \emph{Proceedings of the 27th International Joint Conference on Artificial Intelligence}, pages 4950--4957.

\bibitem[{Touvron et~al.(2023{\natexlab{a}})Touvron, Lavril, Izacard, Martinet, Lachaux, Lacroix, Rozi{\`e}re, Goyal, Hambro, Azhar et~al.}]{touvron2023llama}
Hugo Touvron, Thibaut Lavril, Gautier Izacard, Xavier Martinet, Marie-Anne Lachaux, Timoth{\'e}e Lacroix, Baptiste Rozi{\`e}re, Naman Goyal, Eric Hambro, Faisal Azhar, et~al. 2023{\natexlab{a}}.
\newblock Llama: Open and efficient foundation language models.
\newblock \emph{arXiv preprint arXiv:2302.13971}.

\bibitem[{Touvron et~al.(2023{\natexlab{b}})Touvron, Martin, Stone, Albert, Almahairi, Babaei, Bashlykov, Batra, Bhargava, Bhosale et~al.}]{touvron2023llama2}
Hugo Touvron, Louis Martin, Kevin Stone, Peter Albert, Amjad Almahairi, Yasmine Babaei, Nikolay Bashlykov, Soumya Batra, Prajjwal Bhargava, Shruti Bhosale, et~al. 2023{\natexlab{b}}.
\newblock Llama 2: Open foundation and fine-tuned chat models.
\newblock \emph{arXiv preprint arXiv:2307.09288}.

\bibitem[{Umapathi et~al.(2023)Umapathi, Pal, and Sankarasubbu}]{umapathi2023med}
Logesh~Kumar Umapathi, Ankit Pal, and Malaikannan Sankarasubbu. 2023.
\newblock Med-halt: Medical domain hallucination test for large language models.
\newblock \emph{arXiv preprint arXiv:2307.15343}.

\bibitem[{Varshney et~al.(2023)Varshney, Yao, Zhang, Chen, and Yu}]{varshney2023stitch}
Neeraj Varshney, Wenlin Yao, Hongming Zhang, Jianshu Chen, and Dong Yu. 2023.
\newblock A stitch in time saves nine: Detecting and mitigating hallucinations of llms by validating low-confidence generation.
\newblock \emph{arXiv preprint arXiv:2307.03987}.

\bibitem[{Vaswani et~al.(2017)Vaswani, Shazeer, Parmar, Uszkoreit, Jones, Gomez, Kaiser, and Polosukhin}]{vaswani2017attention}
Ashish Vaswani, Noam Shazeer, Niki Parmar, Jakob Uszkoreit, Llion Jones, Aidan~N Gomez, {\L}ukasz Kaiser, and Illia Polosukhin. 2017.
\newblock Attention is all you need.
\newblock \emph{Advances in neural information processing systems}, 30.

\bibitem[{Wan et~al.(2024)Wan, Huang, Cui, Quan, Bi, and Shi}]{wan2024knowledge}
Fanqi Wan, Xinting Huang, Leyang Cui, Xiaojun Quan, Wei Bi, and Shuming Shi. 2024.
\newblock \href {http://arxiv.org/abs/2401.10768} {Knowledge verification to nip hallucination in the bud}.

\bibitem[{Wang and Sennrich(2020)}]{wang2020exposure}
Chaojun Wang and Rico Sennrich. 2020.
\newblock On exposure bias, hallucination and domain shift in neural machine translation.
\newblock \emph{arXiv preprint arXiv:2005.03642}.

\bibitem[{Wang et~al.(2023{\natexlab{a}})Wang, Xie, Jiang, Mandlekar, Xiao, Zhu, Fan, and Anandkumar}]{wang2023voyager}
Guanzhi Wang, Yuqi Xie, Yunfan Jiang, Ajay Mandlekar, Chaowei Xiao, Yuke Zhu, Linxi Fan, and Anima Anandkumar. 2023{\natexlab{a}}.
\newblock Voyager: An open-ended embodied agent with large language models.
\newblock \emph{arXiv preprint arXiv:2305.16291}.

\bibitem[{Wang(2019)}]{wang-2019-revisiting}
Hongmin Wang. 2019.
\newblock Revisiting challenges in data-to-text generation with fact grounding.
\newblock In \emph{Proceedings of the 12th International Conference on Natural Language Generation}, pages 311--322.

\bibitem[{Wang et~al.(2022)Wang, Wei, Schuurmans, Le, Chi, Narang, Chowdhery, and Zhou}]{wang2022self}
Xuezhi Wang, Jason Wei, Dale Schuurmans, Quoc~V Le, Ed~H Chi, Sharan Narang, Aakanksha Chowdhery, and Denny Zhou. 2022.
\newblock Self-consistency improves chain of thought reasoning in language models.
\newblock In \emph{The Eleventh International Conference on Learning Representations}.

\bibitem[{Wang et~al.(2023{\natexlab{b}})Wang, Ivison, Dasigi, Hessel, Khot, Chandu, Wadden, MacMillan, Smith, Beltagy et~al.}]{wang2023far}
Yizhong Wang, Hamish Ivison, Pradeep Dasigi, Jack Hessel, Tushar Khot, Khyathi~Raghavi Chandu, David Wadden, Kelsey MacMillan, Noah~A Smith, Iz~Beltagy, et~al. 2023{\natexlab{b}}.
\newblock How far can camels go? exploring the state of instruction tuning on open resources.
\newblock \emph{arXiv preprint arXiv:2306.04751}.

\bibitem[{Wang et~al.(2023{\natexlab{c}})Wang, Kordi, Mishra, Liu, Smith, Khashabi, and Hajishirzi}]{wang-etal-2023-self-instruct}
Yizhong Wang, Yeganeh Kordi, Swaroop Mishra, Alisa Liu, Noah~A. Smith, Daniel Khashabi, and Hannaneh Hajishirzi. 2023{\natexlab{c}}.
\newblock Self-instruct: Aligning language models with self-generated instructions.
\newblock In \emph{Proceedings of the 61st Annual Meeting of the Association for Computational Linguistics (Volume 1: Long Papers)}, pages 13484--13508.

\bibitem[{Wang et~al.(2023{\natexlab{d}})Wang, Mao, Wu, Ge, Wei, and Ji}]{wang2023unleashing}
Zhenhailong Wang, Shaoguang Mao, Wenshan Wu, Tao Ge, Furu Wei, and Heng Ji. 2023{\natexlab{d}}.
\newblock Unleashing cognitive synergy in large language models: A task-solving agent through multi-persona self-collaboration.
\newblock \emph{arXiv preprint arXiv:2307.05300}.

\bibitem[{Wei et~al.(2023{\natexlab{a}})Wei, Haghtalab, and Steinhardt}]{wei2023jailbroken}
Alexander Wei, Nika Haghtalab, and Jacob Steinhardt. 2023{\natexlab{a}}.
\newblock Jailbroken: How does llm safety training fail?
\newblock \emph{arXiv preprint arXiv:2307.02483}.

\bibitem[{Wei et~al.(2021)Wei, Bosma, Zhao, Guu, Yu, Lester, Du, Dai, and Le}]{wei2021finetuned}
Jason Wei, Maarten Bosma, Vincent Zhao, Kelvin Guu, Adams~Wei Yu, Brian Lester, Nan Du, Andrew~M Dai, and Quoc~V Le. 2021.
\newblock Finetuned language models are zero-shot learners.
\newblock In \emph{International Conference on Learning Representations}.

\bibitem[{Wei et~al.(2024{\natexlab{a}})Wei, Karina, Chung, Jiao, Papay, Glaese, Schulman, and Fedus}]{simpleqa}
Jason Wei, Nguyen Karina, Hyung~Won Chung, Yunxin~Joy Jiao, Spencer Papay, Amelia Glaese, John Schulman, and William Fedus. 2024{\natexlab{a}}.
\newblock \href {http://arxiv.org/abs/2411.04368} {Measuring short-form factuality in large language models}.

\bibitem[{Wei et~al.(2022)Wei, Wang, Schuurmans, Bosma, Xia, Chi, Le, Zhou et~al.}]{wei2022chain}
Jason Wei, Xuezhi Wang, Dale Schuurmans, Maarten Bosma, Fei Xia, Ed~Chi, Quoc~V Le, Denny Zhou, et~al. 2022.
\newblock Chain-of-thought prompting elicits reasoning in large language models.
\newblock \emph{Advances in Neural Information Processing Systems}, 35:24824--24837.

\bibitem[{Wei et~al.(2023{\natexlab{b}})Wei, Huang, Lu, Zhou, and Le}]{wei2023simple}
Jerry Wei, Da~Huang, Yifeng Lu, Denny Zhou, and Quoc~V Le. 2023{\natexlab{b}}.
\newblock Simple synthetic data reduces sycophancy in large language models.
\newblock \emph{arXiv preprint arXiv:2308.03958}.

\bibitem[{Wei et~al.(2024{\natexlab{b}})Wei, Yang, Song, Lu, Hu, Tran, Peng, Liu, Huang, Du, and Le}]{safe}
Jerry Wei, Chengrun Yang, Xinying Song, Yifeng Lu, Nathan Hu, Dustin Tran, Daiyi Peng, Ruibo Liu, Da~Huang, Cosmo Du, and Quoc~V. Le. 2024{\natexlab{b}}.
\newblock \href {https://doi.org/10.48550/ARXIV.2403.18802} {Long-form factuality in large language models}.
\newblock \emph{CoRR}, abs/2403.18802.

\bibitem[{Wu and Xiong(2023)}]{wuqafacteval}
Alexander R Fabbri Chien-Sheng Wu and Wenhao Liu~Caiming Xiong. 2023.
\newblock Qafacteval: Improved qa-based factual consistency evaluation for summarization.

\bibitem[{Wu et~al.(2023{\natexlab{a}})Wu, Gaur, Chen, Zhou, Zhu, Wang, Li, Liu, Ren, Liu et~al.}]{wu2023decoder}
Jian Wu, Yashesh Gaur, Zhuo Chen, Long Zhou, Yimeng Zhu, Tianrui Wang, Jinyu Li, Shujie Liu, Bo~Ren, Linquan Liu, et~al. 2023{\natexlab{a}}.
\newblock On decoder-only architecture for speech-to-text and large language model integration.
\newblock \emph{arXiv preprint arXiv:2307.03917}.

\bibitem[{Wu et~al.(2023{\natexlab{b}})Wu, Jiang, Jiang, Xie, and Tu}]{wu2023plms}
Weiqi Wu, Chengyue Jiang, Yong Jiang, Pengjun Xie, and Kewei Tu. 2023{\natexlab{b}}.
\newblock Do plms know and understand ontological knowledge?
\newblock \emph{arXiv preprint arXiv:2309.05936}.

\bibitem[{Wu et~al.(2024)Wu, Hee, Hu, and Lee}]{wu2024longgenbench}
Yuhao Wu, Ming~Shan Hee, Zhiqing Hu, and Roy Ka-Wei Lee. 2024.
\newblock Longgenbench: Benchmarking long-form generation in long context llms.
\newblock \emph{arXiv preprint arXiv:2409.02076}.

\bibitem[{Xiao and Wang(2021)}]{xiao-wang-2021-hallucination}
Yijun Xiao and William~Yang Wang. 2021.
\newblock On hallucination and predictive uncertainty in conditional language generation.
\newblock In \emph{Proceedings of the 16th Conference of the European Chapter of the Association for Computational Linguistics: Main Volume}, pages 2734--2744.

\bibitem[{Xie et~al.(2023)Xie, Zhang, Chen, Lou, and Su}]{xie2023adaptive}
Jian Xie, Kai Zhang, Jiangjie Chen, Renze Lou, and Yu~Su. 2023.
\newblock Adaptive chameleon or stubborn sloth: Unraveling the behavior of large language models in knowledge conflicts.
\newblock \emph{arXiv preprint arXiv:2305.13300}.

\bibitem[{Xiong et~al.(2023)Xiong, Hu, Lu, Li, Fu, He, and Hooi}]{xiong2023can}
Miao Xiong, Zhiyuan Hu, Xinyang Lu, Yifei Li, Jie Fu, Junxian He, and Bryan Hooi. 2023.
\newblock Can llms express their uncertainty? an empirical evaluation of confidence elicitation in llms.
\newblock \emph{arXiv preprint arXiv:2306.13063}.

\bibitem[{Xu et~al.(2023{\natexlab{a}})Xu, Guo, Duan, and McAuley}]{xu2023baize}
Canwen Xu, Daya Guo, Nan Duan, and Julian McAuley. 2023{\natexlab{a}}.
\newblock Baize: An open-source chat model with parameter-efficient tuning on self-chat data.
\newblock \emph{arXiv preprint arXiv:2304.01196}.

\bibitem[{Xu et~al.(2023{\natexlab{b}})Xu, Agrawal, Briakou, Martindale, and Carpuat}]{nmt_halu}
Weijia Xu, Sweta Agrawal, Eleftheria Briakou, Marianna~J. Martindale, and Marine Carpuat. 2023{\natexlab{b}}.
\newblock \href {https://doi.org/10.48550/ARXIV.2301.07779} {Understanding and detecting hallucinations in neural machine translation via model introspection}.
\newblock \emph{CoRR}, abs/2301.07779.

\bibitem[{Yang et~al.(2025)Yang, Tu, Liu, Wang, Zheng, Zhang, Cui, Chen, He, Wang, Ong, and Huang}]{yang2025boundary}
Junxiao Yang, Jinzhe Tu, Haoran Liu, Xiaoce Wang, Chujie Zheng, Zhexin Zhang, Shiyao Cui, Caishun Chen, Tiantian He, Hongning Wang, Yew-Soon Ong, and Minlie Huang. 2025.
\newblock \href {http://arxiv.org/abs/2505.13529} {Barrel: Boundary-aware reasoning for factual and reliable lrms}.

\bibitem[{Yao et~al.(2022)Yao, Zhao, Yu, Du, Shafran, Narasimhan, and Cao}]{yao2022react}
Shunyu Yao, Jeffrey Zhao, Dian Yu, Nan Du, Izhak Shafran, Karthik~R Narasimhan, and Yuan Cao. 2022.
\newblock React: Synergizing reasoning and acting in language models.
\newblock In \emph{The Eleventh International Conference on Learning Representations}.

\bibitem[{Ye et~al.(2023)Ye, Xu, Xu, Ye, Yan, Zhou, Wang, Hu, Shi, Shi et~al.}]{ye2023mplug}
Qinghao Ye, Haiyang Xu, Guohai Xu, Jiabo Ye, Ming Yan, Yiyang Zhou, Junyang Wang, Anwen Hu, Pengcheng Shi, Yaya Shi, et~al. 2023.
\newblock mplug-owl: Modularization empowers large language models with multimodality.
\newblock \emph{arXiv preprint arXiv:2304.14178}.

\bibitem[{Yin et~al.(2023)Yin, Sun, Guo, Wu, Qiu, and Huang}]{yin2023large}
Zhangyue Yin, Qiushi Sun, Qipeng Guo, Jiawen Wu, Xipeng Qiu, and Xuanjing Huang. 2023.
\newblock Do large language models know what they don't know?
\newblock \emph{arXiv preprint arXiv:2305.18153}.

\bibitem[{Yu et~al.(2023{\natexlab{a}})Yu, Wang, Tu, Cao, Zhang-Li, Lv, Peng, Yao, Zhang, Li et~al.}]{yu2023kola}
Jifan Yu, Xiaozhi Wang, Shangqing Tu, Shulin Cao, Daniel Zhang-Li, Xin Lv, Hao Peng, Zijun Yao, Xiaohan Zhang, Hanming Li, et~al. 2023{\natexlab{a}}.
\newblock Kola: Carefully benchmarking world knowledge of large language models.
\newblock \emph{arXiv preprint arXiv:2306.09296}.

\bibitem[{Yu et~al.(2023{\natexlab{b}})Yu, Zhang, Liang, Jiang, and Sabharwal}]{yu2023improving}
Wenhao Yu, Zhihan Zhang, Zhenwen Liang, Meng Jiang, and Ashish Sabharwal. 2023{\natexlab{b}}.
\newblock Improving language models via plug-and-play retrieval feedback.
\newblock \emph{arXiv preprint arXiv:2305.14002}.

\bibitem[{Yue et~al.(2023)Yue, Wang, Zhang, Chen, Su, and Sun}]{yue2023automatic}
Xiang Yue, Boshi Wang, Kai Zhang, Ziru Chen, Yu~Su, and Huan Sun. 2023.
\newblock Automatic evaluation of attribution by large language models.
\newblock \emph{arXiv preprint arXiv:2305.06311}.

\bibitem[{Zarrie{\ss} et~al.(2021)Zarrie{\ss}, Voigt, and Sch{\"u}z}]{zarriess2021decoding}
Sina Zarrie{\ss}, Henrik Voigt, and Simeon Sch{\"u}z. 2021.
\newblock Decoding methods in neural language generation: a survey.
\newblock \emph{Information}, 12(9):355.

\bibitem[{Zeng et~al.(2022)Zeng, Liu, Du, Wang, Lai, Ding, Yang, Xu, Zheng, Xia et~al.}]{zeng2022glm}
Aohan Zeng, Xiao Liu, Zhengxiao Du, Zihan Wang, Hanyu Lai, Ming Ding, Zhuoyi Yang, Yifan Xu, Wendi Zheng, Xiao Xia, et~al. 2022.
\newblock Glm-130b: An open bilingual pre-trained model.
\newblock In \emph{The Eleventh International Conference on Learning Representations}.

\bibitem[{Zeng et~al.(2025)Zeng, Xuan, Cui, and Voigt}]{zeng2025reasoningmodel}
Qingcheng Zeng, Weihao Xuan, Leyang Cui, and Rob Voigt. 2025.
\newblock \href {http://arxiv.org/abs/2504.06564} {Do reasoning models show better verbalized calibration?}

\bibitem[{Zha et~al.(2023)Zha, Yang, Li, and Hu}]{zha-etal-2023-alignscore}
Yuheng Zha, Yichi Yang, Ruichen Li, and Zhiting Hu. 2023.
\newblock {A}lign{S}core: Evaluating factual consistency with a unified alignment function.
\newblock In \emph{Proceedings of the 61st Annual Meeting of the Association for Computational Linguistics (Volume 1: Long Papers)}, pages 11328--11348.

\bibitem[{Zhang et~al.(2025)Zhang, Zhang, Kong, Miao, Ye, and Zheng}]{zhang2025lost}
Junhao Zhang, Richong Zhang, Fanshuang Kong, Ziyang Miao, Yanhan Ye, and Yaowei Zheng. 2025.
\newblock Lost-in-the-middle in long-text generation: Synthetic dataset, evaluation framework, and mitigation.
\newblock \emph{arXiv preprint arXiv:2503.06868}.

\bibitem[{Zhang et~al.(2024{\natexlab{a}})Zhang, Hosseini, Bansal, Kazemi, Kumar, and Agarwal}]{zhang2024generativeverifiersrewardmodeling}
Lunjun Zhang, Arian Hosseini, Hritik Bansal, Mehran Kazemi, Aviral Kumar, and Rishabh Agarwal. 2024{\natexlab{a}}.
\newblock \href {http://arxiv.org/abs/2408.15240} {Generative verifiers: Reward modeling as next-token prediction}.

\bibitem[{Zhang and Choi(2023)}]{zhang2023clarify}
Michael J.~Q. Zhang and Eunsol Choi. 2023.
\newblock \href {http://arxiv.org/abs/2311.09469} {Clarify when necessary: Resolving ambiguity through interaction with lms}.

\bibitem[{Zhang et~al.(2023{\natexlab{a}})Zhang, Press, Merrill, Liu, and Smith}]{zhang2023language}
Muru Zhang, Ofir Press, William Merrill, Alisa Liu, and Noah~A Smith. 2023{\natexlab{a}}.
\newblock How language model hallucinations can snowball.
\newblock \emph{arXiv preprint arXiv:2305.13534}.

\bibitem[{Zhang et~al.(2024{\natexlab{b}})Zhang, Yu, and Feng}]{zhang2024truthx}
Shaolei Zhang, Tian Yu, and Yang Feng. 2024{\natexlab{b}}.
\newblock \href {http://arxiv.org/abs/2402.17811} {Truthx: Alleviating hallucinations by editing large language models in truthful space}.

\bibitem[{Zhang et~al.(2023{\natexlab{b}})Zhang, Dong, Li, Zhang, Sun, Wang, Li, Hu, Zhang, Wu et~al.}]{zhang2023instruction}
Shengyu Zhang, Linfeng Dong, Xiaoya Li, Sen Zhang, Xiaofei Sun, Shuhe Wang, Jiwei Li, Runyi Hu, Tianwei Zhang, Fei Wu, et~al. 2023{\natexlab{b}}.
\newblock Instruction tuning for large language models: A survey.
\newblock \emph{arXiv preprint arXiv:2308.10792}.

\bibitem[{Zhang et~al.(2023{\natexlab{c}})Zhang, Pan, Zhao, and Wang}]{zhang2023mitigating}
Shuo Zhang, Liangming Pan, Junzhou Zhao, and William~Yang Wang. 2023{\natexlab{c}}.
\newblock Mitigating language model hallucination with interactive question-knowledge alignment.
\newblock \emph{arXiv preprint arXiv:2305.13669}.

\bibitem[{Zhang et~al.(2019)Zhang, Kishore, Wu, Weinberger, and Artzi}]{zhang2019bertscore}
Tianyi Zhang, Varsha Kishore, Felix Wu, Kilian~Q Weinberger, and Yoav Artzi. 2019.
\newblock Bertscore: Evaluating text generation with bert.
\newblock In \emph{International Conference on Learning Representations}.

\bibitem[{Zhang et~al.(2023{\natexlab{d}})Zhang, Xia, Couturier, Zheng, Rajmohan, and Ruhle}]{zhang2023hybrid}
Xuchao Zhang, Menglin Xia, Camille Couturier, Guoqing Zheng, Saravan Rajmohan, and Victor Ruhle. 2023{\natexlab{d}}.
\newblock Hybrid retrieval-augmented generation for real-time composition assistance.
\newblock \emph{arXiv preprint arXiv:2308.04215}.

\bibitem[{Zhang et~al.(2023{\natexlab{e}})Zhang, Cui, Bi, and Shi}]{zhang2023alleviating}
Yue Zhang, Leyang Cui, Wei Bi, and Shuming Shi. 2023{\natexlab{e}}.
\newblock \href {http://arxiv.org/abs/2312.15710} {Alleviating hallucinations of large language models through induced hallucinations}.

\bibitem[{Zhao et~al.(2023{\natexlab{a}})Zhao, Li, Joty, Qin, and Bing}]{zhao2023verify}
Ruochen Zhao, Xingxuan Li, Shafiq Joty, Chengwei Qin, and Lidong Bing. 2023{\natexlab{a}}.
\newblock Verify-and-edit: A knowledge-enhanced chain-of-thought framework.
\newblock \emph{arXiv preprint arXiv:2305.03268}.

\bibitem[{Zhao et~al.(2023{\natexlab{b}})Zhao, Wei, Preston, and Poon}]{zhao2023automatic}
Theodore Zhao, Mu~Wei, J~Samuel Preston, and Hoifung Poon. 2023{\natexlab{b}}.
\newblock Automatic calibration and error correction for large language models via pareto optimal self-supervision.
\newblock \emph{arXiv preprint arXiv:2306.16564}.

\bibitem[{Zhao et~al.(2022)Zhao, Liu, Ren, and Wen}]{zhao2022dense}
Wayne~Xin Zhao, Jing Liu, Ruiyang Ren, and Ji-Rong Wen. 2022.
\newblock Dense text retrieval based on pretrained language models: A survey.
\newblock \emph{arXiv preprint arXiv:2211.14876}.

\bibitem[{Zhao et~al.(2023{\natexlab{c}})Zhao, Zhou, Li, Tang, Wang, Hou, Min, Zhang, Zhang, Dong et~al.}]{zhao2023survey}
Wayne~Xin Zhao, Kun Zhou, Junyi Li, Tianyi Tang, Xiaolei Wang, Yupeng Hou, Yingqian Min, Beichen Zhang, Junjie Zhang, Zican Dong, et~al. 2023{\natexlab{c}}.
\newblock A survey of large language models.
\newblock \emph{arXiv preprint arXiv:2303.18223}.

\bibitem[{Zhao et~al.(2024)Zhao, Zhang, Chern, Gao, Liu, He et~al.}]{zhao2024felm}
Yiran Zhao, Jinghan Zhang, I~Chern, Siyang Gao, Pengfei Liu, Junxian He, et~al. 2024.
\newblock Felm: Benchmarking factuality evaluation of large language models.
\newblock \emph{Advances in Neural Information Processing Systems}, 36.

\bibitem[{Zhao et~al.(2020)Zhao, Cohen, and Webber}]{sum_rouge}
Zheng Zhao, Shay~B. Cohen, and Bonnie Webber. 2020.
\newblock \href {http://arxiv.org/abs/2009.13312} {Reducing quantity hallucinations in abstractive summarization}.
\newblock \emph{CoRR}, abs/2009.13312.

\bibitem[{Zheng et~al.(2023{\natexlab{a}})Zheng, Li, Dong, Fan, Wu, Xu, and Chang}]{zheng2023can}
Ce~Zheng, Lei Li, Qingxiu Dong, Yuxuan Fan, Zhiyong Wu, Jingjing Xu, and Baobao Chang. 2023{\natexlab{a}}.
\newblock Can we edit factual knowledge by in-context learning?
\newblock \emph{arXiv preprint arXiv:2305.12740}.

\bibitem[{Zheng et~al.(2023{\natexlab{b}})Zheng, Dou, Gao, Shen, Wang, Liu, Jin, Liu, Xiong, Chen et~al.}]{zheng2023secrets}
Rui Zheng, Shihan Dou, Songyang Gao, Wei Shen, Binghai Wang, Yan Liu, Senjie Jin, Qin Liu, Limao Xiong, Lu~Chen, et~al. 2023{\natexlab{b}}.
\newblock Secrets of rlhf in large language models part i: Ppo.
\newblock \emph{arXiv preprint arXiv:2307.04964}.

\bibitem[{Zheng et~al.(2023{\natexlab{c}})Zheng, Huang, and Chang}]{zheng2023does}
Shen Zheng, Jie Huang, and Kevin Chen-Chuan Chang. 2023{\natexlab{c}}.
\newblock Why does chatgpt fall short in providing truthful answers.
\newblock \emph{arXiv preprint arXiv:2304.10513}.

\bibitem[{Zhong et~al.(2021)Zhong, Yin, Yu, Zaidi, Mutuma, Jha, Awadallah, Celikyilmaz, Liu, Qiu, and Radev}]{DBLP:conf/naacl/ZhongYYZMJACLQR21}
Ming Zhong, Da~Yin, Tao Yu, Ahmad Zaidi, Mutethia Mutuma, Rahul Jha, Ahmed~Hassan Awadallah, Asli Celikyilmaz, Yang Liu, Xipeng Qiu, and Dragomir~R. Radev. 2021.
\newblock \href {https://doi.org/10.18653/v1/2021.naacl-main.472} {Qmsum: {A} new benchmark for query-based multi-domain meeting summarization}.
\newblock In \emph{Proceedings of the 2021 Conference of the North American Chapter of the Association for Computational Linguistics: Human Language Technologies, {NAACL-HLT} 2021, Online, June 6-11, 2021}, pages 5905--5921. Association for Computational Linguistics.

\bibitem[{Zhong et~al.(2023)Zhong, Wu, Manning, Potts, and Chen}]{zhong2023mquake}
Zexuan Zhong, Zhengxuan Wu, Christopher~D Manning, Christopher Potts, and Danqi Chen. 2023.
\newblock Mquake: Assessing knowledge editing in language models via multi-hop questions.
\newblock \emph{arXiv preprint arXiv:2305.14795}.

\bibitem[{Zhou et~al.(2023{\natexlab{a}})Zhou, Liu, Xu, Iyer, Sun, Mao, Ma, Efrat, Yu, Yu et~al.}]{zhou2023lima}
Chunting Zhou, Pengfei Liu, Puxin Xu, Srini Iyer, Jiao Sun, Yuning Mao, Xuezhe Ma, Avia Efrat, Ping Yu, Lili Yu, et~al. 2023{\natexlab{a}}.
\newblock Lima: Less is more for alignment.
\newblock \emph{arXiv preprint arXiv:2305.11206}.

\bibitem[{Zhou et~al.(2021)Zhou, Neubig, Gu, Diab, Guzm{\'{a}}n, Zettlemoyer, and Ghazvininejad}]{neural_gen}
Chunting Zhou, Graham Neubig, Jiatao Gu, Mona~T. Diab, Francisco Guzm{\'{a}}n, Luke Zettlemoyer, and Marjan Ghazvininejad. 2021.
\newblock \href {https://doi.org/10.18653/V1/2021.FINDINGS-ACL.120} {Detecting hallucinated content in conditional neural sequence generation}.
\newblock In \emph{Findings of the Association for Computational Linguistics: {ACL/IJCNLP} 2021, Online Event, August 1-6, 2021}, volume {ACL/IJCNLP} 2021 of \emph{Findings of {ACL}}, pages 1393--1404. Association for Computational Linguistics.

\bibitem[{Zhou et~al.(2023{\natexlab{b}})Zhou, Zhang, Poon, and Chen}]{zhou2023context}
Wenxuan Zhou, Sheng Zhang, Hoifung Poon, and Muhao Chen. 2023{\natexlab{b}}.
\newblock Context-faithful prompting for large language models.
\newblock \emph{arXiv preprint arXiv:2303.11315}.

\bibitem[{Zhu et~al.(2021)Zhu, Hinthorn, Xu, Zeng, Zeng, Huang, and Jiang}]{zhu-etal-2021-enhancing}
Chenguang Zhu, William Hinthorn, Ruochen Xu, Qingkai Zeng, Michael Zeng, Xuedong Huang, and Meng Jiang. 2021.
\newblock Enhancing factual consistency of abstractive summarization.
\newblock In \emph{Proceedings of the 2021 Conference of the North American Chapter of the Association for Computational Linguistics: Human Language Technologies}, pages 718--733.

\bibitem[{Zhu et~al.(2023)Zhu, Wang, Zhou, Wang, Chen, Wang, Yang, Ye, Gong, Zhang et~al.}]{zhu2023promptbench}
Kaijie Zhu, Jindong Wang, Jiaheng Zhou, Zichen Wang, Hao Chen, Yidong Wang, Linyi Yang, Wei Ye, Neil~Zhenqiang Gong, Yue Zhang, et~al. 2023.
\newblock Promptbench: Towards evaluating the robustness of large language models on adversarial prompts.
\newblock \emph{arXiv preprint arXiv:2306.04528}.

\bibitem[{Zou et~al.(2023{\natexlab{a}})Zou, Phan, Chen, Campbell, Guo, Ren, Pan, Yin, Mazeika, Dombrowski, Goel, Li, Byun, Wang, Mallen, Basart, Koyejo, Song, Fredrikson, Kolter, and Hendrycks}]{zou2023representation}
Andy Zou, Long Phan, Sarah Chen, James Campbell, Phillip Guo, Richard Ren, Alexander Pan, Xuwang Yin, Mantas Mazeika, Ann-Kathrin Dombrowski, Shashwat Goel, Nathaniel Li, Michael~J. Byun, Zifan Wang, Alex Mallen, Steven Basart, Sanmi Koyejo, Dawn Song, Matt Fredrikson, J.~Zico Kolter, and Dan Hendrycks. 2023{\natexlab{a}}.
\newblock \href {http://arxiv.org/abs/2310.01405} {Representation engineering: A top-down approach to ai transparency}.

\bibitem[{Zou et~al.(2023{\natexlab{b}})Zou, Wang, Kolter, and Fredrikson}]{zou2023universal}
Andy Zou, Zifan Wang, J~Zico Kolter, and Matt Fredrikson. 2023{\natexlab{b}}.
\newblock Universal and transferable adversarial attacks on aligned language models.
\newblock \emph{arXiv preprint arXiv:2307.15043}.

\end{thebibliography}
\bibliographystyle{acl_natbib}

\end{CJK}
\end{document}